\documentclass[12pt]{amsart}
\usepackage{amsbsy,amssymb,amsmath,amsthm,amscd,amsfonts,latexsym,amstext,delarray,
amsmath,graphicx} 
\usepackage{qtree}
\usepackage[margin=.8in]{geometry}
\usepackage{color}
\usepackage[all]{xy}

\newtheorem{thm}{Theorem}[section]
\newtheorem{prop}{Proposition}[section]
\newtheorem{cor}[thm]{Corollary}
\newtheorem{lem}[thm]{Lemma}

\newtheorem{defn}[prop]{Definition}

\usepackage[normalem]{ulem}

\usepackage[]{mdframed}

\usepackage{bigstrut}

\usepackage{tikz}
\usetikzlibrary{positioning}

\numberwithin{equation}{section}

\def\Z{{\mathbb Z}}
\def\N{{\mathbb N}}

\def\cB{{\mathcal B}}

\def\cF{{\mathcal F}}

\def\cH{{\mathcal H}}

\def\cK{{\mathcal K}}
\def\cL{{\mathcal L}}
\def\cM{{\mathcal M}}

\def\cO{{\mathcal O}}
\def\cP{{\mathcal P}}
\def\cQ{{\mathcal Q}}
\def\cR{{\mathcal R}}
\def\cS{{\mathcal S}}

\def\cV{{\mathcal V}}
\def\cW{{\mathcal W}}

\def\bB{{\mathbb B}}

\def\Hom{{\rm Hom}}

\def\fM{{\mathfrak M}}
\def\fT{{\mathfrak T}}
\def\fF{{\mathfrak F}}
\def\fB{{\mathfrak B}}

\def\fc{{\mathfrak c}}
\def\fm{{\mathfrak m}}
\def\fs{{\mathfrak s}}
\def\fz{{\mathfrak z}}
\def\fh{{\mathfrak h}}
\def\fb{{\mathfrak b}}

\title[Extension Condition]{Extension Condition ``violations" and Merge optimality constraints}
\author[M.Marcolli, R.K.Larson, R.Huijbregts]{Matilde Marcolli, Richard Larson, and Riny Huijbregts}
\date{2025}

\address{Department of Mathematics and Department of Computing and Mathematical Sciences, 
California Institute of Technology, CA 91125, USA}
\email{matilde@caltech.edu}
\address{Department of Linguistics, Stony Brook University, NY 11794-4376, USA}
\email{richard.larson@stonybrook.edu}
\address{Department of Linguistics, Utrecht University, Trans 10, 3512 JK Utrecht, The Netherlands}
\email{m.a.c.huijbregts@uu.nl} 

\begin{document}
\maketitle

\begin{abstract}
We analyze, using the mathematical formulation of Merge within the Strong Minimalist Thesis framework, 
a set of linguistic phenomena, including head-to-head movement, phrasal affixes and
syntactic cliticization, verb-particle alternation, and operator-variable phenomena. These are often
regarded as problematic, as violations of the Extension Condition. We show that, in fact, all of these
phenomena can be explained without involving any EC violation. We first show that derivations
using Sideward Merge are possible for all of these cases: these respect EC, though they involve 
some amount of optimality violations, with respect to Resource Restrictions cost functions, and
the amount of violation differs among these cases. We show that all the cases that involve
large optimality violations can be derived in alternative ways involving neither EC nor the use of SM.
The main remaining case (head-to-head movement) only involves SM with minimal violations of
optimality (near equilibrium fluctuations). We analyze explicitly also the cases of multiple wh-fronting,
clusters of clitics in Romance languages and possessor agreement construction in Korean, and how
an explanation of these phenomena based on SM can be made compatible with the colored operad
generators for phases and theta roles. We also show that the EC condition has a clear
algebraic meaning in the mathematical formulation of Merge and is therefore an intrinsic 
structural algebraic constraint of the model, rather than an additional assumption. 
We also show that the minimal optimality violating SM plays a structural role in the Markovian 
properties of Merge, and we compare different optimality conditions coming from Minimal Search and from
Resource Restriction in terms of their effect on the dynamics of the Hopf algebra Markov chain, in a
simple explicit example. 
\end{abstract}

\setcounter{tocdepth}{4}
\tableofcontents

\section{Introduction}

We work within Chomsky's Minimalism based on free symmetric Merge as formulated in
\cite{ChomskyUCLA}, \cite{ChomskyGK}, \cite{ChomskyGE}, \cite{ChomskyKeio},
and \cite{ChomskyElements}, and with the mathematical formulation of this model,
as developed in \cite{MCB}.
We analyze here a range of different syntactic constructions, 
all apparently representing ``violations" to the {\em Extension Condition} (EC),
which we will recall below in \S \ref{ECsec}. 

\smallskip

We show that the Extension Condition (formulated as the requirement that trees
are always grown at the root in the action of Merge on workspaces) is a structural 
part of the mathematical formulation of Merge, hence it does not need to be assumed 
as an additional requirement. This reflects the fact that the grafting operator (grafting
a forest at a common root) satisfies a universal property in the categorical sense, that
other kind of EC-violating growths of structure do not satisfy. This universal property has 
direct consequences in terms of the syntax-semantics interface. 

\smallskip

On the other hand, the formulation of the Merge action on workspaces 
allows for the presence of forms of Sideward Merge (SM) in addition to External and Internal
Merge. These SM operations do {\em not} violate EC,
but they do violate some optimality conditions expressed either in terms of a 
Minimal Search, or preferably for the type of analysis we discuss here, as an
optimization of certain Resource Restriction (RR) conditions, such as Minimal Yield (MY).

\smallskip

Our goal here is to analyze certain linguistic phenomena listed above, which
can be at first viewed as EC violation problems. The main linguistic cases that we 
will analyze in \S \ref{ECsec} fall into four categories:
\begin{enumerate}
\item head-to-head movement,
\item head-to-phrase movement (including phrasal affixes and syntactic cliticization),
\item phrase-to-head movement (like verb-particle alternation),
\item phrase-to-phrase movement (operator-variable phenomena).
\end{enumerate}

\smallskip

We discuss how to reformulate all of these cases
so that {\em no EC-violations occur}. We first describe
derivations, for each of these cases, involving Sideward Merge (SM). 
Because SM violates optimality constraints, an explanation based on SM 
requires a quantitative analysis of the optimality violation, in order to
assess the reliability of the explanation. We provide a quantitative
measure of how far from optimality each SM derivation is. We then
show that all the cases involving large violations are amenable to
other types of non-EC-violating derivations that do not involve SM
(but that in some cases require the introduction of additional structure
that is currently not yet fully part of the mathematical formulation of Minimalism).
The remaining case of head-to-head movement
only involves SM derivations with minimal optimality violations, hence
SM appears to be the best explanation. We also show that there
are structural mathematical reasons for expecting the existence of
forms of SM with minimal optimality violation, related to 
the Markovian property of Merge. From the analysis of the Hopf
algebra Markov chain properties we also see, in a different way,
a mathematical reason for the need to weight the different type of 
Merge operations (External Merge, Internal Merge, Sideward Merge) 
via some appropriate cost function. 

\smallskip

One can interpret our analysis of optimality violations in the 
same way one would in an analogous mathematical
model in the physical sciences. Optimality constraints (like the minima
of an action functional) are different in nature from structural
constraints of the underlying kinematics of the physical model. The
latter are {\em hard constraints} (as is, in our case the Extension Condition:
it is a structural constraint built into the algebra of the model). The
action/energy minimizations, on the other hand, are {\em soft constraints}.
The terminology hard/soft constraint is of standard use in the mathematical
theory of constrained optimization: it does not mean, of course, that
soft constraint are less important than hard ones, just that they need
to be analyzed in a different way. For example,
one can expand an action/energy functional around a minimum and one can consider
points that are closer or farther away from a minimizing solution. Thus,
in physical models, while hard constraints are directly built into the
underlying geometry, soft constrains are typically added as terms 
(Lagrange multipliers) into the action functional itself.
The same distinction between hard and soft constraints exists in
other settings in the general theory of constrained optimization, 
where hard constraints are structural to the problem (here part
of the algebraic structure) whereas soft constraints are also 
typically associated with an increasing cost function, {\em close to}
but not necessarily {\em at} its minimum points. Here Resource Restrictions constraints
should be regarded as the {\em soft constraint}, as opposed
to the Extension Condition that is a {\em hard constraint}. This
means that derivations based on SM are violations of a soft
constraint and therefore come with an associated quantitative
degree of (un)acceptability, unlike violations of hard constraints
that are simply not possible in the model. In other words,
while in a physical system violation of structural constraints
is ruled out (and would constitute a falsification of the theory), 
violations of optimization soft constraints are not impossible, 
but are bound to happen very rarely, all the
more rarely the more significant the violation. 

\smallskip

We approach the analysis of these cases using the mathematical formulation
of Merge of \cite{MCB}. The main aspects of the mathematical model that are
relevant to this question are the following:
\begin{itemize}
\item EC is a more fundamental part of the model than the minimality constraints
(MS, RR, and MY): derivations involving EC violations are excluded by the model, while
derivations involving SM are part of the model, though disfavored by optimality.
\item These optimality measures are completely quantified and the amount of violation 
of minimality can be explicitly computed.
\item Our computation of violations shows that the four different linguistic cases 
we analyze correspond  exactly to different possible degrees of violation of the Merge 
optimality constraints.
\item Only the first class of cases (head-to-head movement) corresponds to
a minimal violation of all the optimality RR constraints: the three remaining cases
involve at least one optimality condition (or all of them simultaneously, as in the last case) that 
can be arbitrarily large.
\end{itemize}

\smallskip

Thus, when we analyze the phenomena listed above from the
point of view of a proposed derivation in terms of SM, we
find that an SM-based derivation is strongly disfavored by
optimality in all but the first case, with a gradient of
(un)acceptability that corresponds to the order, from (1) to (4), in which
they are listed above: with the second case presenting 
a less significant violation of the third case, which in turn
presents a less significant violation of optimality than the
fourth case. This reflects the empirical linguistic evidence
around the different classes of phenomena that we analyze. 

\smallskip

This ranking of the optimality violations shows that,
in all but the first case (which has minimal violations),
the explanation using SM is disfavored (large fluctuation
very far from equilibrium/optimality are highly unlikely).
This implies that these large-violations cases should
have an alternative explanation that still does not involve
EC violations but also does not require SM. 

\smallskip

We propose in \S \ref{ExtSec}  a different analysis for those phenomena 
that involve large violations of optimality, and alternative possible
explanations by derivations not involving the use of SM,
that maintain the RR-optimality intact (in addition to avoiding EC-violations). 

\smallskip

Our analysis of the various cases of apparent EC violations, and the
behavior of SM and optimality violations, 
can be regarded as providing a simple test of ``linguistic predictions" that
the mathematical model can make, as shown in \S \ref{MYSec} where
the computation of degrees of violation of RR-optimality matches 
the four different types of linguistic phenomena. At the same time,
we also see that the analysis of these specific linguistic phenomena
and possible alternative explanations aimed at avoiding large RR-optimality
violations suggest further aspects of the mathematical model that are indeed
still under development and that we will discuss briefly at the 
end of the paper. These involve, on the one hand,  further new structures in the 
theory, such as Chomsky's new Box Theory, \cite{ChomskyKeio}, \cite{Chomsky23a},
\cite{Chomsky23b}, or possible additional algebraic structures related to 
the morphosyntax interface, as modeled
in \cite{SentMar}.

\smallskip
\subsection{Structure of the paper}\label{PlanSec} 

We first provide in \S \ref{ECsec} a description of the
four types of linguistic phenomena that we analyze in this
paper and we discuss how these are sometimes 
presented in the linguistics literature as possible problems
of violations of the Extension Condition (EC).
We then introduce alternative derivations that avoid
EC-violations. In \S \ref{AltSMsec} we present a possible
description in terms of Sideward Merge (SM) which
does {\em not} violate EC but does violate 
Resource Restriction optimality constraints. We delay to
 \S \ref{MYSec} to show how one ranks the SM derivations 
of these four cases according to how much optimality is violated,
since that argument requires first recalling some mathematical
background. In \S \ref{ExtSec} we discuss how some of these
cases also admit a possible derivation that 
avoids both structural EC-violations and optimality RR-violations,
and we highlight what additional structure are needed for these
alternative derivations. 
We also discuss why, on linguistic ground, the remaining
case of head-to-head movement should always involve
an SM immediately followed by an EM. A mathematical
argument for why this is the case is given in \S \ref{EMSMnearIM}.

\smallskip

In the rest of the paper we discuss these cases and their possible 
explanations from the point of view of the mathematical formulation of Merge
of \cite{MCB}. 

\smallskip

We fist give in \S \ref{MathMinSec}  a quick summary for the readers
of the mathematical formulation of Marcolli--Chomsky--Berwick (as developed in \cite{MCB})
of Merge within the general framework of Chomsky's
Strong Minimalist Thesis (SMT), \cite{ChomskyElements}.

In this section, we focus in particular on three aspects of the model, all of them 
already presented in \cite{MCB}:
\begin{itemize}
\item the way in which EC occurs as an intrinsic algebraic constraint
of the model (and is {\em not} taken as an assumption) 
\item the different types of Resource Restriction conditions present in the model
(Minimal Yield and the No Complexity Loss principle)
\item the formulation of Minimal Search as zero-cost condition for an
appropriate cost functional. 
\end{itemize}
In addition to recalling these properties, we develop in \S \ref{MYSec} a new
detailed discussion of the different forms of violation of RR-optimality and
how to evaluate quantitatively the magnitude of the violation. 

\smallskip

The fact that all the four cases analyzed appear at first to be
violations of EC, but turn out to have other forms
of derivation (with or without SM) that do not violate EC
is a fact that can be a priori explained by the
mathematical model. In fact, although the type of
movement that gives EC violations (insertions at
lower levels of the tree away from the root) is not
structurally admissible in the mathematical model of the 
action of Merge on workspaces, it is describable
in that model as the dual ``insertion Lie algebra"
of the Hopf algebra of workspaces that defines
Merge, as discussed in \S 1.7 of \cite{MCB}, 
and this has as direct consequence that such insertions
are not additional operations but are derivable from
the given algebraic structure. This implies that 
constructions based on such insertions
should admit a different yet equivalent construction 
that only uses data from the Hopf algebra of
workspaces itself, which includes the Merge
operations (including SM). 

\smallskip

In \S \ref{ECdualHsec} we analyze how the Extension Condition
arises from the algebraic properties of the model, by showing that,
unlike EC-violating structure formation operations, the grafting 
of trees at the root involved in the Merge operation has a specific
algebraic property (Hochschild cocycle), which satisfies a universal
property that has direct relevance to the syntax-semantics interface.
Thus, EC is not an additional assumption, but a direct consequence of
the algebraic structure.

\smallskip

In \S \ref{CostsSec} we discuss different types of cost functions:
Minimal Search, Minimal Yield, and No Complexity Loss (the last
two combined as forms of Resource Restriction). We show that
External Merge and Internal Merge are cost optimizing while
different forms of Sideward Merge have different levels of
optimality violation and can be ranked accordingly. 

\smallskip

Using the mathematical formulation of Merge and Minimalism 
developed in \cite{MCB}, we then show in \S \ref{MYSec} that the four different cases
analyzed correspond to four different types of RR-optimality
violations, in increasing order of difficulty, in the sense of
incurring in larger and more severe forms of violations. Thus,
from this quantitative analysis we derive the fact that only
the first type could represent a ``minimal RR-optimality violation",
whereas all the other three types allow for arbitrarily large violations
of increasingly problematic form. This makes the SM-based
proposed explanation more unlikely for three out of the four cases,
and more likely for the first case. 

\smallskip

In \S \ref{BulgSMsec} we return to analyze the SM model
of multiple wh-fronting and the tucking-in problem, and in
\S \ref{SMdisemSec}
we discuss the implications that this explanation has on 
the role of SM in the dichotomy in semantics between EM and IM
with respect to the theta-roles structure. While, as mentioned
in \cite{MarLar} it looks at first like SM will act like an EM in this
dichotomy, this kind of constructions suggest a more complex
role for SM that is partly behaving like an EM and partly like an IM
with respect to this dichotomy.

\smallskip

In \S \ref{EMSMling} we analyze different phenomena, namely
head-to-head movement, multiple wh-fronting,
clusters of clitics in Romance languages and possessor agreement 
construction in Korean, from the point of view of an explanation
based on Sideward Merge and we explain how such explanation
can be made compatible with the modeling of theta roles and
of the structure of phases via colored operads and bud generating
systems developed in \cite{MarLar} and \cite{MHL}. We show that
a simple modification of the generators accounts for all of these
cases, and avoids unwanted construction while favoring a
combination of an SM immediately followed by an EM (which we
already observed in previous sections is, in terms of cost
counting, most similar to an IM). 

\smallskip

In \S \ref{HopfMArkovSec}
we discuss how the existence of some form of SM is structurally
needed for the Markovian properties of Merge. We illustrate this
fact by computing a very simple explicit example of the dynamics
of the Hopf algebra Markov chain given by the action of
Merge on workspaces: the case of workspaces with only three leaves.
The advantage of this case is that it is simple enough that the
entire behavior of the dynamics can be computed explicitly by hand,
and one can directly see the reason why SM plays a crucial structural
role in the dynamics, as well as the crucial role of the cost functions.
We do not analyze in this paper the more general case of an arbitrary
number of leaves, as that will be discussed elsewhere and this
simple example suffices to prove the necessity of SM to ensure
the strong connectedness property needed for the Markovian behavior of Merge. 
We do prove one general result, that applies to the Hopf algebra Markov chain
of Merge on workspaces of arbitrary size and that is important for
the problem we are discussing in this paper. We show that to obtain the
strong connectedness property the form of SM with the minimal optimality
violation suffices. Thus, on purely mathematical ground, we see that
one expects the existence of linguistic phenomena that require
minimal-optimality-violating SM, which in this paper we identify as
being the head-to-head movement. A more general discussion of
the Merge Hopf algebra Markov chain as a dynamical system and
the general properties of the dynamics is dealt separately in \cite{MarSkig}:
for the purpose of the present paper we only present the simplest
example of the dynamics, with only three leaves.

\smallskip

We then discuss the implications for the current theoretical
model of the alternative  explanations presented in \S \ref{ExtSec} 
that do not use SM and have neither EC-violations nor RR-optimality
violations. We show that some of these alternative proposals for the 
cases involving SM with large optimality violations will require extensions 
of the theoretical model, which would include developing the appropriate 
mathematical formulation of Chomsky's new Box Theory 
(see \cite{ChomskyKeio},  \cite{Chomsky23a}, \cite{Chomsky23b}),
as well as certain transformations that are best described in the
context of the Morphosyntax extension of the mathematical model
of Merge developed in \cite{SentMar}.

\smallskip

Readers primarily interested in the linguistic implications of our results
may skip several of the more extensive mathematical parts of the paper
(\S \ref{ECdualHsec}, \S \ref{CostsSec}, and \S \ref{HopfMArkovSec}) 
and focus on \S \ref{ECsec}, \S \ref{ExtSec}, \S \ref{MYSec}, and \S \ref{ThSec} 
where the problem and the main 
linguistic consequences of the mathematical models are discussed.

\smallskip
\section{A list of apparent Extension Condition ``violations"} \label{ECsec} 

Chomsky \cite{Chomsky93}, \cite{Chomsky95} proposes that syntactic 
composition obeys an {\em Extension Condition} (EC), which Matushansky \cite{Matu} 
formulates concisely as follows.

\begin{defn}\label{ECdef}
{\rm Extension Condition}: Merge should be effected at the root. 
\end{defn}

Given Merge as the sole structure-building operation in grammar, EC amounts to 
requiring that 
\begin{enumerate}
\item syntactic compositions always ``grow" or expand structure; the result 
should be larger than the components; 
\item such growth happens at the
root of the tree and not by insertions at any lower vertices/edges. 
\end{enumerate}
We will discuss in \S \ref{MathMinSec} how exactly both of these conditions are 
reflected in the mathematical formulation of \cite{MCB}. The first condition is what
is referred to in \cite{MCB} and here as the No Complexity Loss condition, while the
second, which we focus on especially as the crucial part of the Extension Condition,
is intrinsic in the algebraic properties of the model of \cite{MCB}. The first condition here
will be subsumed in one of the cost functions that describe optimality. It will turn out
that this cost function is even more basic and fundamental than Minimal Search
in terms of the effect it has on the dynamics of the action of Merge on workspaces
(as we discuss in \S \ref{HopfMArkovSec}). The second condition, on the other hand,
the core of EC, is a purely algebraic condition, whose relevance and important role
we discuss in \S \ref{ECdualHsec}. 

\smallskip

As noted in Chomsky \cite{Chomsky93}, \cite{Chomsky95}, all cases of External Merge (EM) 
satisfy EC trivially.  Thus, if syntactic objects $X$ and $Y$ are in the workspace together \eqref{2a} 
and are externally merged \eqref{2b}, then the result expands the root and grows the structure, 
whichever element is considered root (labels the root), and whether $X$ and $Y$ are both atoms 
(elementary syntactic objects), one is an atom and the other a phrase, or both are phrases.
\begin{align}\label{2a}
 X, Y  \ \ \ \ &  \text{ (or } X\sqcup Y \text{ in the notation of \cite{MCB}) }  
\end{align}
\begin{align}\label{2b}
\text{\bf EM}:   \ \   [ X, Y ] \ \   \ \  &  \text{ (or }  \fM(X,Y)=\Tree[ $X$ $Y$ ]  \text{ in the notation of \cite{MCB}) } 
\end{align}

\smallskip

Conformity with EC also holds in the most familiar cases of Internal Merge (IM), 
such as movement to specifier position \eqref{3a}--\eqref{3b} or adjunction to a containing phrase 
\eqref{4a}--\eqref{4b}. 
\begin{align}\label{3a}
 [ \, X\,\,  [ \ldots YP \ldots ] ]  & \ \ \    \text{ or with tree notation:}  \Tree[ $X$ [$\cdots$ [ [ [ $\cdots$ $YP$ ] $\cdots$ ] $\cdots$ ] ] ]   
\end{align}
\begin{align}\label{3b}
\text{\bf  IM}:  \ \  [ \, YP\,\,  [ \, X\,\,  [ \ldots \text{\sout{$YP$}} \ldots ] ] ]  &   \ \ \  \text{ or with tree notation:}  \Tree[ $YP$ [ $X$ [$\cdots$ [ [ [ $\cdots$ \text{\sout{$YP$}} ] $\cdots$ ] $\cdots$ ] ] ] ]   
\end{align}
\begin{align}\label{4a}
 [ \ldots XP \ldots ]  &  \ \ \ \  \text{ or with tree notation }    \Tree[ $\cdots$ [ [ $\cdots$ $XP$ ] $\cdots$ ] ]   
\end{align}
\begin{align}\label{4b}
\text{\bf  IM}: \ \ [ \, XP\, [ \ldots \text{\sout{$XP$}} \ldots ] ]  \ \ \ \   &  \text{ or with tree notation }    
 \Tree[ $XP$ [ $\cdots$ [ [ $\cdots$ \text{\sout{$XP$}} ] $\cdots$ ] ] ]
\end{align}
In both cases, Merge is effected at the root, satisfying EC.

\smallskip

Nonetheless, conformity with EC is not ``automatic" in the case of IM, as it is with EM. 
Consider the schematic derivation in \eqref{5a}--\eqref{5c}, where ``$X-Y$" represents adjunction of $Y$ to $X$.
\begin{align}\label{5a}
X, \,\,   [ \ldots Y \ldots ]   &    \ \ \text{ or with tree notation: } \ \  X\,  \sqcup \, \,  \Tree[ $\cdots$  [ [ $Y$ $\cdots$ ]   $\cdots$ ] ] 
\end{align}
\begin{align}\label{5b}
 \text{\bf EM}: \ \ \   [\, X \,\,  [ \ldots Y \ldots ] ]  & \ \   \text{ or with tree notation: } \ \ \Tree[ $X$ [ $\cdots$  [ [ $Y$ $\cdots$ ]   $\cdots$ ] ] ] 
\end{align}
\begin{align}\label{5c}
 \text{\bf ``IM"}: \ \     [\, X-Y  \,\,  [ \ldots \text{\sout{$Y$}} \ldots ] ]   & \ \   \text{ or with tree notation: } \ \  \Tree[ [ $X$ $Y$ ] [ $\cdots$  [ [ \text{\sout{$Y$}} $\cdots$ ]  $\cdots$ ] ] ] 
\end{align}
Movement in \eqref{5c} is purely internal to \eqref{5b}. In a scenario like \eqref{5c} we would have a putative
form of  ``IM" that takes place, not at the root as in \eqref{5a}--\eqref{5b}, 
but rather at one of its subterms ($X$). Derivations of the form in \eqref{5a}--\eqref{5c} 
therefore violate the Extension Condition.

We will show in \S \ref{ECdualHsec} that the mathematical formulation of \cite{MCB} excludes operations 
like \eqref{5c} with insertions at non-root vertices for purely algebraic reasons.

\smallskip

Interestingly, derivations as in \eqref{5a}--\eqref{5c}  have been proposed in the linguistics 
literature for all the possible combinations of $X$ and $Y$: where $X$ and $Y$ are both atoms, 
where one is an atom and the other a phrase, and where both are phrases. 
Moreover, these analyses have involved empirical phenomena that many linguists 
would regard as core. The status of EC therefore acquires considerable importance in this context.

\smallskip
\subsection{Head-to-Head Movement}\label{HtoHsec}

The case of \eqref{5a}--\eqref{5c}, where $X$ and $Y$ are syntactic atoms; i.e., heads, 
exemplifies head-to-head movement \eqref{6a}--\eqref{6b}: 
\begin{align}\label{6a}
X, \,\,   [_{\rm YP} \ldots Y \ldots ]   &  \ \  \stackrel{\text{\bf EM}}{\longrightarrow} & [\, X \,\,  [_{\rm YP} \ldots Y \ldots ] ]  &
 \ \  \stackrel{\text{\bf ``IM"}}{\longrightarrow} &  [\, X-Y  \,\,  [_{\rm YP} \ldots \text{\sout{$Y$}} \ldots ] ] 
\end{align}
or equivalently, written in the tree notation,
\begin{equation}\label{6b}
X \sqcup \Tree[ .${\rm YP}$ $\cdots$ [ [ $\cdots$ $Y$ ] $\cdots$ ] ]   \,  \stackrel{\text{\bf EM}}{\longrightarrow} \,
\Tree[ $X$ [.${\rm YP}$ $\cdots$ [ [ $\cdots$ $Y$ ] $\cdots$ ] ] ]  \,  \stackrel{\text{\bf ``IM"}}{\longrightarrow}  \, 
\Tree[ [ $X$ $Y$ ]  [.${\rm YP}$ $\cdots$ [ [ $\cdots$ \text{\sout{$Y$}} ] $\cdots$ ] ] ]  
\end{equation}

Head movement has been proposed to account for a wide variety of apparent $X^0$ 
displacements, including English ``subject auxiliary inversion" (\ref{7ad}.a), French V-to-T movement (\ref{7ad}.b), 
Germanic ``verb second" (\ref{7ad}.c), verb-initial word order in VSO languages like Welsh (\ref{7ad}.d), 
and all manner of incorporations, like (\ref{7ad}.e) from Malayalam \cite{Baker}.
\begin{equation}\label{7ad}
\includegraphics[scale=0.55]{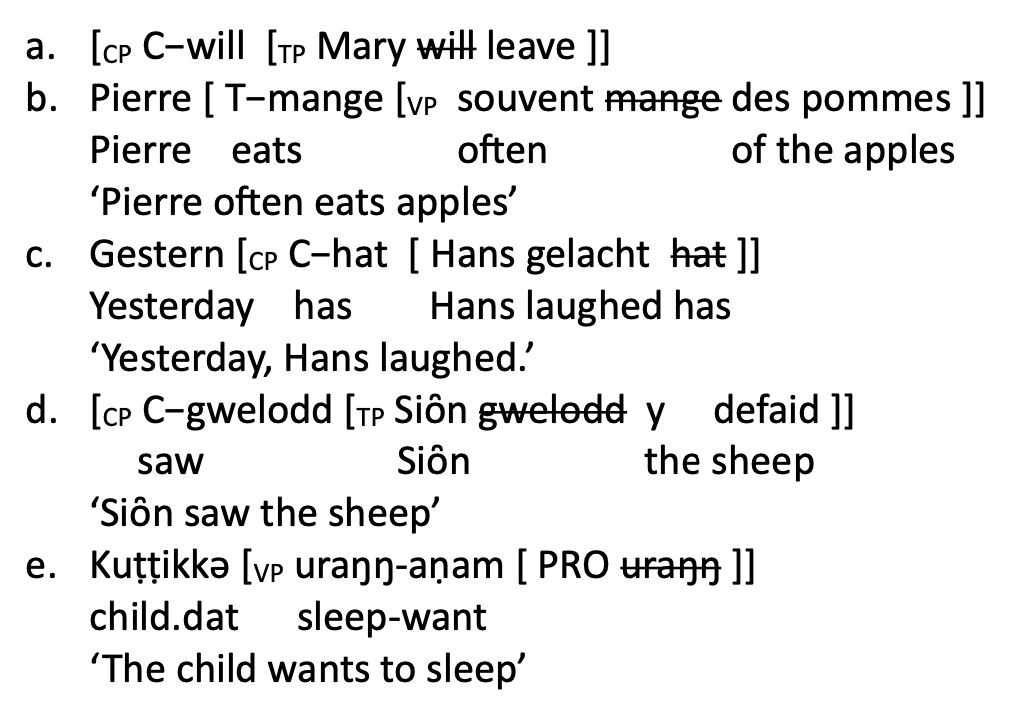}
\end{equation}
Exclusion of the \eqref{6a}--\eqref{6b}, where $X$ and $Y$ are atoms, would therefore 
demand throughgoing re-evaluation of many empirical results that crucially assume its availability.

We will discuss how to resolve this problem without EC-violations in 
 \S \ref{ECsec}, \S \ref{ExtSec}, and \S \ref{MYSec}.

\smallskip
\subsection{Head-to-Phrase Movement}\label{HtoPsec}

Consider next the case of \eqref{5a}--\eqref{5c} where $X$ is a phrase and $Y$ is an atom -- i.e., 
head-to-phrase movement \eqref{8a}--\eqref{8b}: 
\begin{align}\label{8a}
{\rm XP}, \,\,   [_{\rm YP} \ldots Y \ldots ]   &  \ \  \stackrel{\text{\bf EM}}{\longrightarrow} & [\, XP \,\,  [_{\rm YP} \ldots Y \ldots ] ]   &
 \ \  \stackrel{\text{\bf ``IM"}}{\longrightarrow} &   [\, XP-Y  \,\,  [_{\rm YP} \ldots \text{\sout{$Y$}} \ldots ] ] 
\end{align}
or equivalently, written in the tree notation,
\begin{equation}\label{8b}
{\rm XP} \sqcup \Tree[ .${\rm YP}$ $\cdots$ [ [ $\cdots$ $Y$ ] $\cdots$ ] ]   \,  \stackrel{\text{\bf EM}}{\longrightarrow} \,
\Tree[ ${\rm XP}$ [.${\rm YP}$ $\cdots$ [ [ $\cdots$ $Y$ ] $\cdots$ ] ] ]  \,  \stackrel{\text{\bf ``IM"}}{\longrightarrow}  \, 
\Tree[ [ ${\rm XP}$ $Y$ ]  [.${\rm YP}$ $\cdots$ [ [ $\cdots$ \text{\sout{$Y$}} ] $\cdots$ ] ] ]  
\end{equation}

Plausible instances of \eqref{8a}--\eqref{8b} include ``phrasal affixes" or ``syntactic clitics", 
the English genitive morpheme --'s being a well-known example. As noted by Anderson, \cite{Anderson}, 
the English prenominal genitive inflection is realized on a variety of items, from lexical words (\ref{9ae}.a) 
to much larger phrases (\ref{9ae}.b--e). In all cases, 's appears at a right edge, whatever the 
categorial identity or grammatical function of the item it attaches to.
\begin{equation}\label{9ae}
\includegraphics[scale=0.55]{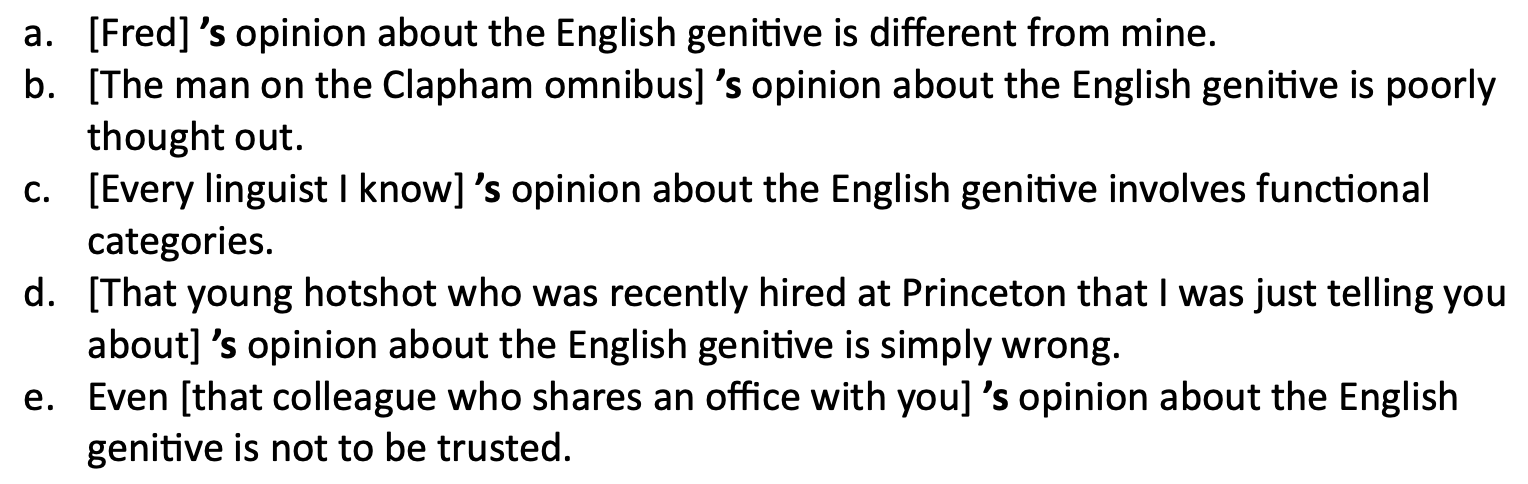}
\end{equation}

A widely discussed account of the construction, due to Abney \cite{Abney}, 
analyzes 's as a determiner, heading a DP, with the possessor occupying its 
Spec position  (\ref{10}.a), cf.~(\ref{9ae}.c). Final positioning of the D head 
derives by adjunction to the possessor phrase, as in  (\ref{10}.b), which 
exemplifies \eqref{8b}.   
\begin{equation}\label{10}
\includegraphics[scale=0.55]{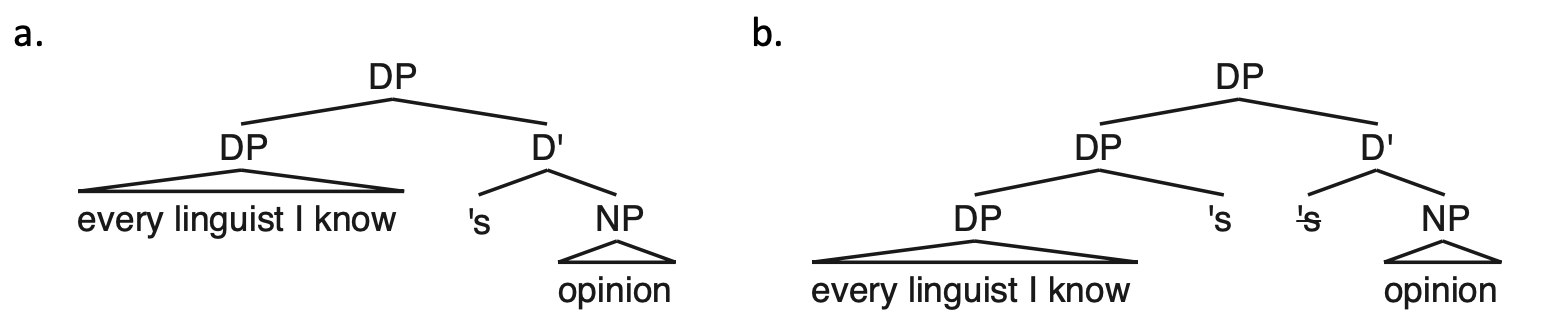}
\end{equation}

Phrasal affixation/syntactic cliticization is a common process in world languages, 
plausibly analyzed as head-to-phrase movement \cite{Anderson}. Once again, 
therefore, proscription of derivations equivalent to \eqref{8a}--\eqref{8b} would 
demand serious reevaluation of a raft of empirical results.

We will return to discuss this case in  \S \ref{ECsec}, \S \ref{ExtSec}, \S \ref{MYSec}.

\smallskip
\subsection{Phrase-to-Head Movement} \label{PtoHsec}

The counterpart of \eqref{8a}--\eqref{8b} is \eqref{11a}--\eqref{11b}, where $X$ is a head 
and $Y$ is phrase -- i.e., phrase-to-head movement: 
\begin{align}\label{11a}
X, \,\,   [_{ZP} \ldots YP \ldots ]  &  \ \  \stackrel{\text{\bf EM}}{\longrightarrow} & [\, X \,\,  [_{ZP} \ldots Y \ldots ] ]   &
 \ \  \stackrel{\text{\bf ``IM"}}{\longrightarrow} &  [\, X-YP  \,\,  [_{ZP} \ldots \text{\sout{$YP$}} \ldots ] ] 
\end{align}
or equivalently, written in the tree notation,
\begin{equation}\label{11b}
X \sqcup \Tree[ .${\rm ZP}$ $\cdots$ [ [ $\cdots$ ${\rm YP}$ ] $\cdots$ ] ]   \,  \stackrel{\text{\bf EM}}{\longrightarrow} \,
\Tree[ $X$ [.${\rm ZP}$ $\cdots$ [ [ $\cdots$ ${\rm YP}$ ] $\cdots$ ] ] ]  \,  \stackrel{\text{\bf ``IM"}}{\longrightarrow}  \, 
\Tree[ [ $X$ ${\rm YP}$ ]  [.${\rm ZP}$ $\cdots$ [ [ $\cdots$ \text{\sout{ ${\rm YP}$ }} ] $\cdots$ ] ] ]  
\end{equation}

Once again, derivations equivalent to \eqref{11a}--\eqref{11b} have 
been proposed in the literature.  Consider the so-called ``verb-particle 
alternation" illustrated in (\ref{12}.a--c).  In view of their notional 
status as a single, complex predicate, the elements look and up 
are widely regarded as forming an underlying syntactic constituent, 
whose separation as in (\ref{12}.b) is due to some form of displacement. 
\begin{equation}\label{12}
\includegraphics[scale=0.55]{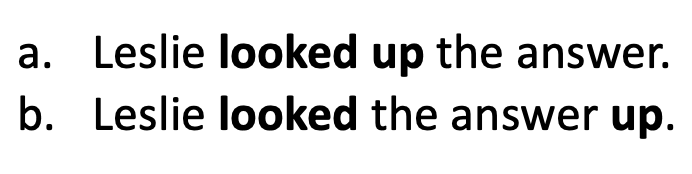}
\end{equation}

One idea, offered in \cite{Chomsky55}, is that (\ref{12}.b) derives by movement 
of the particle rightward. A more modern view, following developments in \cite{Chomsky93}, 
analyzes both (\ref{12}.a) and (\ref{12}.b) as involving movement, but differing in 
what/how much moves. Specifically, verb phrases start with an articulated v-VP structure 
shown in (\ref{13}.a) with the predicate then raising to v. If the verb alone raises, 
as in (\ref{13}.b), we get (\ref{12}.b), with the verb and particle separated. 
This is an example of head-to-head movement. But if the complex predicate raises 
as a group, as in (\ref{13}.c), we get (\ref{12}.a), with the verb and particle 
maintaining contiguity. 
\begin{equation}\label{13}
\includegraphics[scale=0.55]{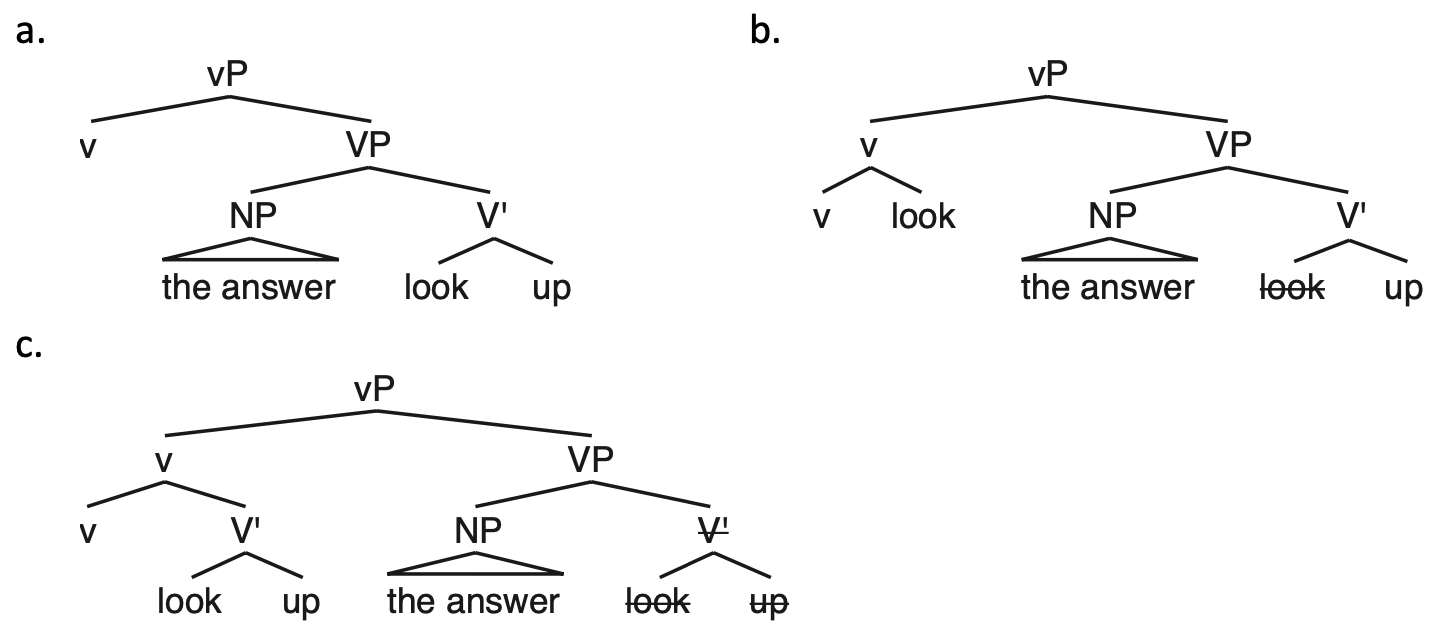}
\end{equation}

Here (\ref{13}.c) exemplifies \eqref{11b}: phrase-to-head movement. 
Perhaps more intuitively, (\ref{13}.c) illustrates the case where the 
grammar treats a constituent as a complex head -- a complex $X^0$, 
a phenomenon observed elsewhere in the grammar, particularly 
in the domain of word formation,~\cite{Bruening}.  

We will discuss these cases further in  \S \ref{ECsec}, \S \ref{ExtSec}, \S \ref{MYSec}.

\smallskip

\subsection{Phrase-to-Phrase Movement} \label{PtoPsec}

Finally, we may consider \eqref{14a}--\eqref{14b}, where $X$ and $Y$ are both phrases -- i.e., phrase-to-phrase movement: 
\begin{align}\label{14a}
{\rm XP}, \,\,   [_{\rm ZP} \ldots {\rm YP} \ldots ]  &  \ \  \stackrel{\text{\bf EM}}{\longrightarrow} & [\, {\rm XP} \,\,  [_{\rm ZP} \ldots {\rm YP} \ldots ] ]  &
 \ \  \stackrel{\text{\bf ``IM"}}{\longrightarrow} &   [\, {\rm XP}-{\rm YP}  \,\,  [_{\rm ZP} \ldots \text{\sout{ YP }} \ldots ] ]
\end{align}
or equivalently, written in the tree notation,
\begin{equation}\label{14b}
{\rm XP} \sqcup \Tree[ .${\rm ZP}$ $\cdots$ [ [ $\cdots$ ${\rm YP}$ ] $\cdots$ ] ]   \,  \stackrel{\text{\bf EM}}{\longrightarrow} \,
\Tree[ ${\rm XP}$ [.${\rm ZP}$ $\cdots$ [ [ $\cdots$ ${\rm YP}$ ] $\cdots$ ] ] ]  \,  \stackrel{\text{\bf ``IM"}}{\longrightarrow}  \, 
\Tree[ [ ${\rm XP}$ ${\rm YP}$ ]  [.${\rm ZP}$ $\cdots$ [ [ $\cdots$ \text{\sout{ ${\rm YP}$ }} ] $\cdots$ ] ] ]  
\end{equation}

Derivations of this type have been offered in the literature, particularly in 
the domain of operator-variable phenomena. Thus, Higginbotham and May  \cite{Higg} 
propose that, under conditions they specify, unary quantifiers in adjoined position 
may merge at LF to form a single binary quantifier. (\ref{15}.a--c) illustrate 
this ``Quantifier Absorption" process schematically for the multiple {\em wh}-question {\em Who read what?}
\begin{equation}\label{15}
\includegraphics[scale=0.55]{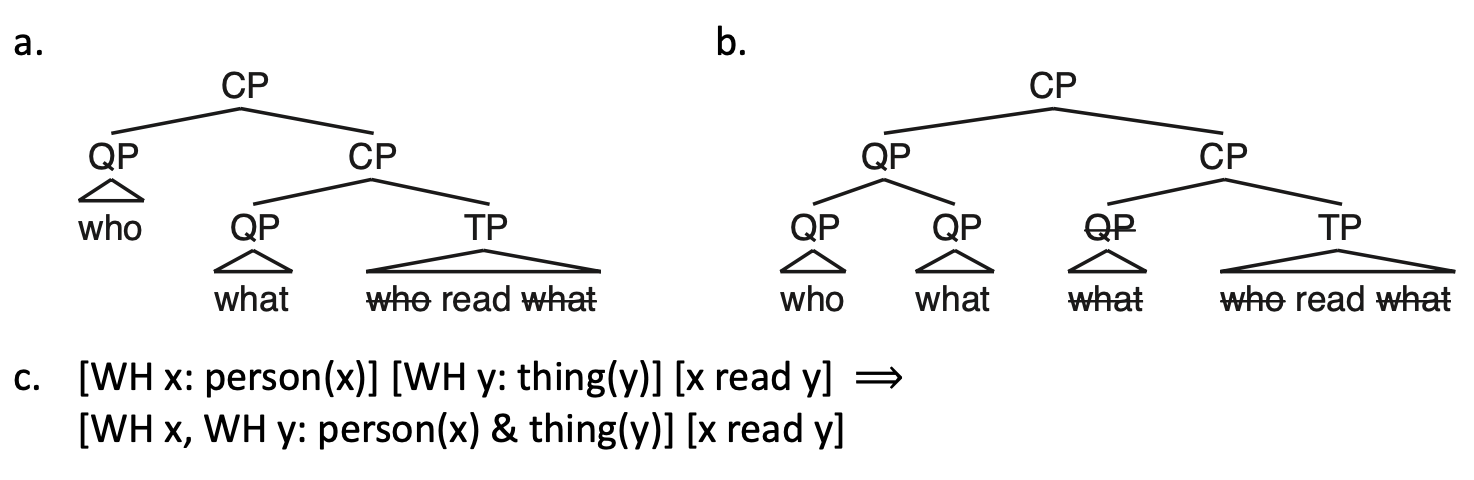}
\end{equation}

Relatedly, Rudin \cite{Rudin} proposes that in certain languages exhibiting multiple {\em wh}-fronting, 
including Bulgarian and Romanian, all {\em wh}’s reside within a single Spec position, 
attached to each other: \eqref{16} illustrates a derivation for the Bulgarian sentence {\em Koj 
vi\v{z}da kogo} `who saw what'.
\begin{equation}\label{16}
\includegraphics[scale=0.55]{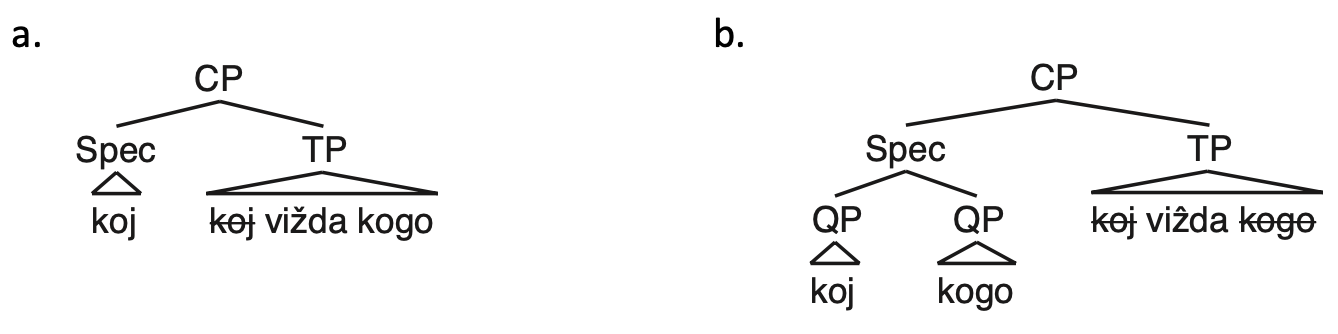}
\end{equation}

Presumably the step from (\ref{16}.a) to (\ref{16}.b) involves 
adjunction to the {\em wh} already present. If so then both 
(\ref{15}.a--b) and (\ref{16}.a--b) would be instances of \eqref{14a}--\eqref{14b}.

\smallskip

We will return to discuss this case in  \S \ref{ECsec}, \S \ref{ExtSec}, \S \ref{MYSec}.
We start in \S \ref{AltSMsec} by presenting alternative derivations for all of these
cases, that use Sideward Merge and do {\em not} involve any EC violation.
We will then analyze these SM solutions in terms of cost functions and violation
of optimality and identify cases for which SM does indeed look like the
correct explanations and those cases where other explanations involving
neither EC-violation nor large optimality violations are possible.

\smallskip
\section{Alternative Derivation using Sideward Merge} \label{AltSMsec}

We see, then, that all instances of  \eqref{5a}--\eqref{5c}, which are uniformly 
proscribed by the Extension Condition, have been proposed in analyses of 
empirical phenomena in the literature, including instances like head-to-head 
movement that have been very widely advocated indeed. In view of this, it is 
worth considering whether a derivational alternative to \eqref{5a}--\eqref{5c} 
is available, which would secure its results while avoiding its flaws. 

\smallskip

Observe \eqref{17a}--\eqref{17c} in this context, a generalization of a proposal by 
Bobaljik and Brown \cite{Boba}, which we place side by side with \eqref{5a}--\eqref{5c} 
for comparison, and where we ignore the status of $X$ and $Y$ as heads or phrases:
\begin{align}
X, \,\,   [ \ldots Y \ldots ]  &    &       X, \,\,   [ \ldots Y \ldots ] & \label{17a}  \\
 X-Y, \,\,  [ \ldots \text{\sout{$Y$}} \ldots ]   & \ \  {\rm {\bf SM}}      &
  [\, X \,\,  [ \ldots Y \ldots ] ] & \ \  {\rm EM}   \label{17b} \\
[\, X-Y \,\,  [ \ldots \text{\sout{$Y$}} \ldots ] ]  & \ \  {\rm EM}   &   [\, X-Y  \,\,  [ \ldots \text{\sout{$Y$}} \ldots ] ]   & \ \  {\rm ``IM"}  \label{17c}
 \end{align}
 In the tree notation the new chain of derivations takes the form
\begin{equation}\label{17all}
X \sqcup \Tree[ $\cdots$ [ [ $\cdots$ $Y$ ] $\cdots$ ] ]  \ \stackrel{\text{\bf SM}}{\longrightarrow} \ 
\Tree[ $X$ $Y$ ] \sqcup  \Tree[ $\cdots$ [ [ $\cdots$ \text{\sout{ $Y$ }} ] $\cdots$ ] ] \ \stackrel{\text{\bf EM}}{\longrightarrow} \
\Tree[ [ $X$ $Y$ ]  [ $\cdots$ [ [ $\cdots$ \text{\sout{ $Y$ }} ] $\cdots$ ] ] ] 
\end{equation}

Note that in its final step \eqref{17a}--\eqref{17c} (shown in the left column) obtains the same structural result 
as \eqref{5a}--\eqref{5c} (shown in the right column of \eqref{17a}--\eqref{17c}), although by EM versus a
putative EC-violating ``IM". 
Crucially different, however, is the intermediate \eqref{17b}-step. Whereas \eqref{5b} employs EM, \eqref{17b} 
uses sideways merge (SM). Importantly, no step in \eqref{17a}--\eqref{17c} violates the 
Extension Condition. The effects of phrase-internal movement are therefore captured, and violation 
of EC evaded, by appeal to sideway merge. This result applies uniformly to all of the cases surveyed 
above where the status of $X$ and $Y$ as head vs. phrases are alternated.

\smallskip
\subsection{Comment on cost counting in the Head-to-Head cases}\label{SMhead2headSec}

In the case discussed in \eqref{13}, note that we can consider as possible derivations
\begin{enumerate}
\item SM movements $v^*$-$V$ followed by an EM
\begin{equation}\label{lookupTree1}
 v^* \,\, \sqcup \, \Tree[  [ look  up ] [ the answer ] ]   \stackrel{\text{\bf SM}}{\longmapsto}  \Tree[ $v^*$ look ] \sqcup \Tree[  [ \text{\sout{ look }} up ] [ the answer ] ]   
 \end{equation}
 $$  \stackrel{\text{\bf EM}}{\longmapsto} \Tree[  [ $v^*$ look ] [  [ \text{\sout{ look }} up ] [ the answer ] ] ] = \Tree[ [ $v^*$ look ] [ [ the answer ] [ \text{\sout{ look }} up ] ] ]  $$
where the last equality holds because trees are non-planar. In this form this is  
a head-to-head movement, that extracts a single atomic element and SM-merges is to another single atomic element.
\item One can also consider a possible derivation with phrase-to-head as mentioned in (\ref{13}.c) that SM-merges
a V-Prt phrase to $v^*$ in the form 
\begin{equation}\label{lookupTree2}
 v^* \,\, \sqcup \, \Tree[  [ look  up ] [ the answer ] ]   \stackrel{\text{\bf SM}}{\longmapsto} \Tree [ $v^*$  [ look  up ] ] \sqcup \Tree[ 
\text{\sout{look up}} [ the answer ] ] 
\end{equation}
$$ \stackrel{\text{\bf EM}}{\longmapsto} \Tree[ [ $v^*$  [ look  up ] ]   [ 
\text{\sout{look up}} [ the answer ] ]     ]  $$
However, one expects this to be disfavored over the previous case as less optimal (we will show
in \S \ref{MYSec} that an explicit computation of costs confirms this).
\item A derivation incorporating an additional IM would not change the result 
and just lengthen the derivation hence it is also be disfavored, 
\begin{equation}\label{lookupTree3}
 \Tree[  [  $v^*$ look ] [ [ the answer ] [  [ \text{\sout{ look }} up ] \text{\sout{the answer}}  ] ] ] 
\end{equation} 
\end{enumerate}

We will explicitly compare cost functions in \S \ref{MYSec} and argue that 
a ranking of optimality violations selects between possible options.

\smallskip

Note, however, that if we replace ``{\em the answer}" in the examples above with a 
larger structure such as, for example, ``{\em the answer to the problem that I gave 
as homework to the student in my class last week}", then ``{\em look up the answer to...}"
would be fine but ``{\em look the answer to... up}" would not. One can argue that this
difference is simply a parsing problem outside of I-language (an Externalization problem).
On the other hand, it is clear that the difference reflects some form of cost counting,
so one might as well deal with all cost counting problems in a single step in the model.
If this is indeed the case, then this would require that an increased cost associated
to the SM-movement of ``{\em look up}" in (2) over the movement of ``{\em look}" in (1) should 
become less significant (out of a total counting of costs) in the case of a ``heavier" NP like ``{\em the answer to...}". 

\smallskip

One can further compare the case discussed here with a similar case of the form
\begin{align*}
\text{\bf John looked the answer right up.} \\
\text{\bf *John looked right up the answer.}
\end{align*}

Note that, even in this case, if one replaces ``{\em the answer}" with
a heavier NP as above, ``{\em the answer to...}", then the case ``{\em John looked the answer to... right up}"
would also incur in the same problem mentioned above, while ``{\em John looked right up the answer to...}"
may seem somewhat ``better" than ``{\em John looked right up the answer}". 

It is also helpful to consider the difference between the non-viable ``{\em *looked right up the answer}" with 
the prepositional case of the viable  ``{\em crawled right up his sleeve}", and the difference 
between an ungrammatical   ``{\em *crawled it right up}" versus a viable ``{\em looked it right up}".

\smallskip

Comparison between these cases suggests the possibility of interaction between
the Sideward Merge movement and another mechanism, the flexible boundary of
Morphosyntax discussed in \cite{SentMar}. We will return to discuss this briefly in \S \ref{MSboundarySec}.

\smallskip
\subsection{Sideward Merge and cost counting}\label{preSMcostsSec}

The existence of an alternative derivation like \eqref{17all}, employing an alternative 
form of merge operation, is technically interesting. But it also raises questions that, in earlier versions 
of the Minimalist Program, would have been difficult to answer confidently. External and Internal Merge 
are familiar operations in the history of generative grammar, their applications being very well-explored 
in the literature. Sideways Merge is, by contrast, a relatively recent innovation, \cite{Nunes}. 
The results above suggest that we can derive results by SM without violating constraints that we 
cannot derive by IM/EM without violating constraints. Does this mean that SM is a more 
powerful operation, whose admission extends the generative capacity of the grammar in undesirable ways? 
Relatedly, is there a sense in which EM and IM are ``better," ``more preferable" operations than SM? 
And if they are ``better," how much better, and according to what properties? Are all instances 
of EM and IM preferable, by these measures, to all instances of SM? Are some instances of 
SM less preferable than other instances of SM?

\smallskip

In Chomsky's new formulation of Minimalism, IM and EM are indeed always preferable to SM,
as they satisfy both an optimization condition with respect to Minimal Search (see \S 3.3.2 of
\cite{ChomskyElements}) and Resource Restriction constraints or Minimal Yield (see \S 4 of
\cite{ChomskyElements}).  However, the question of whether different forms of SM may
have different levels of acceptability (or different degrees of violation of the 
optimality constraints that IM and EM satisfy) is not directly addressed. 

\smallskip

With the recent development of a fully formal version of the Minimalist Program, with the 
mathematical formulation of \cite{MCB}, it has become possible to phrase these questions, 
and evaluate their answers in a precise quantitative way, by
providing a specific cost function with respect to which Minimal Search (as formulated
in \cite{ChomskyElements}, or in a slightly different form here) is indeed minimal, and by a precise quantification of the
Resource Restriction constraints. In the following sections, we recall 
the mathematical model of Merge and various constraints on it, including Minimal Search (MS), 
Minimal Yield (MY) and No Complexity Loss (NCL), under which EM and IM do indeed emerge 
as optimal satisfiers. We can then pose the question of proposed SM derivations of the sort reviewed above, 
demonstrating that their divergence from optimality is not uniform: that some kinds of SM 
operations are less optimal than others. We show that these constraints effectively rank 
the analyses in \S \ref{HtoHsec}--\ref{PtoPsec} and degree of deviance, and briefly 
review the expectations of this ranking, typologically.  The conclusion of this
analysis, as we discuss in \S \ref{MYSec}, is that, beyond the first case (especially in the form that
describes head-to-head movement) all other cases of proposed SM-based
derivations allow at least one form of arbitrarily large
violation, making the SM explanation unlikely.  These other cases, we argue, also
admit alternative explanations that do not involve EC-violations and also do not
require large optimality violations through SM.

\smallskip

Before discussing in \S \ref{CostsSec} and \S \ref{MYSec} the details
of how one analyzes these RR-optimality violations of different forms of SM and correspondingly
ranks the SM operations involved in the different cases, 
we present in \S \ref{ExtSec}  a proposal for
alternative explanations for the cases for which we will show that SM incurs in large optimality
violations. This will confirm the fact that only SM with violations that are very close to optimality plays
a role and that large optimality violations can be replaced by other explanations, not requiring
Merge transformations that are``far from equilibrium" (hence unlikely). 
We will also highlight in \S \ref{ExtSec}  what additional structures need to be incorporated in the
mathematical model for these alternate explanation. We will return to discussing those in \S \ref{ThSec}.

\medskip
\section{Alternative Derivation without Sideward Merge}\label{ExtSec} 

 We present here alternative accounts for some of the phenomena presented above
 that do not involve the use of Sideward Merge with large optimality violations and that 
 still do not involve any EC-violations. We conclude from this analysis that the
 head-to-head case is the only one for which an SM operation may be required, while
 all the other cases may be accountable for otherwise. 
 
 \smallskip
 
 There are in particular two cases that are well known to pose serious 
 empirical problems. One is collecting verbal root and affixes to a single 
 complex verb (English {\em drinks}, French {\em mangeais}, Latin {\em audiveram}). 
 The second is {\em multiple wh-movement} MWM in (South-)Slavic languages. 
 These are different problems, that we discuss in \S \ref{amalgSec} and in \S \ref{MWMsec}. 
 
 \smallskip
 \subsection{Head-to-Head movement: the Amalgamation problem}  \label{amalgSec}

The problem of how to account for head-to-head movement 
under SMT-based Merge Theory without introducing either 
EC-violations or RR-optimality violations
is referred to as {\em Amalgamation} in Chomsky's \cite{ChomskyGK} 
and later work. There it is proposed that Amalgamation is {\em not} a
Merge operation and that it probably belongs to Externalization.

\smallskip

As already mentioned in \S \ref{ECsec}, Matushansky’s 
original proposal \cite{Matu} to account for this phenomenon
violates the Extension Condition, hence it is ruled out as
not structurally formulable as a Merge operation within the 
SMT-based Merge theory,  in Chomsky's formulation 
presented in \cite{ChomskyElements}. Indeed this original proposal 
involves raising a head, say $V$, by IM to yield, e.g., $\{V, \{T, \{EA, \{V, \ldots \} \} \} \}$, 
followed by an additional rule, referred to as $m$-Merge, ``amalgamating" $V$ and $T$.
This $m$-Merge violates EC hence it is not formulable in the current Merge theory.

\smallskip

As we will discuss in \S \ref{ECdualHsec}, in the mathematical formulation 
in \cite{MCB} such EC-violating $m$-Merge operations as proposed in \cite{Matu} 
(as illustrated in \eqref{5c} above) is ruled out for a different
reason, related to its algebraic properties as the insertion 
pre-Lie operation in the dual Lie algebra of the Hopf algebra of workspaces.

\smallskip

Furthermore, a proposal that $m$-Merge is in Externalization 
leads to inconsistency since $m$-Merge both follows and precedes IM. 
Consider, for instance, a structure like $$\{ \{ \{ V, v^* \}, T \}, \{ EA, \{ \{ V, v^* \}, \{V, IA \} \} \} \}.$$
Here, $V$ internally merged with $v^*P$ must then be $m$-Merged with $v^*$ to form a 
complex verb in Externalization, which must itself be raised to $TP$ 
by IM in syntax (I-language) before it can be $m$-Merged with $T$ in Externalization 
again in order to ultimately yield $\{ \{ V, v^* \}, T \}$. 

\smallskip

The conclusion, therefore, is that $m$-Merge is ruled out in I-language (due to EC-violations) 
and yields an incoherent result in Externalization. Therefore, in \cite{ChomskyGK}, Chomsky 
suggests ``Amalgamation" in Externalization that does not require prior application 
of head raising in I-language. Defined in this way, ``Amalgamation” is a term for a set of 
problems that must be solved in Externalization.    

\smallskip
\subsubsection{Amalgamation in SMT-based Merge theory} \label{AmalgOperadSec}

 There are two different possible accounts for the Amalgamation-problems within 
 SMT-based Merge theory, that avoid both EC-violations and RR-optimality violations
 (hence in particular that do not require the use of SM), that differ in where these 
 problems are fixed, inside I-Language or outside I-language in Externalization. 
 Consider the example in \eqref{Am1eq}, with (a) externalized from (b) and with 
 a more specified form of (a) given in (c). 
  \begin{equation}\label{Am1eq}
 \includegraphics[scale=0.55]{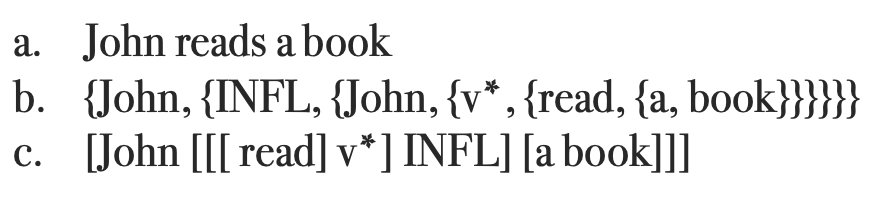}
 \end{equation}
 
 \smallskip
 
\noindent {\bf Hypothesis A}:  Amalgam-problem is part of I-language and is {\em phase-based}. 
Briefly, there’s no SM involved but instead a combination of External Merge (EM) and FormCopy (FC), 
where the latter is not really a principle but a local CC-configuration of identical inscriptions.

\smallskip

With this hypothesis, everything required is formulated within the mathematical
model, where FC takes the form of a restriction to diagonals as discussed in \S 3.8 of \cite{MCB}.

\smallskip

Illustrated in the example of \eqref{Am1eq}, consider a workspace with SO’s that have resulted from previous Merge-operations, written in the form
$$ \text{WS1} = \left[ \text{John},  \{ v^*, \text{read} \},   \{ \text{INFL}, \{ v^*, \text{read} \} \}, \{ \text{read}, \{ \text{a}, \text{book} \} \} \right]  $$
in the notation of \cite{ChomskyElements} or equivalently as
$$ F = \text{John} \, \sqcup \, \Tree[ $v^*$ \text{read} ]\, \sqcup \, \Tree[ INFL [ $v^*$ read ] ] \, \sqcup\,
\Tree[ read [ a  book ] ]   $$
in the notation of \cite{MCB}, where all the trees are {\em non-planar}. 
In particular, inscriptions $\{ v^*, \text{read} \}$ and $\{ \text{INFL}, \{ v^*, \text{read} \}\}$ 
may be EM-results of previous Merge operations applying to workspaces. 
This is fine since WSs are multi-sets of syntactic objects (or equivalently forests in 
the \cite{MCB} formulation). Alternatively, the inscriptions may derive from Merge-based 
lexical processes that generate inflectional morphology on a par with 
derivational morphology. Successive Merge-operations will then yield:
\begin{equation}\label{Am2eq}
\includegraphics[scale=0.55]{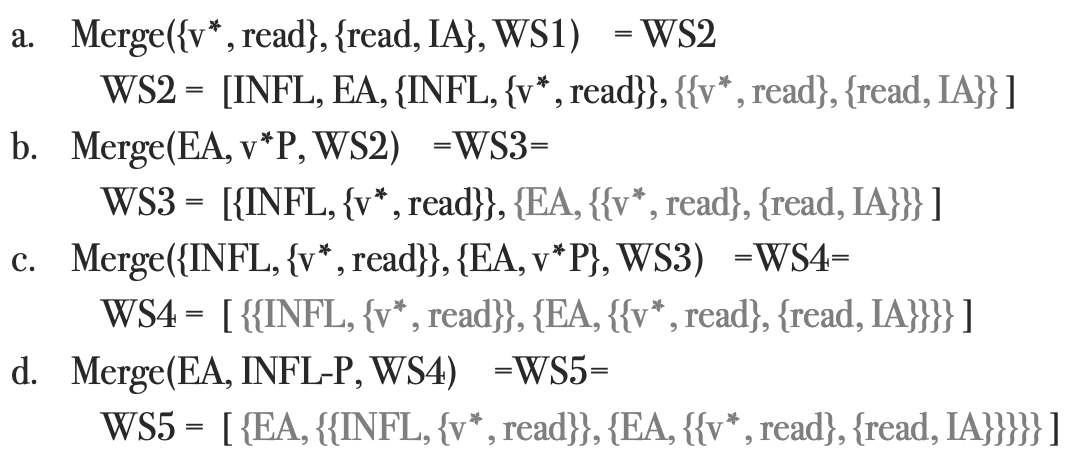}
\end{equation}

Here EM applies in (a-c) and IM applies in (d). The roots {\em read1} and {\em read2} in $$\{ \{ v^*, read2 \}, \{ read1, IA \} \}$$ 
of WS2 are copy-paired under FC (this needs a slight reformulation of $c$-command). The situation is analogous for $\{ v^*, read2 \}$ and $\{ v^*, read3 \}$ in the structure $$\{ \{ \text{INFL}, \{ v^*, read3 \} \},$$ $$\{ \text{EA}, \{ \{ v^*, read2 \}, \text{VP} \} \} \}$$ of WS4. Copy deletion, a computational efficiency principle of the sensorimotor (SM) systems  
(formulated in the mathematical model as the difference between the use of the $\Delta^c$ coproduct at the CI interface
and the $\Delta^d$ coproduct at the SM interface) then yields 
$$ \{ \text{EA}, \{ \{ \text{INFL}, \{ v^*, \text{read} \} \}, \{  \text{\sout{EA}} , 
 \{  \text{\sout{$\{ v^*, \text{read} \}$}}  , \{ \text{\sout{read}} , \text{IA} \} \} \} \} \} , $$ 
i.e, the structure $\{ \text{EA}, \{ \{ \text{INFL}, \{ v^*, \text{read} \} \}, \text{IA} \} \} \}$, 
which is ultimately externalized in the form {\em John [[ read ] -s] a book} as in \eqref{Am1eq}.

\smallskip
 
Here we only used freely applying the basic forms of Merge (EM/IM) and FC. Everything 
conforms to SMT-based Merge Theory as in \cite{ChomskyElements} and is accounted for in 
the mathematical model of \cite{MCB}. No structural principle like EC is violated, and we can also
do without the use of the MS and RR-optimality violating SM.   
Here Amalgamation is just the name of a problem set that can be solved under Merge with optimality 
constraints and is not introduced as a separate rule.  This formulation fits with the setting of \cite{MCB}
using the ``restriction to diagonals" and coproduct extraction according to this restriction, as discussed
in \S 3.8.2 of \cite{MCB}. The only issue that needs to be discussed, to which we will
return in \S \ref{ThSec}, is a quantitative comparison of cost functions, because the use of
only IM/EM eliminates RR-optimality violations of SM, but the FormCopy operation also should 
have an assigned computable cost, so the question of comparing this explanation with
the one based on SM also requires comparing the cost of the type of SM 
involved to the cost of the combination of the FC used here.

Also, among the cases mentioned in \S \ref{HtoHsec}, we should distinguish between
cases like the German ``verb second" V2 and the V-to-T cases. The V2 case can be accounted
for as a standard case of IM, merging INFL to TP: INFL raising to INFL-P by IM defines a root clause and aborts the derivation. Only a single further IM to the newly IM-derived root phase is possible, and when it applies it accounts for the 
Continental Germanic V2 root clauses. Thus, the V-to-T cases are genuine cases of head-to-head, but the T-to-C
are not. 

\smallskip

In terms of the mathematical model of \cite{MCB}, all the constraints on movement by IM based on the
structure of {\em phases}, as in the case discussed here, are implementable through a coloring algorithm 
formulated as a colored operad in \cite{MHL}.  An analysis of the costs of the derivation via FormCopy and
cancellation of the deeper copies compared to the costs of the Sideward Merge derivation is discussed in \S \ref{FCsec}.

\medskip

\noindent {\bf Hypothesis B}: Amalgam-problem is part of Externalization and is dealt 
with at the phase level. In particular, in this proposal 
Amalgamation applies in Externalization and is a rule 
$R(X,Y)$ that substitutes $Y$ for head of  $X$.  

\smallskip

Briefly, inflectional morphology is Merge-based and applies in 
the Lexicon. Again no SM is needed. Notice that substitution
of $Y$ for head of $X$ would violate the EC condition because
it is an operad-type insertion at a leaf that would grow the
tree away from the root. However, here this is assumed to
take place in Externalization and not in the structure formation
process of I-language. We discuss in \S \ref{ThSec} below
what kind of additional structures of the theoretical and
mathematical model would be required to accommodate the 
presence of such substitution rules in Externalization.

\smallskip

We again illustrate this with the example of \eqref{Am1eq}. Here we start with a 
workspace that consists of lexical items and syntactic features (a forest with no edges, 
consisting of single labelled leaves),  WS1 = [ INFL, $v^*$, John, read, a, book ], in
the notation of \cite{ChomskyElements}. Let  $\{ v^*, \text{read} \}$ and 
$\{ \text{INFL}, \{ v^*, \text{read}\}\}$ be merge-generated morphological 
structures of the Lexicon.

Merge-derivation yields I-language syntactic object
$$ SO = \{ \text{EA}, \{ \text{INFL}, \{ \text{EA} , \{ v^*, \{ \text{read}, \{ \text{a, book} \} \} \} \} \} \} $$
To illustrate the substitution rule in Externalization that produces Amalgamation
consider the following setting:
\begin{equation}\label{Am3eq}
\includegraphics[scale=0.55]{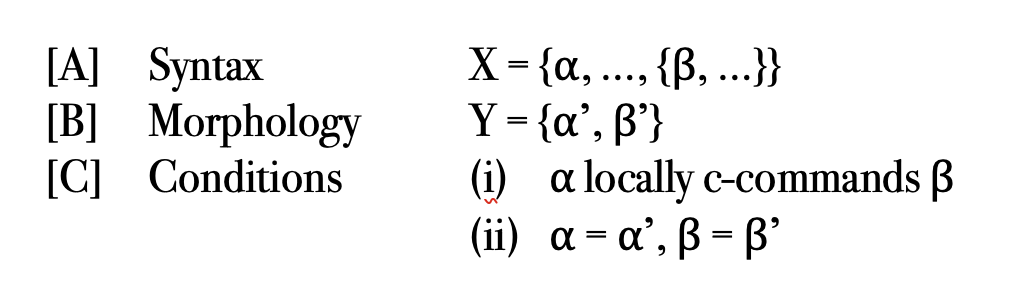}
\end{equation}
Here the local c-command relation is defined as follows.

\begin{defn}\label{locccom}
Head X locally c-commands head Y if X c-commands Y and there’s 
no head Z (Z $\neq$ X, Y) that is c-commanded by X and c-commands Y.
\end{defn}

Then the Amalgamation substitution rule in Externalization is defined in the following way.

\begin{defn}\label{AmalgRule}
In the setting of \eqref{Am3eq}, 
${\rm Amalg}(X, \{ \alpha', \beta' \})$ substitutes lexical morphology $\{\alpha', \beta' \}$ for $\alpha$, 
the syntactic head of X. 
\end{defn}

We will return to discussing this definition and its implications in \S \ref{ThSec} below. 

\smallskip

When applied again to the example of \eqref{Am1eq}, we obtain the following:
\begin{equation}\label{Am4eq}
\includegraphics[scale=0.55]{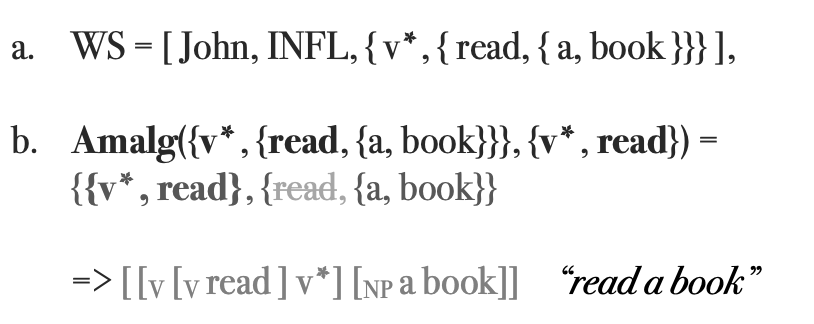}
\end{equation}
\begin{equation}\label{Am5eq}
\includegraphics[scale=0.55]{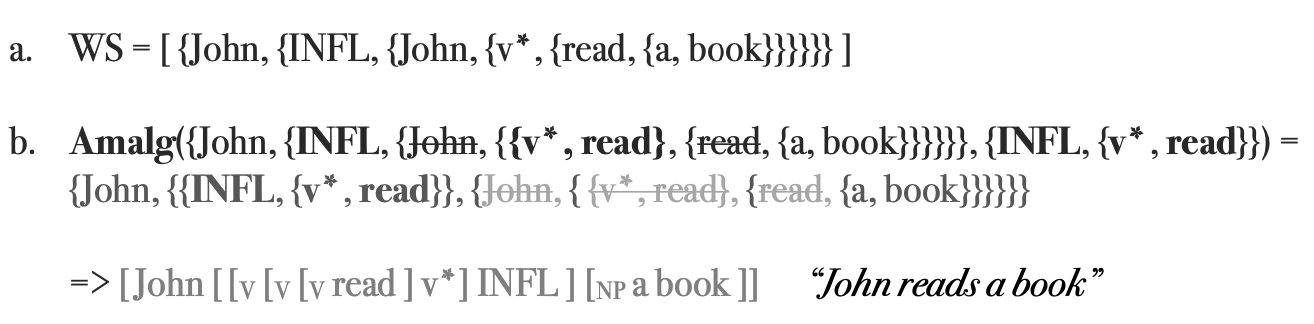}
\end{equation}
where in \eqref{Am4eq}
Amalgamation at the $v^*P$ phase level yields (b) from (a). Analogously, in \eqref{Am5eq}  
Amalgamation at the $CP$ phase level yields (b) from (a). 
The merge-generated lexical structure $\{ \text{INFL}, \{ v^*, \text{Root} \} \}$ gets linearized/externalized as
\begin{equation}\label{Am6eq}
\includegraphics[scale=0.55]{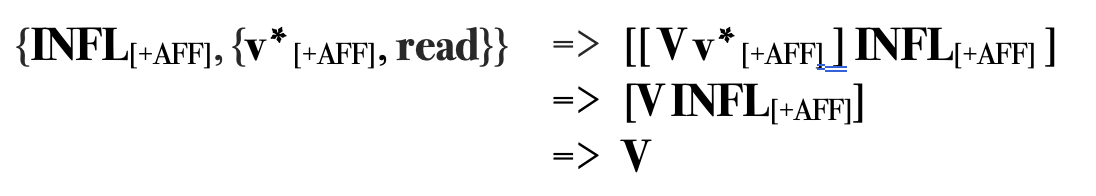}
\end{equation}

The Amalg-operation takes a merge-generated phase structure and a lexical-derived category C,
 substituting C for the head of the phase under some homomorphism.  The merge-generated 
 structure fed into the CI interface has only simple verbal heads, V, $v^*$, and INFL. The 
 amalgamated structure in Externalization has the complex verbs, [V $v^*$] and [[V $v^*$] INFL]. 
 The complexity in Externalization has no semantic effect in CI. 

In contrast, head-to-head movement like V-to-T (of the type that introduces violations RR-optimality
or of EC) would predict that [V T] = T whereas in fact it is [V T] = V 
without any semantic effects. Under this proposal, Amalgamation in Externalization yields complex 
morphology without semantic effects.

The alternative, $EM( \{ V,T \}, \{ EA, \{ V, IA \} \} )$, has copy-paired Vs: one has theta structure properties, 
the other has a morphological function, hosting an affix T. Again one has that [V T] = V and EM-ing $\{ V,T \}$ 
with $\{ EA, v^* P \}$ adds informational structure (tense/aspect/mood) for further interpretation at CI. 

\smallskip

This second possibility also avoids the violation of either structural EC constraints or 
RR-optimality constraints (at the level of the structure formation in I-language), 
but this time at the cost of introducing certain substitution
rules like Amalg of Definition~\ref{AmalgRule} acting on Externalization. As we will
discuss in \S \ref{ThSec} below, this requires a significant extension of the
theory (in terms of modeling Externalization) at the level of its algebraic structure.
The kind of ``operadic insertions" described by the formulation of Amalgamation in 
Definition~\ref{AmalgRule} appears to be compatible with the formulation
of the syntax-morphology interface in \cite{SentMar}, and again the main
question is an evaluation of costs associated to all the operations involved
and a comparison of those costs to the minimal RR-optimality violation introduced
by the use of an SM-based formulation. We will return to discuss this in \S \ref{ThSec}. 

\medskip

\subsection{Head-to-Phrase}\label{XtoXPnoSMsec}

The case of the English genitive morpheme {\em --'s} discussed in \S \ref{HtoPsec} can be approached
with a different analysis as Case Theory, assigned at Externalization. We will comment on this briefly
in the context of the interface of syntax and morphology in \S \ref{MSboundarySec}. On the other hand
a similar case that can give rise to further example of SM-movement is given by the Romance clitics.
These, however, are forms of head-to-head raising, not of head-to-phrase.

\medskip

\subsection{Phrase-to-Head} \label{XPtoXPnoSMsec}

In \S \ref{PtoHsec} we considered the case of phrases like ``{\em look up the answer}" or
``{\em look the answer up}" as possible sources of phrase-to-head movement. 
We discussed briefly in \S \ref{SMhead2headSec} how a modeling of such phrases using
SM should also incorporate a mechanism that accounts for seemingly different costs 
associated to the type of NP involved.
In the same perspective, we can observe that the two cases \eqref{lookupTree1}
and \eqref{lookupTree2} in \S \ref{SMhead2headSec} correspond to
movement starting with 
$$ \{ v^*, \{ {\rm NP}, \{\{ V, {\rm Prt} \}, {\rm NP} \} \} \} $$ 
that give, respectively, either movement of $V$ or of $\{ V, {\rm Prt}\}$.
The verb {\em look up} is $X = \{ {\rm RT}, {\rm Prt} \}$ and either  the 
root ${\rm RT}$ or $X$ is merged via SM to $v^*$, resulting in the case of \eqref{lookupTree1},
$$ \{\{ v^*, V \}, \{ {\rm NP}, \{ \{ V, {\rm Prt} \}, {\rm NP} \} \} \} = \text{ look the answer up } $$
or the case of \eqref{lookupTree2} 
$$ \{ v^*, \{ V, {\rm Prt} \} \}, \{ {\rm NP}, \{ \{ V, {\rm Prt} \}, {\rm NP} \} \} \} = \text{ look up the answer } $$
While the second case looks like a
phrase-to-head movement, because it moves a structure $X = \{ {\rm RT}, {\rm Prt} \}$, 
it should really be regarded, just like the first case, as a case of head-to-head movement.

\medskip
 \subsection{Phrase-to-Phrase and multiple wh-movement} \label{MWMsec}
 
{\em Multiple wh-movement} (MWM) per se 
can be thought of as just a sequence of Internal Merge (IM) applications. 
On the other hand, when one takes into consideration the compatibility
of IM movement with phases, IM either moves to the (unique)
Spec-of-Phase position, or within the interior of a phase to a Spec-of-INFL
or a Spec-of-Root position.  

\smallskip

The proposed ``Quantifier Absorption” (Higginbotham \& May \cite{Higg}), 
as described in \S \ref{PtoPsec} above, cannot be a structure-building 
 operation of I-language if it is an operation at all, since it violates EC. In the alternative formulation 
 discussed in \S \ref{AltSMsec} using SM it does not violate EC but it requires a violation of RR-optimality. 
 Moreover, it has problems with the coloring compatibility for the SM+EM combination
 (versus the IM of a single wh-movement), in terms of coloring by theta positions, see \cite{MHL}, \cite{MarLar}.
 
 \smallskip
 
 The wh-elements do not constitute a constituent, [koj kakvo]. 
 There is no empirical evidence they do, other than the fact that there is a single Spec-of-C 
 that must host all wh-elements. But the X-bar theoretic notion of specifier of X-bar 
 has no meaning in Merge theory, posing an additional problem: Why multiple wh-movement? 
 Why right-adjunction of ``what" to ``who"? 
 
 \smallskip
 
 On the other hand, as we will return to discuss in \S \ref{MSboundarySec}, the
 quantifier absorption possibility is compatible with the ``movable boundary of
 morphosyntax" as formulated within the mathematical model of Minimalism in
 \cite{SentMar}.
 
 \smallskip
 
Moreover, there is an additional problem (referred to as the ``tucking in" problem discussed by Norvin 
Richards in \cite{NRich}) that receives a principled explanation under 
Chomsky's new Box Theory (see \cite{ChomskyKeio},  \cite{Chomsky23a}, \cite{Chomsky23b}) 
that is not available under the simpler current EM/IM model. 
 
 \smallskip
 
 Multiple wh-movement (MWM) in South-Slavic languages is illustrated in (a) of \eqref{MWM1eq} 
 for Bulgarian. 
 \begin{equation}\label{MWM1eq}
\includegraphics[scale=0.52]{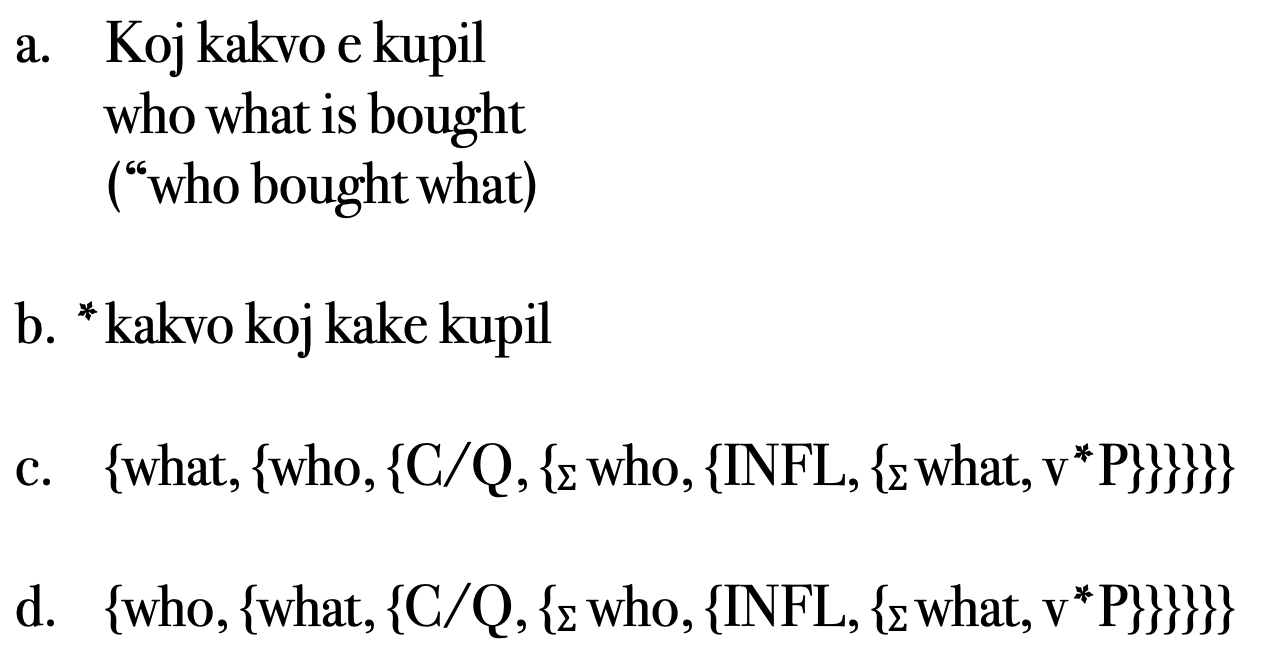}
\end{equation}
 Not only do these languages exhibit multiple wh-movement but some, e.g., Bulgarian, show basic 
 superiority effects. I.e., case (b) of \eqref{MWM1eq}  is excluded. The problem is how to derive 
 superiority under minimality-constrained Merge application. Preserving superiority under Merge 
 implies a ``tucking-in" derivation where a lower wh-element {\em wh2}  is ``tucked-in" under a 
 previously raised wh-element {\em wh1}, violating the Extension Condition. I.e., [wh1 [wh2 [ C … ]]] 
 rather than [wh2 [wh1 [C …]]]. 
 As we have discussed in \S \ref{AltSMsec}, 
 the former result only complies with the Extension Condition if minimality constraint are
 violated. This is a problem since a priori, the type of SM that would account
 for this case can present arbitrarily large violations of RR-minimality, as we show
 in \S \ref{MYSec}. Even if we only restrict to only cases where the resulting violation
 of optimality is as small as possible (resulting in a violation comparable to that of
 an SM-based head-to-head movement), an explanation that avoids optimality violations
 would still be preferable. As in the case of the head-to-head movement discussed above
 we see that for MWM also such an explanation without minimality violations occurs, but
 again it requires significantly extending the algebraic structure of the theory, in this case to
 accommodate the new Box Theory proposed by Chomsky in \cite{ChomskyKeio},  
 \cite{Chomsky23a}, \cite{Chomsky23b}. 
 
 \smallskip
 
 At C/Q phase level, Merge selects {\em who} by MS and raises {\em who} by IM to SPEC-CP. 
 Next, Merge selects {\em what} and internally merges it with the previous merge result. 
 But the result (c) of \eqref{MWM1eq} would be ungrammatical. EA must preserve ``superiority” 
 over IA in wh-question interrogatives. It does not in (c) of \eqref{MWM1eq}. However, the way 
 to achieve this would be to raise {\em what} by IM to SPEC-CP, yielding (d) of \eqref{MWM1eq}. 
 But this ``tucking in" operation violates EC. The SM-based derivation described in \S \ref{AltSMsec}
 above avoids the EC violation but at the cost of violating RR-optimality. 
 
 \smallskip
 
 This is indeed a problem of SMT-based Merge theory. But it’s a problem that is explained 
 under Chomsky's new Box Theory. IM to SPEC-Phase moves an element outside the 
 propositional domain of Merge-derivation. The element gets ``boxed,” or more precisely 
 gets ``immunized” for further Merge application. An immediate consequence is the 
 elimination of successive-cyclicity for A-bar movement. The consequence is strengthened 
 to A-movement as well (omitted here). End of the road for IM-ed elements. Merge theory 
 basically structure-builds theta structure, the propositional domain ($v^*$P, INFL-P). But the 
 propositional domain must be linked to the clausal domain (CP) to carry the computation
 forward. Linking is effected by a convention allowing higher phase heads to access the boxed 
 elements in $\{$ SPEC, Phase $\}$ for interpretation (semantic/CI and 
 spell-out/Externalization). 
 
 \smallskip
 
 This can be illustrated by following \eqref{MWM2eq} below. Merge applications yield (a) of \eqref{MWM2eq}. 
 The boxed elements will be (cyclically) accessed by the matrix phase head C/Q.  
 \begin{equation}\label{MWM2eq}
\includegraphics[scale=0.55]{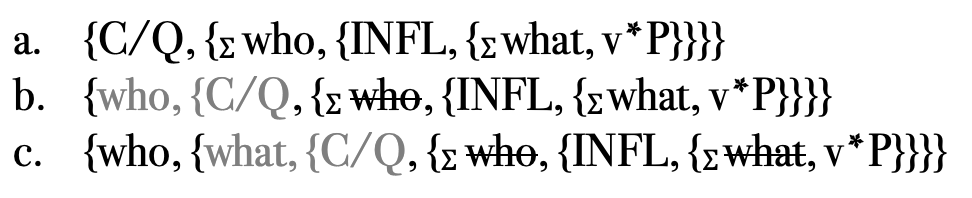}
\end{equation}

\begin{itemize}
\item {\em Step 1}: C/Q accesses {\em who} in $\{$ SPEC, INFL-P $\}$ under MS and both interprets 
(at the Conceptual-Intensional interface) {\em who} and spells out (at the Sensory Motor interface) {\em who} in SPEC-of-Phase. 
See (b) of \eqref{MWM2eq}. This first step of C/Q accessing ``who” translates the Vacuous 
Movement Hypothesis into Merge/Box-Theory. It is slightly more complicated though (see \cite{Chomsky23a}),
as we will discuss more in detail below. 
\item {\em Step 2}: C/Q accesses {\em what} in $\{$ SPEC, $v^*$P $\}$ under MS and both interprets (at the CI
interface) and spells out (at the Sensory Motor interface) what in SPEC-of-Phase. See (c) of \eqref{MWM2eq}. 
\end{itemize} 

\smallskip

There is just one SPEC in a  $\{$ SPEC, Phase $\}$ configuration. XP in $\{ \text{XP}, \{ \text{SPEC}, \text{Phase} \} \}$ is no such configuration since $\{ \text{SPEC}, \text{Phase} \}$ is not a Phase. Note that at the derivational 
stage (a) of \eqref{MWM2eq}, ``who" is accessed and externalized in SPEC-of-C/Q, yielding a 
$\{ \text{SPEC}, \text{Phase} \}$ configuration. When subsequently ``what" is accessed by the phase head C/Q, it must be externalized in 
SPEC-of-C/Q with a unique result $\{$ who, $\{$ what, $\{ \text{C/Q}, \ldots \} \}\}$.  

\smallskip

The result is that ``tucking in" naturally follows and is no longer a problem for Merge. 
Merge unexceptionally conforms to EC and also to RR-optimality. It does not ``counter-cyclically” 
tuck in lower elements underneath higher elements. The multiple wh-movement (MWM) of case 
(c) of \eqref{MWM2eq} is accounted for outside optimal-Merge theory, which only generates (a) 
of \eqref{MWM2eq}, but results from C/Q accessing boxed elements in Box Theory at the phasal level.   

\begin{itemize}
\item Tucking-in results from C/Q accessing $\Sigma = \{ \text{SPEC}, \{ \Phi, \ldots \}\}$ 
constrained by optimality as Minimal Search (MS).
\end{itemize}

The ``tucking-in" property contradicts the applicability of Merge operation (since it entails either a violation of EC
or in the SM-based proposal a violation of RR-optimality) but falls out from Box Theory connecting propositional 
and clausal domains (since C/Q accesses boxed elements under MS). In South-Slavic languages, in particular 
Bulgarian, C/Q has multiple access. In English-type languages C/Q accessing of boxed elements is restricted. 
This parameterized difference is independent of ``tucking-in", which has an independent explanation under 
Box Theory. 

\smallskip

Thus, in the setting described here based on Box Theory, there’s no need for quantifier absorption. 
Accessing of boxed elements in SPEC-of-Phase by matrix C/Q for interpretation at the CI-interface 
(and the SM-interface) suffices.  On the other hand, this relies on a form of Minimalism (Box Theory)
that is not at present realized within the mathematical formalism of \cite{MCB}: while requiring
that elements moved by IM to SPEC-Phase are ``boxed” and inaccessible to further Merge application
can be easily implemented by a small change to the colored operad generators for phases as
formulated in \cite{MHL}, the part that, at the moment, does not have a suitable mathematical formulation
is the meaning of ``C/Q accessing boxed elements". Thus, this possible explanation
based on Box Theory remains a question within this model. 

\smallskip

There is a related question. The Slavic languages themselves fall in two categories:
\begin{equation}\label{MWM3eq}
\includegraphics[scale=0.42]{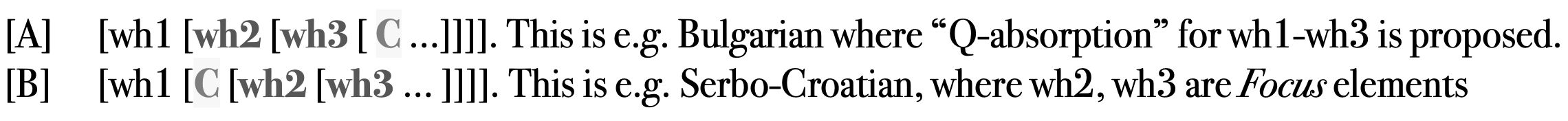}
\end{equation}

\smallskip

Basic Superiority holds for the [A]-type language but is waived for the [B]-type language. 
No theory has an explanatory account yet for the multiple wh-Focused cases in [B]. 
Focus and Scrambling need some looking into. As regards [A]-type languages,
 C/Q accessing both ``who" and ``what" gives the pair-list semantics of \eqref{MWM2eq} 
 on a pair with English ``Who bought what?" The latter case shows C/Q accessing {\em who} 
 first and accessing {\em what} second as in Bulgarian but externalizing only {\em who} 
 in SPEC-of-C/Q and spelling out “what” in situ.
 
 \smallskip
 
 We will revisit in \S \ref{BulgSMsec} the proposed explanation via Sideward Merge
 for the multiple wh-fronting and the tucking in problem. In comparison with the
 alternative explanation proposed here with the use of Box Theory, the SM formulation
 we discuss in \S \ref{BulgSMsec} can be fully implemented within the framework of
 the mathematical model of Minimalism, as it is at the persent stage. 
  
 \smallskip

\section{Mathematical Minimalism} \label{MathMinSec}

As shown in \S 1.5 and \S 1.6 of \cite{MCB}, the constraints of Minimal Search
and Resource Restriction described in \cite{ChomskyElements} can be quantified. For
Minimal Search, quantification corresponds to the introduction of a cost function,
determined by the coproduct of the Hopf algebra of workspaces, with respect to
which Minimal Search is indeed ``minimal" in the sense that it selects External Merge (EM)
and Internal Merge (IM) as the leading order terms, and the remaining forms of
Sideward Merge (SM) as subdominant with respect to EM/IM. Resource Restriction
is articulated in \S 1.6 of \cite{MCB} as two types of constraints: Minimal Yield 
about constraints on different counting of size of workspaces, and a ``no
complexity loss" principle. In the case of Minimal Yield, the quantification is 
in terms of a ``no divergence" condition on (number of components of the
workspace is non-increasing) and a ``no information loss principle" (number of
accessible terms is non-decreasing). The remaining ``no complexity loss"
principle expresses the property that Merge should recursively build 
structures of increasing complexity, where a simple indicator of the ``complexity" 
of the obtained structures is provided by the degree in the Hopf algebra of
workspaces, so degrees of all individual workspace components should be
non-decreasing. It is proved in \S 1.6 of \cite{MCB} that EM, IM, and only
one form of SM meet the ``no divergence" and ``no information loss principle",
and only EM and IM meet the ``no complexity loss" principle.

\smallskip

Given the mathematical model of \cite{MCB}, we can directly compute 
what type of structures realize the ``minimal violations" to the optimality 
constraints, namely when one or more of them are violated in the ``minimal possible way".
Such phenomena can be seen as minimal extensions of the EM/IM
structure building operations via the only other accessible operations in
the mathematical formulation of the action of Merge on workspaces, namely
forms of Sideward Merge. The resulting SM operations will be those that,
although not of leading order and less cost-effective than EM/IM, are still preferable 
to any other form of SM, in representing movement that is ``as close as possible" 
to IM and that violates minimally the Resource Restriction constraints. 

\smallskip

We summarize here the mathematical model of Merge and Minimalism
of \cite{MCB} and we then proceed in \S \ref{MYSec}
to compute the amount of violation of the optimality constraints 
associated to different forms of SM and we compare the results with
the four linguistic cases discussed in \S \ref{ECsec}.

\smallskip
\subsection{Merge model summary}

We recall the fundamental aspects of the mathematical model
of Merge of \cite{MCB} that we will need to use in the following.
\begin{itemize}
\item $\cS\cO_0$ is the (finite) set of {\em lexical items and syntactic features}.
\item $\cS\cO$ is the (countable) set of {\em syntactic objects}, with the algebraic
structure of free commutative nonassociative magma generated by the set $\cS\cO_0$,
with binary operation $\fM$,
$$ \cS\cO = {\rm Magma}_{c,na}(\cS\cO_0, \fM)\cong \fT_{\cS\cO_0}\, , $$
canonically isomorphic to the set of (non-planar) binary rooted trees with
leaves decorated by elements of $\cS\cO_0$.
\item The (countable) set $\fF_{\cS\cO_0}$ of binary rooted forests with
connected components in $\fT_{\cS\cO_0}$ is the set of {\em workspaces}.
\item An {\em accessible terms} $T_v\subset T$ of a syntactic objects is
the subtree of all descendants of a non-root vertex $v$ of $T$.
\item The vector space $\cV(\fF_{\cS\cO_0})$ spanned by the set of workspaces
has the structure of a commutative graded connected Hopf algebra, with product
$\sqcup$ given by the disjoint union of workspaces, and coproduct given by
the extraction of accessible terms,
\begin{equation}\label{coprod}
 \Delta(T)=\sum_{\underline{v}} F_{\underline{v}} \otimes T/F_{\underline{v}} \, , 
\end{equation} 
with $F_{\underline{v}}=T_{v_1}\sqcup\cdots \sqcup T_{v_n}$ a collection of disjoint
accessible terms of $T$ and $T/F_{\underline{v}}$ the resulting quotient (cancellation
of the deeper copy).\footnote{We refer the reader to \S 1.2 of \cite{MCB} for  a
detailed discussion of various subtleties in the definition of the quotient $T/F_{\underline{v}}$,
and for the linguistic role of the different forms of this quotient.}
\item The {\em action of Merge on workspaces} is defined by the collection of linear operator
$\fM_{S,S'}$ on $\cV(\fF_{\cS\cO_0})$, parameterized by pairs $S,S'$ of syntactic objects,
given by
\begin{equation}\label{opMerge}
 \fM_{S,S'} = \sqcup \circ (\fB \otimes {\rm id})\circ \delta_{S,S'} \circ \Delta \, , 
\end{equation} 
where $\Delta$ is the Hopf algebra coproduct, $\delta_{S,S'}$ is the Kronecker-delta
operator that is the identity on terms in the range of $\Delta$ that have $S\sqcup S'$ in
the left-hand-side of the coproduct (and zero otherwise), $\fB$ is the grafting operator
from forests to trees $\fB(T\sqcup T')=\fM(T,T')$, and $\sqcup$ is the coproduct that
reassembles the new workspace at the end of this operation. These operators include
the case where $S$ or $S'$ can be the product unit $1$ (namely the formal empty forest),
as this case is needed in the case of Internal Merge.
\item The operations $\fM_{S,S'}$ and their composites include External and Internal
Merge and three different forms of Sideward Merge.
\item A cost function that weights the terms of the coproduct, by assigning a cost
to both the extraction of an accessible term and its contraction in the quotient (the
cancellation of the deeper copy) identifies External and Internal Merge as the leading
terms with minimal cost, while all the forms of Sideward Merge appear as subdominant
(higher cost) terms. This cost function realizes {\em Minimal Search} as the extraction of
the leading order terms.
\item {\em Resource Restriction} constraints involving the effect of the Merge operators
$\fM_{S,S'}$ on different size measures ({\em Minimal Yield}) on workspaces identify 
External and Internal Merge, as well as 
one of the three forms of Sideward Merge, as optimal. A further Resource Restriction
constraint of ``no complexity loss", determined by the Hopf algebra grading, selects
External and Internal Merge as the only optimal forms.
\end{itemize}

We refer the readers to Chapter~1 of \cite{MCB} for a detailed account of all the
aspects recalled here. We just recall here the three forms of SM that are obtained
as cases of the operators $\fM_{S,S'}$ above.

We will also make use in this paper of some additional developments of the
mathematical model of Minimalism of \cite{MCB}, especially the formulation
in \cite{MarLar} and in \cite{MHL} of theta roles and of the structure of phases
in terms of colored operads and their bud generating systems.

\smallskip

\section{Extension Condition and Hochschild cocycles} \label{ECdualHsec}

In the mathematical formulation of Merge and Minimalism in \cite{MCB},
{\em the Extension Condition is always automatically satisfied}, namely it is
what we referred to here as a ``hard constraint" of the model. The reason
is that all structure formation in this model is effected by two algebraic
structures: the magma of syntactic objects, where the only available operation
(the magma multiplication) acts only at the root of trees, and the
grafting operator that determines Merge, and this
operator is also forced to act only at the root of trees by
its algebraic properties. As we will discuss in this section, the operation
of merging trees at the root (as opposed to EC-violating insertions at
edges inside the tree) is characterized by a specific algebraic
property: being a Hochschild cocycle on the Hopf algebra of workspaces. 
As we explain here, this property has an important consequence on
the semantics interface, which characterizes Merge as an operation
that is ``optimal" (in the category-theoretic sense of satisfying a 
universal property) for interfacing with any model of semantics
that can be endowed with compositional properties. Another way
to say this is that, as shown in \cite{MCB} models of semantics do
not need to be compositional for a viable syntax-semantics interface
to exist (as syntax provides the compositional structure), but also
any model of semantics that is compositional has to proceed from
syntax in a way that we make precise in this section. This
universal property is {\em not} shared by EC-violating forms
of tree-growth like insertions happening lower into the tree. 
Thus, any operation that grows a tree at any other place that is not the root is not
part of the structure formation mechanism in this model because it does not
satisfy the same algebraic properties. 

\smallskip

There is an interesting subtlety, though, as discussed in \S 1.3.2 of \cite{MCB}.
Operations that grow a tree at non-root vertices do exist in this formalism, not directly
as part of the Hopf algebra of workspaces with its grafting cocycle, but in 
the {\em dual} Hopf algebra.

\smallskip

More technically, the operations that insert a tree into another one at one of 
the lower vertices determines the pre-Lie structure of the Lie algebra of 
primitive elements of the dual Hopf algebra. We will not discuss this here
in details, but we refer the reader to \S 1.3.2 of \cite{MCB}.

\smallskip

What is relevant here, though, is that this mathematical fact suggests 
heuristically that proposed constructions that appear to violate the
Extension Condition should (by this duality structure) in fact admit 
an alternate derivation that only uses what is available in the Hopf
algebra of workspaces. This is exactly the case for the constructions
that we analyzed in \S \ref{ECsec}, with the alternate derivation in
terms of Sideward Merge that we presented in \S \ref{AltSMsec}:
SM is indeed fully defined in terms of the Hopf algebra of workspaces 
endowed with its grafting cocycle that defines the Merge operator. 
So we can regard our discussion of \S \ref{AltSMsec} as an instantiation 
of this general principle.

\smallskip

We prove here that the Extension Condition is a structural algebraic necessity of the
model, in as it ensures an algebraic property, the cocycle property,
and that no EC-violating insertion, at locations other than the root, would
satisfy the same condition.

\smallskip

We first recall briefly the mathematical notion of Hochschild cocycles
and Hochschild cohomology for bialgebras and then we analyze how
it relates to EC.

\smallskip
\subsection{Hochschild  cocycles}\label{HochCocycleSec}

In general Hochschild homology is a way of defining
homological invariants of algebras (which dualizes to a
Hochschild cohomology of coalgebras), in such a way
that, when the algebra consists of functions on a smooth space, it
recovers the geometric theory of differential forms on that space.,
and that the lower homology groups captures well known
properties of algebras, such as the existence of trace
functionals and derivations. 

In the mathematics literature one encounters primarily the case of 
an {\em algebra} $A$, where Hochschild homology ${\rm HH}_n(A,M)$ is built 
using a bimodule $M$ and a complex $(\Hom(A^{\otimes n}, M), b)$, 
for which we do not write out here explicitly the differential $b$, since 
we will rather focus here on its dual version in the case of a {\em coalgebra}. 

We recall the basis definition of bicomodules for a coalgebra, and the
definition of Hochschild cohomology.

\begin{defn}\label{bicomoddef}
A bicomodule $M$ for a coalgebra $C$ is both a left and a right
comodule, with a compatibility condition, where the structure of left comodule is given by a linear map
$\rho_L \in \Hom(M, C\otimes M)$ satisfying
$$( {\rm id}_C\otimes \rho_L) \circ \rho_L = (\Delta \otimes {\rm id}_M) \circ \rho_L  \ \ \ \text{ and } \ \ \ 
(\epsilon \otimes {\rm id}_M) \circ \rho_L = {\rm id}_M\, , $$
with $\Delta$ and $\epsilon$ the coproduct and counit of $C$. The right comodule
structure is defined similarly with a linear map $\rho_R \in \Hom(M, M\otimes C)$ staisfying
$$ ( \rho_R \otimes {\rm id}_C) \circ \rho_R = ({\rm id}_M \otimes  \Delta) \circ \rho_R  \ \ \ \text{ and } \ \ \ 
({\rm id}_M \otimes  \epsilon) \circ \rho_R = {\rm id}_M\, . $$
The compatibility of the left/right structures is expressed by the identity 
\begin{equation}\label{compatcomod}
( {\rm id}_C\otimes \rho_R) \circ \rho_L =(\rho_L \otimes {\rm id}_C) \circ \rho_R\, .
\end{equation}
In particular, a coalgebra $C$ is a bicomodule over itself with $\rho_L=\rho_R=\Delta$.
\end{defn}

\smallskip

\begin{defn}\label{HHcoalgdef}
To construct the Hochschild cohomology of a coalgebra $C$, one considers a bicomodule $M$
and a complex $(\Hom(M, C^{\otimes n}), b_n)$, where
$\Hom(M, C^{\otimes n})$ are  linear maps, with differential
\begin{equation}\label{bndiff}
 b_n = \sum_{i=0}^{n+1} (-1)^i b_{n,i}, 
\end{equation}
\begin{equation}\label{bndiff2} 
 b_{n,i}(\psi) = \left\{ \begin{array}{ll} ({\rm id}\otimes \psi)\circ \rho_L & i=0, \\
({\rm id}^{\otimes i-1} \otimes \Delta \otimes {\rm id}^{\otimes n-i}) \circ \psi &  1\leq i \leq n \\
(\psi \otimes {\rm id})\circ \rho_R & i=n+1 . 
\end{array}\right. 
\end{equation}
This is indeed a differential, satisfying $b_{n+1}\circ b_n=0$. Thus, one can compute
the resulting cohomology given by the quotient
$$ {\rm HH}^n(C,M):= \frac{{\rm Ker}(b_n: \Hom(M, C^{\otimes n}) \to \Hom(M, C^{\otimes n+1}))}{ {\rm Im}(b_{n-1}: \Hom(M, C^{\otimes n-1}) \to \Hom(M, C^{\otimes n}))}\, . $$
One refers to the elements in ${\rm Ker}(b_n: \Hom(M, C^{\otimes n}) \to \Hom(M, C^{\otimes n+1}))$, that is,
the linear maps $\psi\in \Hom(M, C^{\otimes n})$ with $b_n(\psi)=0$, as the $n$-Hochschild cocycles, and
to the elements in the image ${\rm Im}(b_{n-1}: \Hom(M, C^{\otimes n-1}) \to \Hom(M, C^{\otimes n}))$ as the
Hochschild coboundaries. 
\end{defn}

\smallskip
\subsubsection{Hochschild 1-cocycles}

In particular, a Hochschild $1$-cocycle $\psi : M \to C$ satisfies $b_1 \psi =0$ (the cocycle condition) with
$b_1$ as in \eqref{bndiff}, \eqref{bndiff2}. This reads as the equation
\begin{equation}\label{b1diff}
b_1\psi=(b_{1,0}-b_{1,1}+b_{1,2})\psi = ({\rm id}\otimes\psi)\circ \rho_L - \Delta \circ \psi + (\psi \otimes {\rm id}) \circ \rho_R = 0
\end{equation}
or equivalently
\begin{equation}\label{b1diff2}
\Delta \circ \psi = ({\rm id}\otimes\psi)\circ \rho_L  + (\psi \otimes {\rm id}) \circ \rho_R \, . 
\end{equation}
where the left-hand-side depends on the bialgebra coproduct $\Delta$ of $C$ and the right-hand-side
depends on the bicomodule structure $\rho_L, \rho_C$ of $M$. Thus, one can view a 
Hochschild $1$-cocycle  as a linear map $\psi : M \to C$ that intertwines in an optimal way
the bicomodule structure on the source with the coalgebra structure on the target. 

The identity \eqref{b1diff2} is an identity between linear maps from $M$ to $C \otimes C$.
For any given element $x\in M$, it gives an identity
$$ \Delta(\psi(x)) = ({\rm id}\otimes\psi)(\rho_L(x))  + (\psi \otimes {\rm id})(\rho_R(x)) $$
between vectors in the vector space $C\otimes C$.

\smallskip
\subsection{Hochschild 1-cocycles for bialgebras}\label{bialgHH1sec}

If instead of a coalgebra we consider a bialgebra $\cH$, then there are different possible ways of constructing
a bicomodule structure on $\cH$ over itself. Since $\cH$ is also a coalgebra, we can take, as above
$\rho_L=\rho_R=\Delta$. However, since $\cH$ also has a unit $1$ for the algebra structure, we can
also use $\rho_L=1\otimes {\rm id}$ and $\rho_R = {\rm id}\otimes 1$. Moreover, one can also
consider the cases with $\rho_L=\Delta$ and $\rho_R = {\rm id}\otimes 1$ or 
$\rho_R=\Delta$ and $\rho_L=1\otimes {\rm id}$. All of these satisfy the compatibility, since the
coproduct is an algebra homomorphism. Thus, for each of these possible combinations, one
can write the resulting Hoschschild $1$-cocycle condition.

\smallskip

\begin{itemize}
\item The first case, $\rho_L=\rho_R=\Delta$, gives as the corresponding the $1$-cocycle equation
$$ \Delta\circ \psi= ({\rm id}\otimes \psi + \psi \otimes {\rm id}) \circ \Delta\, , $$
which is also known as the {\em coderivation} (which is dual to the Leibniz rule
satisfied by a {\em derivation} $D$ on an algebra: $D \circ \mu =\mu \circ (D \otimes {\rm id} + {\rm id} \otimes D)$,
with $\mu$ denoting the algebra multiplication).  

\smallskip

\item The second case,  $\rho_L=1\otimes {\rm id}$ and $\rho_R = {\rm id}\otimes 1$, gives as $1$-cocycle equation
$$  \Delta\circ \psi = 1 \otimes \psi + \psi \otimes 1 \, , $$
which denotes a linear map $\psi: \cH \to \cH$ whose range is contained in the subspace of 
primitive elements for the bialgebra coproduct. 

\smallskip

\item In the remaining two cases, one finds a different, less symmetric, form of the resulting $1$-cocycle equation: 
for the case of $\rho_L=\Delta$ and $\rho_R = {\rm id}\otimes 1$ one finds
\begin{equation}\label{1coeq}
  \Delta\circ \psi = \psi \otimes 1 + ({\rm id}\otimes \psi) \circ \Delta\, , 
\end{equation}  
and for the case of $\rho_R=\Delta$ and $\rho_L=1\otimes {\rm id}$ one has 
\begin{equation}\label{1coeqW}
 \Delta\circ \psi = 1 \otimes \psi + (\psi \otimes {\rm id}) \circ \Delta \, . 
\end{equation} 
\end{itemize}

\smallskip

The case $\rho_L=\Delta$ and $\rho_R = {\rm id}\otimes 1$, with \eqref{1coeq} as the resulting
$1$-cocycle equation, is the one that plays a crucial role in physics, see for instance
the discussion in \cite{Panzer}. The case $\rho_R=\Delta$ and $\rho_L=1\otimes {\rm id}$ with \eqref{1coeqW} 
will also be directly relevant in the linguistic setting, as we will discuss below.

\smallskip
\subsection{The grafting operator as a cocycle}\label{graftBcocycleSec}

As in \cite{MCB}, \S 1.9 and  \S 1.16, we consider a projection $\gamma_{(k)}$ on the
vector space $\cV(\fF_{\cS\cO_0})$ that projects onto the subspace spanned by the
basis elements $F\in \fF_{\cS\cO_0}$ given by forests with $k$ components,
$b_0(F)=k$. In particular, for $k=2$, and $\Pi_{(2)}=\gamma_{(2)}\otimes {\rm id}$,
we obtain as in \S 1.9 of \cite{MCB} the linear operator
$$ \cK = \sum_{S,S'} \fM_{S,S'} = \sqcup \circ (\cB \otimes 1) \circ \Pi_{(2)} \circ \Delta \, , $$
where $\cB$ is the grafting operator
\begin{equation}\label{grafting}
 \cB(T_1 \sqcup \cdots \sqcup T_n)= \Tree[ $T_1$ $T_2$ $\cdots$ $T_N$ ]\, , 
\end{equation} 
and the projection $\Pi_{(2)}$ ensures that it only applies to terms with two components, producing binary trees. 
The operator $\cK$ 
describes as a single operation all the possible applications of Merge, $\cK=\sum_{S,S'} \fM_{S,S'}$, to a given workspace.

\smallskip

In the form \eqref{grafting}, the operator $\cB$ is not defined on $\cV(\fF_{\cS\cO_0})$ as it
generates trees that have a non-binary vertex, whenever applied to forests with more than two
connected components. Indeed, $\cB$ is defined as a linear map on the larger vector space.
We denote here by $\tilde\fF_{\cS\cO_0}$ the set of rooted
forests of arbitrary arity with both leaves and internal vertices 
decorated by elements in $\cS\cO_0$. We can then view $\cB$ as an endomorphism of $\cV(\tilde\fF_{\cS\cO_0})$.
This vector space can be endowed with a Hopf algebra structure, the commutative graded connected
Connes--Kreimer Hopf  algebra $\cH_{CK}=(\cV(\tilde\fF_{\cS\cO_0}), \sqcup, \Delta_{CK})$,
graded by number of leaves, with product given by the disjoint 
union $\sqcup$ and coproduct $\Delta_{CK}$ given by summing over all the admissible cuts in the form
\begin{equation}\label{DeltaCK}
 \Delta_{CK}(T)=\sum_C \pi_C(T) \otimes \rho_C(T) \, ,
\end{equation} 
as in the case of the coproduct $\Delta^\rho$ on the Hopf algebra of workspaces.

\smallskip

The grafting $\cB$ of \eqref{grafting} satisfies the Hochschild $1$-cocycle condition \eqref{1coeq} for this
Connes--Kreimer Hopf algebra $\cH_{CK}$ of rooted trees (of arbitrary arity, not necessarily binary),
\begin{equation}\label{CKcocycle}
 \Delta_{CK} \circ \cB = \cB \otimes 1 + ( {\rm id} \otimes \cB) \circ \Delta_{CK} \, , 
\end{equation} 
with the coproduct $\Delta_{CK}$ given by the admissible cuts, 
see Proposition~28 of \cite{Foissy}. 

\smallskip

We also denote by $\tilde\fF_{\cS\cO_0}^{\leq n}$ the
set of ``at most $n$-ary" forests with leaves and internal vertices labelled by $\cS\cO_0$. 
If $F \in \tilde\fF_{\cS\cO_0}^{\leq n}$, then also 
$\Delta_{CK}(F)\in \cV(\tilde\fF_{\cS\cO_0}^{\leq n})\otimes \cV(\tilde\fF_{\cS\cO_0}^{\leq n})$. 
Since $\tilde\fF_{\cS\cO_0,\Omega}^{\leq n}$ is also preserved under $\sqcup$, we obtain a Hopf subalgebra
$\cH_{CK}^{\leq n}$ of $\cH_{CK}$. We are in particular interested in the case of the Hopf subalgebra
$\cH_{CK}^{\leq 2}$. 

\smallskip

The subspace $\cV(\tilde\fF_{\cS\cO_0}^{\leq 2})$ of binary forests (with non-branching vertices allowed) 
is not preserved by the linear operator
$\cB$. However, the linear operator
$$ \sqcup \circ (\cB \otimes {\rm id}) \circ \Pi_{(2)} \circ \Delta_{CK} $$
does preserve the subspace $\cV(\tilde\fF_{\cS\cO_0}^{\leq 2})$ and agrees with 
the the Merge operator on workspaces (with the $\Delta^\rho$ choice of the coproduct). 

\smallskip

On the larger vector space $\cV(\tilde\fF_{\cS\cO_0})$ one can similarly construct the
Hopf algebra Markov chain for the $\cH_{CK}$ Hopf algebra of the form
\begin{equation}\label{CKMarkov}
\cK_{CK}:= \sqcup \circ (\cB\otimes {\rm id})\circ \Delta_{CK} \, ,
\end{equation}
where we do not need to introduce any projection after $\Delta_{CK}$, since in
$\cV(\tilde\fF_{\cS\cO_0})$ the operator $\cB$ can be applied to any arbitrary
term of the coproduct $\Delta_{CK}$. 
The Merge Hopf algebra Markov chain
$$ \cK= \sqcup \circ (\cB\otimes {\rm id})\circ \Pi_{(2)} \circ \Delta $$
on workspaces consisting of binary forests is the restriction of
the Markov chain \eqref{CKMarkov}.

\smallskip
\subsection{Labelled grafting}\label{labelBSec}  

There is a subtlety here that has to do with the labels. We want to work with trees and forests
that have labels at the leaves, taken from a finite set, here $\cS\cO_0$. In the Connes--Kreimer Hopf 
algebra $\cH_{CK}$ with the grafting $\cB$, the grafting should satisfy $\cB(1)=\bullet$ (a tree with
a single vertex), in order to satisfy \eqref{1coeq}. 
One possibility is just to assume that the set of labels contains an ``empty label" and interpret $\cB(1)=\bullet$ as the
single leaf with empty label. However, this is not compatible with the linguistic interpretation of
the set of labels $\cS\cO_0$, which should be lexical items and syntactic features. 

A better solution is to consider, as in the physics literature, a 
family of grafting operators $\{ \cB^\alpha \}_{\alpha \in \cS\cO_0}$
\begin{equation}\label{Balpha}
\cB^\alpha (T_1 \sqcup \cdots \sqcup T_n)= \Tree[ $T_1$ $T_2$ $\cdots$ $T_N$ ].$\alpha$
\end{equation}
that act as in \eqref{grafting}, but also label the {\em root} of the resulting structure by $\alpha$.

\smallskip
\subsubsection{Colored Merge and colored grafting}

In fact, once we also take into consideration the filtering of syntactic objects by theta roles 
and well formed phases, as in \cite{MHL}, \cite{MarLar}, one considers a family
$\{ \cB^{(\alpha,\fc)} \}_{\alpha \in \cS\cO_0, \fc\in \Omega}$, where $\Omega$ denotes
a set of colors, as discussed in \cite{MHL}, that label a colored operad that simultaneously
implements the theta-roles coloring and the phases-coloring. 

With these colorings taken into consideration, the form of the Merge operators
$\fM_{S,S'}$ now requires that $S,S'\in \cL(\bB_{\Phi,\Theta})$ are syntactic objects
in the language of the colored operad $\cO(\bB_{\Phi,\Theta})$ of both theta-roles and
phases, and
\begin{equation}\label{bicoloredMerge}
\fM_{S,S'} = \sqcup \sum_{\alpha \in \cS\cO_0, \fc\in \Omega} (\cB^{(\alpha,\fc)} \otimes {\rm id}) \circ \delta_{\fc_S, \fc_{S'}}^\fc \,\delta_{\alpha, \alpha(h(\fM(S,S'))} \, \delta_{S,S'} \circ \Delta \, ,
\end{equation}
where $\delta_{S,S'}$ is as in \eqref{opMerge}, and $\delta_{\fc_S, \fc_{S'}}^\fc=1$ if there is a colored operad generator
in $\cO(\bB_{\Phi,\Theta})$ with root color $\fc$ and leaf colors $\fc_S, \fc_{S'}$, respectively the root colors of $S,S'$,
and is zero otherwise, while $\delta_{\alpha, \alpha(h(\fM(S,S'))}=1$ if $\alpha$ is equal to the element in $\cS\cO_0$
assigned to the head $h(\fM(S,S'))$ of the tree $\fM(S,S')$ and zero otherwise. 
(Note that the condition $\delta_{\fc_S, \fc_{S'}}^\fc=1$
ensures that $\fM(S,S')\in \cL(\bB_{\Phi,\Theta})$, hence in particular it has a well defined head, so this condition
is meaningful.)

\smallskip

\smallskip
\subsection{The universal property}\label{univpropBsec}

An very important property of the grafting operator $\cB$ satisfying the Hochschild
cocycle condition \eqref{CKcocycle} on the Connes-Kreimer Hopf algebra $\cH_{CK}$
is a {\em universal property} for the datum $(\cH_{CK}, \cB)$, as shown in Theorem~31 
of \cite{Foissy} and Theorem~2.4.6 of \cite{Panzer}). 

This property can be stated in the following way.
\begin{itemize}
\item Given unital associative commutative algebra $\cR$ (over the same field of definition as $\cH_{CK}$),
and a linear map $\psi: \cR \to \cR$, there exists a unique algebra homomorphism $\eta: \cH_{CK} \to \cR$
satisfying
\begin{equation}\label{univpropBeq}
\eta \circ \cB = \psi \circ \eta.
\end{equation}
\item If $\cH$ is a commutative bialgebra (or a Hopf algebra), and the linear map $\psi: \cH \to \cH$ is a Hochschild $1$-cocycle
on $\cH$, then the unique homomorphism of algebras $\eta: \cH_{CK} \to \cH$ is in fact also a bialgebra (or a Hopf algebra)
homomorphism.
\end{itemize}

The construction of the algebra homomorphism $\eta$ is obtained inductively as the solution to
\begin{equation}\label{eqFoissy}
\eta (T) = \psi \circ \eta \circ \pi_{C_{\rm root}}(T)\, ,
\end{equation}
where $F=\pi_{C_{\rm root}}(T)$ is the forest obtained by the admissible cut $C_{\rm root}$ that cuts every edge 
attached to the root of $T$. (In the literature on Hopf algebras of rooted trees, one typically uses $\cB^+$ 
to denote the grafting operator $\cB$ of \eqref{grafting} and $\cB^-$ to denote this cutting
operator.) 

\smallskip

In the colored case, we look at trees and forests where the internal, non-leaf vertices are also labelled, and the
universality property of \cite{Foissy} is in fact stated for these labelled $\cB^{(\alpha,\fc)}$, namely
the inductive property of the homomoprhism $\eta$ in \eqref{eqFoissy} takes the form
\begin{equation}\label{eqFoissyalpha}
\eta (T) = \psi_{(\alpha,\fc)} \circ \eta \circ \pi_{C_{\rm root}}(T)\, ,
\end{equation}
where $(\alpha,\fc)$ is the label carried by the root of $T$, and $\{ \psi_{(\alpha,\fc)} \}_{(\alpha,\fc)}$
is a family of linear maps on the target algebra $\cH$, that are 1-cocycles in the case where $\cH$ is a
bialgebra or Hopf algebra. 

\smallskip

This universal property has interesting implications for models of syntax-semantics interface as
developed in \cite{MCB}, but in order to analyze that more carefully we first need to adjust
our setting, passing from the Connes-Kreimer Hopf algebra $\cH_{CK}$ to the Hopf algebra
of workspaces. We will not discuss this in full here, as some of the more technical mathematical 
aspects will be treated elsewhere.

\smallskip
\subsection{The Hopf algebra of workspaces}\label{HopfWorkSec}

When we consider the ``Hopf algebra of workspaces" of \cite{MCB}, we need to pay attention to the
subtleties concerning the form of the coproduct and the corresponding algebraic
structure, as discussed in \S 1.2 and 1.3 of \cite{MCB}. Indeed, what is referred to, for
simplicity, as the ``Hopf algebra of workspaces" consists in fact of three different levels of
structure, for the three forms $\Delta^c$, $\Delta^\rho$, and $\Delta^d$ of the coproduct:
\begin{enumerate}
\item {\em Bialgebra of workspaces}: $\cH=(\cV(\fF_{\cS\cO_0}), \sqcup, \Delta^c)$, which becomes a
Hopf algebra modulo the ideal generated by ${\bf 1}-\cS\cO_0$.
\item {\em Hopf algebra and comodule}: $\cH_{CK}^{\leq 2}=(\cV(\fF^{\leq 2}_{\cS\cO_0}), \sqcup, \Delta^\rho)$
with $(M=\cV(\fF_{\cS\cO_0}), \rho_R=\Delta^\rho|_{\cV(\fF_{\cS\cO_0}})$ a right-comodule.
\item Hopf algebra: $\cH=(\cV(\fF_{\cS\cO_0}), \sqcup, \Delta^d)$, coassociative with restrictions on
admissible cuts or a ``generalized quasi-Hopf algebra" with a correction to
co-associativity.
\end{enumerate}
As discussed in \cite{MCB}, the coproducts $\Delta^c$ and $\Delta^d$ produce the forms with
or without the {\em trace} of the deleted deeper copy in the copy theory of movement, that
correspond to the CI and the SM interfaces, respectively, with $\Delta^\rho$ relating them. 
The coproduct $\Delta^\rho$ also plays an important role at the interface between syntax
and morphology. 

The details about the coassociativity property for the third form $\Delta^d$ of the coproduct 
will be analyzed in more detail elsewhere, so we do not discuss this further here. One should
just keep in mind that the main issue in the coassociativity of $\Delta^d$ arises in how to
handle the terms $\pi_C(T) \otimes \rho^d_C(T)$ of the coproduct $\Delta^d(T)$, where
the admissible cut $C$ cuts both edges below a given vertex $v$. 

We will not analyze in the present paper the Hochschild cocycle equations for
these different settings and their meaning. These will be discussed elsewhere.
Note, however, that the restriction $\cB \circ \Pi_{(2)}$ of $\cB$ to binary forests
is not by itself a Hochschild cocycle for the Hopf algebra of workspaces. The
Hochschild cocycle property of $\cB$ relies on the existence of the larger
Hopf algebra $\cH_{CK}$. 

\smallskip

We make the following observation that we will return to in \S \ref{implcocycleSec}.
Consider the Connes-Kreimer Hopf algebra $\cH_{CK}$ with the Hopf algebra Markov
chain \eqref{CKMarkov} determined by the cocycle $\cB$. The universal property for $(\cH_{CK},\cB)$
implies that, given $(\cH,\psi)$, another bialgebra or Hopf algebra $\cH$ with a Hochschild $1$-cocycle $\psi$,
the bialgebra (Hopf algebra) homomorphism $\eta: \cH_{CK} \to \cH$ determined by the universal
property also intertwines the Hopf algebra Markov chains determined by $\cB$ and $\psi$, respectively.

\begin{lem}\label{etaMarkov}
For $(\cH,\psi)$ a bialgebra (or Hopf algebra) $\cH=(\cV,\star,\Delta)$ with a Hochschild $1$-cocycle $\psi$, let
\begin{equation}\label{Kpsi}
\cK_\psi := \star \circ (\psi \otimes {\rm id}) \circ \Delta 
\end{equation}
be the associated Hopf algebra Markov chain determined by $\psi$. The unique homomorphism
$\eta: \cH_{CK} \to \cH$ satisfying the universal property also satisfies
\begin{equation}\label{etaK}
\cK_\psi \circ \eta = \eta \circ \cK_{CK} \, .
\end{equation}
\end{lem}

\proof We know that the homomorphism $\eta: \cH_{CK} \to \cH$ satisfying the universal property is
a bialgebra (or Hopf algebra) homomorphism satisfying $\eta \circ \cB = \psi \circ \eta$. In particular,
it also satisfies $\eta(F\sqcup F')=\eta(F)\star \eta(F')$ and $(\eta\otimes\eta)\circ \Delta_{CK}(F) = \Delta \eta(F)$.
Thus, we have
$$ \cK_\psi \circ \eta = \star \circ (\psi \otimes {\rm id}) \circ \Delta \circ \eta =
\star \circ (\psi \otimes {\rm id}) \circ (\eta\otimes\eta)\circ \Delta_{CK} $$
$$ = \star \circ (\psi \circ \eta \otimes \eta) \circ \Delta_{CK} = \star \circ (\eta\otimes \eta) \circ (\cB \otimes {\rm id}) \circ  \Delta_{CK} $$
$$ = \eta \circ \sqcup \circ (\cB \otimes {\rm id}) \circ  \Delta_{CK} = \eta \circ \cK_{CK}  \, . $$
\endproof

This shows that any Hopf algebra Markov chain determined by a Hochschild $1$-cocycle
is induced by the Hopf algebra Markov $\cK_{CK}$ on the Connes-Kreimer Hopf algebra. 
This can be read as a universal property for dynamical systems of the form \eqref{Kpsi},
where $\psi$ is a Hochschild $1$-cocycle. 

\smallskip

\begin{prop}\label{MergeKeta}
Consider then the Hopf subalgebra $\cH_{CK}^{\leq 2}$ with
the induced Merge Markov chain
$$ \cK= \sqcup \circ (\cB \otimes {\rm id}) \circ \Pi_{(2)} \circ \Delta^\rho \, , $$
and the image $\eta(\cH_{CK}^{\leq 2})\subset \cH$. Let 
$\cL \subset \eta(\cH_{CK}^{\leq 2})$ be the linear subspace spanned by
the images $\eta(F)$ of those forests $F\in \tilde\fF_{\cS\cO_0}^{\leq 2}$ that have two
connected components $F=T_1 \sqcup T_2$, and let $L_{(2)}$ be the
linear projection onto this subspace. If $L_{(2)}$ satisfies
\begin{equation}\label{etaL2}
\eta \circ \Pi_{(2)}=L_{(2)} \circ \eta \, ,
\end{equation}
then the Hopf algebra Markov chain
\begin{equation}\label{Kpsi2}
\cK_\psi^{(2)}:= \star \circ (\psi \otimes {\rm id}) \circ L_{(2)} \circ \Delta
\end{equation}
is induced by the Merge Hopf algebra Markov chain $\cK=\sqcup\circ (\cB\otimes {\rm id})\circ \Pi_{(2)}\circ \Delta^\rho$, 
namely it satisfies
\begin{equation}\label{KetaK}
\cK_\psi^{(2)}\circ \eta = \eta \circ \cK \, .
\end{equation}
as linear maps $\cV(\tilde\cF_{\cS\cO_0}^{\leq 2}) \to \eta(\cV(\tilde\cF_{\cS\cO_0}^{\leq 2}))$. 
\end{prop}

\proof
This follows exactly as in Lemma~\ref{etaMarkov}, with the additional property \eqref{etaL2} giving
$$ \eta \circ \cK = \eta \circ \sqcup \circ (\cB\otimes {\rm id})\circ \Pi_{(2)}\circ \Delta^\rho = 
\star \circ (\eta \otimes \eta) \circ (\cB\otimes {\rm id})\circ \Pi_{(2)}\circ \Delta^\rho $$
$$ = \star \circ (\psi \otimes {\rm id})\circ (\eta \otimes \eta)  \circ \Pi_{(2)} \circ \Delta^\rho =
 \star \circ (\psi \otimes {\rm id})\circ L_{(2)} \circ (\eta \otimes \eta)  \circ \Delta_{CK} $$ $$ =
  \star \circ (\psi \otimes {\rm id})\circ L_{(2)} \circ \Delta \circ \eta = \cK_\psi^{(2)}\circ \eta\, . $$
  \endproof

\smallskip
\subsection{Implications of the cocycle condition}\label{implcocycleSec}

As we have mentioned, the Hochschild 1-cocycle condition for the
grafting operator $\cB$ on the Connes-Kreimer Hopf algebra $\cH_{CK}$ plays an
important role in the Hopf algebra approach to renormalization in physics. 
There are in fact two main ways in which the cocycle condition is important. 
We will see that the second one is the one that is more directly relevant to 
our linguistic context.

A first role that the cocycle condition of the grafting operator $\cB$ on the Connes--Kreimer
Hopf algebra plays in perturbative quantum field theory is in the construction of
solutions to the equations of motion, the Dyson--Schwinger equations. Some of this 
was already discussed in \cite{MCB} and we review it briefly here. 

The combinatorial Dyson--Schwinger equations can be formulated as 
a fixed-point equation, $X=\cB(P(X))$, with $X$ a variable $X=\sum_\ell X_\ell$
with $X_\ell\in \cH_\ell$ the graded subspaces of the Hopf algebra, with $P$
a polynomial or formal power series. This fixed-point equation, when written
recursively by degree, encodes the generative process of the combinatorial
objects that span a Hopf subalgebra of the Hopf algebra $\cH_{CK}$. The
cocycle condition satisfied by $\cB$ ensures that the sums of generators 
of the Hopf algebra that occur in these recursive solutions of 
combinatorial Dyson--Schwinger equations indeed generate Hopf subalgebra.

In the case of Merge, the analogous equation $X=\cB(X^2)$, which is the
simplest form the above fixed-point problem can take, this was
discussed in \S 1.17 of \cite{MCB}, where it is shown that the recursive
solution encodes the generative process of the syntactic objects. Thus,
one can view the core generative process
of Merge as the simplest and most fundamental case of
such DS equations, with a single quadratic term as polynomial. 

There is also a second role in physics for the cocycle condition satisfied by $\cB$,
which is related to the universal property of $(\cH_{CK}, \cB)$ that we discussed above.

In the theory of renormalization one considers formal Feynman rules that are
characters of the Hopf algebra $\cH_{CK}$, meaning morphisms of commutative
algebras $\phi: \cH_{CK} \to \cR$ with values in a target algebra (or semiring in the
linguistics case), where the target algebra admits a factorization in the form 
of a Rota--Baxter structure. The latter is used to inductively separate a meaningful
(convergent) part from a meaningless (divergent) part. Normally, it is not required that
the target algebra $\cR$ would also have a bialgebra or Hopf algebra structure and that
the morphism $\phi$ would also preserve this additional structure. The renormalization
procedure in physics does not require these additional structures.

The analogous setting in linguistics is the syntax-semantics interface and 
semantic parsing, as discussed in Chapter~3 of \cite{MCB}. 
In Chapter~3 of \cite{MCB} it is argued that in order to obtain a viable model of syntax-semantics
interface, one does not need to require a compositional structure on the side of semantics that
mirrors the Merge operation of syntax. In fact, the syntax-semantics interface is syntax-driven
in the sense that the compositional structure of syntax suffices for semantic parsing. All that
is needed on the side of semantics is some topological notion of proximity (or metric
distance, or similarity) and a movable threshold on the amount of similarity/proximity.
These properties can be subsumed into the notion of a Rota-Baxter algebra (or Rota-Baxter 
semiring). This makes it possible, for example, to have a working syntax-semantics interface
in settings like vector space models of semantics, where the semantics side is very impoverished.

However, it is observed for instance in \cite{Panzer} that, in the physics
setting, some of the physically relevant Feynman rules do in fact carry
additional structure, where the target $\cR$ is also a bialgebra and
the Feynman rule morphism $\phi: \cH_{CK} \to \cR$ is also a morphism of
bialgebras that intertwines the 1-cocycle $\cB$ and a 1-cocycle on $\cR$,
as in the universality property of \cite{Foissy}, recalled above. In fact, the
universal property specifies that any such target that carries the same
formal structure (Hopf algebra with a 1-cocycle) must come from $\cH_{CK}$
with the cocycle $\cB$ via a homomorphism. The pair $(\cH_{CK},\cB)$ is
universal in that sense, as we have recalled above. This additional structure
can carry advantages with respect to more impoverished forms of the target algebra $\cR$.

In the linguistic case one can see a similar situation. 
While indeed semantics does not need to have its own compositional structure, for a
viable model of syntax-semantics interface to work, nonetheless there are models of 
semantics that do incorporate compositionality.  For instance, Pietroski's semantics was
discussed at length in Chapter~3 of \cite{MCB}. In that formalism a Combine operation
mirrors the structure building operation Merge of syntax as a compositional structure
for concepts. It is shown in \cite{MCB} that this operation can in fact be directly induced
by the syntactic Merge without the need to be independently defined in semantics. It is also
shown in Chapter~3 of \cite{MCB} that other compositional models of semantics, such
as Heim-Kratzer semantics, fit into a Merge-driven model of syntax-semantics interface.

More generally, the universal property of $(\cH_{CK}, \cB)$ discussed above
suggests that one can consider a type of ``compositional" models of semantics
where the composition operation is affected by a $1$-cocycle $\psi$, namely
models based on an operation like what we described in \eqref{Kpsi}. In
such cases, the universal property of $(\cH_{CK}, \cB)$ and the restriction
described in Proposition~\ref{MergeKeta} imply that such compositional
structures can also be implemented in a model of syntax-semantics
interface based on the action of Merge on syntax, through the
intermediate role of the Hopf algebra Markov chain $\cK_{CK}$ of \eqref{CKMarkov}
on the Connes-Kreimer Hopf algebra. We will not discuss this further in the present paper.

\smallskip
\subsection{EC-violating insertions and Lie algebra}\label{ECLieSec}

We have seen that the grafting of trees at the root, the basic operation of
Merge, is implemented by the grafting operator $\cB$ that has a specific
algebraic property, the Hochschild cocycle condition, which plays a role
in an associated universal property. We can now compare, on the basis of 
algebraic properties, this grafting at the root (which does not violate EC)
with other insertions in the interior of the tree, that do violate EC. 

We consider the EC-violating insertion operations proposed
in Countercyclic Merge and Late Merge, and we show that these do not
satisfy the cocycle condition satisfied by the operator $\cB$ of grafting at the root.
Indeed, we show that these other insertions obey different algebraic properties 
with respect to the product and coproduct operations of the Hopf algebra.
This implies that the Extension Condition has an algebraic characterization,
hence it should be regarded as a hard constraint of the mathematical model
and not as a soft-constraint of cost optimization. 

\smallskip

Consider instead the insertion operation that is involved in countercyclic movement
(such as Late Merge) that causes violation of the Extension Condition, by growing
a tree structure at a location different from the root. As discussed in \S 1.7 of \cite{MCB},
insertions of a binary rooted tree $T$ into another such tree $T'$ at a location other than
the root vertex, modeling Late Merge, consists of the pre-Lie insertion operation. 
We assume that $T'$ (the structure where the insertion takes place) has a non-empty set
of edges (some Merge operations should have taken place for Late Merge to be
considered). Given an edge $e \in E(T')$, one defines $T' \lhd_e T$ to be the binary
rooted tree obtained by splitting $e$ into two edges $e', e''$ by the insertion of a new
nonbranching vertex $v$, then attaching the root of $T$ to $v$ via a new edge $\tilde e$.
We write
$$ \delta_T (T')=\sum_{e\in E(T')} T' \lhd_e T \, , $$
sometimes also written as $\delta_T(T')=T' \lhd T$, where $\lhd$ is the pre-Lie
product of the dual Lie algebra (see the discussion in  \S 1.7 of \cite{MCB}). 
For simplicity, we analyze here the case of the linear operators $\delta_\alpha$
with $\alpha \in \cS\cO_0$ (i.e., a tree consisting of a single labelled leaf). This
case will suffice to illustrate why the type of cocycle condition associated to
the grafting operator $\cB$ and the resulting $\tilde\cB^\alpha_{(2)}$ no longer
holds for the operator $\delta_\alpha$ (and a fortiori for the $\delta_T$ with
more complex structures $T$).

\smallskip

\begin{cor}\label{nococins}
The insertion operators $\delta_\alpha$ are derivations on the bialgebra
$(\cV(\fF_{\cS\cO_0}),\sqcup, \Delta^c)$ but are not 1-cocycles of the form \eqref{1coeq}.
\end{cor}

\proof
First observe that, because the insertion $\delta_\alpha$ acts on trees with
non-empty set of leaves, we want to consider the quotient of $(\cV(\fF_{\cS\cO_0}),\sqcup)$
by the ideal generated by $1-\alpha$ as in the first case of the bialgebra of workspaced
recalled above, graded by number of edges, with the coproduct $\Delta^c$. 
The insertion $\delta_\alpha$ is a derivation with respect to the
product $\sqcup$, namely $\delta_\alpha (T_1\sqcup T_2)=\delta_\alpha (T_1)\sqcup T_2+T_1 \sqcup \delta_\alpha (T_2)$,
since the edge where the insertion happens is in either $T_1$ or $T_2$. Notice then that 
the $\delta_\alpha$ defined in this way will not satisfy the cocycle condition \eqref{1coeq} for
the bialgebra $(\cV(\fF_{\cS\cO_0}),\sqcup, \Delta^c)$, since in the left-hand-side
$\Delta^c \circ \delta_\alpha (T)$ will contain terms of the form $\delta_\alpha(\pi_C(T)) \otimes \rho_C(T)$,
for nontrivial admissible cuts of $T$, which do not occur on
the right-hand-side, $(\delta_\alpha\otimes 1+ ({\rm id}\otimes \delta_\alpha)\circ \Delta)(T)$, where
all the insertions only occur in terms $\pi_C(T) \otimes \delta_\alpha(\rho_C(T))$ or $\delta_\alpha(T) \otimes 1$. 
\endproof

By a similar argument, $\delta_\alpha$ also does not satisfy the other cocycle
relation, as $\Delta^c \circ \delta_\alpha (T)$ includes 
terms $\pi_C(T)) \otimes \delta_\alpha(\rho_C(T))$, while $(\delta_\alpha\otimes {\rm id})\circ \Delta(T)$
does not.

\endproof

\section{Cost functions and forms of Merge}\label{CostsSec}

In this section we analyze different forms of cost functions (Minimal Search and 
Resurce Restrictions, in particular Minimal Yield and No Complexity Loss).
We summarize how different forms of Merge (External Merge, Internal Merge,
Sideward Merge) present in the mathematical model are weighted differently
by these cost functions, identifying External Merge and Internal Merge as
optimal and Sideward Merge as suboptimal, with different levels of violation
of optimality. 

\smallskip
\subsection{Different forms of Sideward Merge} \label{typesSMsec}

As discussed in \S 1.3 of \cite{MCB}, there are three different forms of
Sideward Merge that occur as operators $\fM_{S,S'}$ of the form \eqref{opMerge},
which we illustrate here, for simplicity, with a workspace $F=T\sqcup T'$ that
contains only two syntactic objects (for larger workspaces, other components
will remain untouched in the resulting new workspace, so this case suffices to
completely describe the result of these operations):
\begin{enumerate}
\item $S=T_v\subset T$ an accessible term of the syntactic object $T$, and $S'=T'$ 
the other syntactic object, with resulting new workspace
$$ \fM(T_v, T') \sqcup T/T_v\, . $$
\item $S=T_v\subset T$ as above and $S"=T'_w\subset T'$ an accessible term of the syntactic object $T'$, 
with resulting new workspace
$$ \fM(T_v, T'_w) \sqcup T/T_v \sqcup T'/T'_w \, . $$
\item $S=T_v\subset T$ and $S'=T_w \subset T$ two disjoint accessible terms of the same
syntactic object $T$, with resulting new workspace
$$ \fM(T_v, T_w) \sqcup T/(T_v \sqcup T_w) \sqcup T' \, . $$
\end{enumerate}

\smallskip
\subsection{Merge constraints: a discussion of Minimal Search}  \label{MSsev}

We recall more specifically the constraints on Merge mentioned above, 
starting with a more detailed discussion of Minimal Search. Though the
main focus for the purpose of the analysis in this paper will be on
Resource Restriction optimality constraints (which we
discuss in \S \ref{MYsec} and \S \ref{NCLsec} below), we also 
include a brief  discussion of Minimal Search here, both for completeness
and in view of the discussion on the Hopf algebra Markov chain property of
Merge, in \S \ref{HopfMArkovSec}. 

\smallskip

The formulation of Minimal Search given in \S 1.5 of \cite{MCB} is based on a
construction of a cost function that realizes the minimality, summarized as follows. 
Let $d_v$ denote the distance of a non-root vertex of $T\in \fT_{\cS\cO_0}$
from the root of $T$. A cost function is assigned in \S 1.5 of \cite{MCB} to the 
operations of extracting an accessible term $T_v \subset T$, of taking the
quotient $T/T_v$, and of grafting together a forest to a resulting tree.
The cost function $c\in \Z$ has an associated weight $\epsilon^{c}$ for a
real parameter $\epsilon>0$. 
\begin{itemize}
\item The cost of extraction of $T_v \subset T$ depends on how deep $T_v$ is
located inside $T$ (which accounts for length of search), so the cost of this
operation is set equal to $c(T_v)=d_v$, hence weight $\epsilon^{d_v}$.
\item The cost of contraction of $T_v$ to form $T/T_v$ is larger the more
significant the difference between the original tree $T$ and $T/T_v$, which
is the more easily detectable the closer $v$ is to the root of $T$, so that
the corresponding weight should be inversely proportional to the case
of extraction. This gives cost $c(T/T_v)=-d_v$ and weight $\epsilon^{-d_v}$.
\item the resulting $\fM(T, T')$ carries a weight $\epsilon^{c(\fM(T,T'))}=\epsilon^{d+d'}$,
for $c(T)=d$ and $c(T')=d'$. 
 \end{itemize}
 The only zero-cost Merge operations are Internal and External Merge, while the
 Sideward Merge cases carry a non-zero cost: in the EM case $d=d'=0$ and
 in the IM case $d=d_v=-d'$. In the three forms of SM, on the other hand, 
 the term $\fM(T_v, T')$ has cost $d_v$, the term $\fM(T_v,T'_w)>0$ has cost
 $d_v+d'_{w}>0$, and $\fM(T_v,T_w)$ has cost $d_v+d_w>0$.
 
 \smallskip
 
 Note that the condition for a zero-cost operation is that
 it has cost zero (weight one) {\em for all} workspaces $F$ 
 it applies to. This does not count as zero-cost operations
 for which there may exist a special choice of $F$ on which
 it has cost zero while having non-zero cost on other workspaces.
 \footnote{Such special cases may occur when one considers compositions
 of Merge operations where some are allowed to be of the form
 $\fM_{S,1}$, but such cases do not count as zero-cost in the sense 
 of \cite{MCB}.}
 
 \smallskip
 
 While it is Resource Restriction/Minimal Yield rather than Minimal Search 
 that matters for the answer to the question on different levels of
 acceptability of different forms of SM, we can make here some further 
 general observations that are useful to clarify the role and functioning
 of the Minimal Search constraint. In particular, we propose here 
 a slightly different way of implementing a cost function that realizes
 minimality. This will be equivalent in terms of minimization, but will
 have two advantages over the construction of \cite{MCB} recalled above:
 \begin{itemize}
 \item The cost function only depends on the grading of the Hopf algebra, hence it is completely
 intrinsically defined by the algebra of the model (while the datum $d_v$ is part of the
 syntactic objects, it does not explicitly play a role in the algebraic structure of the action of
 Merge on workspaces).
 \item The cost function we construct here is compatible with the fact that all structures
 are built bottom-up rather than top-down, hence it is more natural to think of the search
 that selects computational material for the action of Merge as also occurring in the same
 bottom-up direction (while the cost function based on $d_v$ presumes a top-down
 search into the tree starting at the root).
 \end{itemize}
 
 \smallskip
 
  Indeed, the cost and weight functions
  described above reflect a common assumption in computational processes
  based on tree-structures, that a search starts at the root of the tree and
  proceeds towards the leaves, making the search more costly the further
  away one moves from the root. This is reflected here in the assignment
  of a cost $c(T_v)=d_v$ to the extraction of an accessible term $T_v\subset T$,
  with the cost increasing with the distance $d_v$ of the root vertex $v$ of $T_V$
  from the root of $T$. Thus, in the limit $\epsilon \to 0$ the dominant terms with
  respect to the weight $\epsilon^{d_v}$ would be those with $d_v$ small, 
  that is, $v$ close to the root. This description may not be
 completely satisfactory in the case of syntax, as it does not reflect
 the fact that syntactic objects are built from the bottom up, rather than from the
 top down, and that the root (and the internal nodes) carry linguistic significance only
 by identifying constituencies (phrases) $T_v$, which are themselves bottom-up constructions
 from a subset $L(T_v)\subset L(T)$ of the lexical material.  
 This suggest that it should be more natural to think of the search for 
 accessible terms $T_v \subset T$ as a procedure that also unfolds bottom-up, 
 namely with first looking for accessible
 terms that are ``atomic" (leaves of the tree $T$) and then for accessible terms $T_v$
 that are larger in terms of the size $\ell(T_v)$, hence necessarily 
 involving vertices $v$ that are higher up towards the root. 
  
  \smallskip
  
  The use of the distance $d_v$ can be
  replaced by other measures that express the same optimality of 
  IM and EM as the only zero-cost cases, while focusing on a
  bottom-up search in which the search involving accessible terms that
  are atomic (leaves) comes first with respect to a choice that
  involves larger structures. 
  
  \smallskip 
  
  For $T\in \fT_{\cS\cO_0}$ a syntactic object or $F\in \fF_{\cS\cO_0}$ a workspace, we write
  $L=L(T)$ (respectively, $L=L(F)$) for the set of leaves, with $\ell(T)=\# L(T)$ 
   and $\ell(F)=\sum_a \ell(T_a)$ for $F=\sqcup_a T_a$ the grading of
  the vector space $\cV(\fF_{\cS\cO_0})$. 
  
  \begin{defn}\label{costdef}
  For $T\in \fT_{\cS\cO_0}$, define the {\rm cost of
  extraction} of an accessible term $T_v \subset T$ as 
  \begin{equation}\label{costTv}
  \fc(T_v):= \frac{\ell(T_v)}{\ell(T)} \, .
  \end{equation}
  Similarly, we define the cost of the associated quotient as
   \begin{equation}\label{costTTv}
  \fc(T/T_v):= 1-\frac{\ell(T_v)}{\ell(T)} \, .
  \end{equation}
  We also associate a cost to the Merge operation in the form
  \begin{equation}\label{costM}
  \fc(\fM(A,B)):=\fb(A,B) - \fc(A)-\fc(B)\, ,
  \end{equation}
  where $\fb(A,B)$ counts the number of connected components in the workspace from which the
  Merge material $A,B$ is taken:  $\fb(A,B)=1$ if $A$ and $B$ both come from the same connected
  component of the workspace and $\fb(A,B)=2$ if $A$ and $B$ belong to different components.
  The cost of a composition of Merge operations is the sum of the costs of the individual operations.
  \end{defn}
  
  Defining the cost $\fc(T_v)$ as in \eqref{costTv} amounts to estimating the cost of
  finding the structure $T_v$ by searching for it at the level of the leaves, namely
  the fraction of the size of the set $L(T)$ that needs to be extracted when taking the
  accessible term $T_v$: the fraction of the size is exactly $\ell(T_v)/\ell(T)$. With this
  type of cost function, extracting a single leaf from a large tree is very low cost, but
  extracting an accessible term with a large size $\ell(T_v)$ compared to the size
  of the ambient tree has a higher cost. Selecting an entire component has cost $1$ (the maximal normalized cost).
  The associated cancellation also comes with a cost, as in the setting discussed
  in \cite{MCB}: setting this cost $\fc(T/T_v)$ to be equal to $1-\fc(T_v)$ is again
  expressing the idea that the cost of $T_v$ is ``compensated" by the cost of $T/T_v$,
  the easier it is to extract an accessible term, the more costly it is to delete it and
  viceversa. The justification in this case is different and simpler than the corresponding
  discussion in \cite{MCB}: here one can simply write 
  $$ 1-\fc(T_v)=\frac{\ell(T)-\ell(T_v)}{\ell(T_v)} = \frac{\ell(T/T_v)}{\ell(T)} \, , $$
  provided the quotient is either $T/^d T_v$ or $T/^\rho T_v$ (see \S 1.2 of \cite{MCB}).
  Thus, the cost $\fc(T/T_v)$ of \eqref{costTTv} is again the same as \eqref{costTv},
  that is, the relative size of $T/T_v$ compared to the original $T$. Finally,
  the definition of the cost of the operation $\fM$ in terms of $\fb$ reflects the
  idea, in linguistics, that Internal Merge should in some way be more essential
  and simpler than External Merge. Here, although both end up being zero-cost
  operations (as in the formulation in \cite{MCB}), they nonetheless differ in
  the fact that the contribution from the number of components involved is
  higher in EM than in IM. (This can be compared with the discussion at the end
  of \S 1.6 of \cite{MCB} on EM as ``more complex" than IM.)  The following properties
  then follow directly from this definition of the cost function.
  
  \begin{prop}\label{MSprop}
  The cost function as in Definition~\ref{costdef} has the following properties:
  \begin{itemize}
  \item The minimal cost of extraction of an accessible term is the atomic case (a single leaf) with
  cost $\fc(\alpha)=1/\ell(T)$, the maximal cost is $1$, which is realized only in the case
  of selecting an entire component.
  \item Internal Merge and External Merge are the only zero-cost operations:
  $$ \begin{array}{ll} {\bf {\rm EM}}: &  \fc(\fM(T,T'))=2-\fc(T)-\fc(T')=0 \\  
  {\bf {\rm IM}}: & \fc(\fM(T_v,T/T_v))=1-\fc(T_v)-\fc(T/T_v) =0 \, .  \end{array} $$
  \item The three forms of Sideward Merge have cost:
  $$ \begin{array}{ll} {\bf {\rm SM(1)}}: &    \fc(\fM(T_v,T'))=2-\fc(T_v)-1=1- \fc(T_v) > 0    \\ 
   {\bf {\rm SM(2)}}: &  \fc(\fM(T_v, T'_w)) = 2 -\fc(T_v) - \fc(T_w) >0  \\  
  {\bf {\rm SM(3)}}:  & \fc(\fM(T_v, T_w)) = 1 -\fc(T_v) - \fc(T_w) >0 \, ,  \end{array} $$
  \end{itemize}
  where in the last case $F_{\underline{v}}=T_v\sqcup T_w \subset T$ with $\ell(F_{\underline{v}})=\ell(T_v)+\ell(T_w) < \ell(T)$,
  so that we obtain $\fc(\fM(T_v, T_w)) >0$. 
  \item The cost $\fc(T_v)$ is smallest for an atomic component (a leaf), $\fc(\alpha)=1/\ell(T)$ and maximal at the
  largest of the two accessible terms below the root, $\max_v \fc(T_v)=\max\{ \fc(T_1), \fc(T_2) \}$ for $T=\fM(T_1,T_2)$.
  \item The cost for a Sideward Merge operation is maximal when the accessible terms $T_v$ and $T_w$ are atomic.
  \end{prop}
  
  The last property appears at odds with the previous one: the cost of extraction of $T_v$ is minimal for a single
  leaf, $T_v=\alpha$, but the cost of the resulting Sideward Merge is maximal in this case. The fact that
  SM is less costly when involving accessible terms near the root than deeper in the tree is similar
  to the behavior of the cost function used in \cite{MCB}, albeit differently defined, and in fact this
  behavior is pretty much fixed by the condition of having a single simple expression for the
  cost that assigns cost zero to both External and Internal Merge. As we will discuss below, other cost
  functions related to Minimal Yield and Resource Restriction counting, do not have this property and
  will assign minimal cost to the SM with atomic elements. 
  
  \smallskip
  
  Moreover, one additional cost (regardless of the choice of cost function) in using SM operations
  that extract a single leaf are the extra costs caused by necessarily increasing the
  length of the resulting derivation (the number of Merge operations performed, and this is 
  part of what the cost function of the Merge operations accounts for singe they add over the
  chain of operations in a derivation). To illustrate this, consider
  the following example. There are two ways of obtaining the forest
  $$ F = \Tree[ [ $\alpha$ $\beta$ ] [ $\gamma$ $\delta$ ] ] \sqcup  \Tree[ $\eta$ $\xi$ ] $$
  from the tree
  $$ T = \Tree[ [ $\eta$ [ [ $\alpha$ $\beta$ ] [ $\gamma$ $\delta$ ] ] ] $\xi$ ] $$
  by applications of Sideward Merge. One is as the single SM operation
  $$ T \mapsto  \Tree[ $\alpha$ $\beta$ ] \sqcup \Tree[ $\gamma$ $\delta$ ] \otimes \Tree[ $\eta$ $\xi$ ] \mapsto
  \Tree[ [ $\alpha$ $\beta$ ] [ $\gamma$ $\delta$ ] ] \otimes \Tree[ $\eta$ $\xi$ ] \mapsto F $$
  and the other is as two successive applications of SM followed by one application of EM: first
   $$ T \mapsto \alpha \sqcup \beta \otimes   \Tree[ [ $\eta$  [ $\gamma$ $\delta$ ] ]  $\xi$ ]  \mapsto  
   \Tree[ $\alpha$ $\beta$ ] \otimes \Tree[ [ $\eta$ [ $\gamma$ $\delta$ ] ] $\xi$ ] \mapsto F_1=\Tree[ $\alpha$ $\beta$ ] \sqcup 
  \Tree[ [ $\eta$ [ $\gamma$ $\delta$ ] ] $\xi$ ] $$
  then
  $$ F_1 \mapsto \gamma \sqcup \delta\,\,  \otimes   \Tree[ $\eta$ $\xi$ ]   \sqcup  \Tree[ $\alpha$ $\beta$ ] \mapsto 
   \Tree[ $\gamma$ $\delta$ ] \otimes \Tree[ $\eta$ $\xi$ ] \sqcup \Tree[ $\alpha$ $\beta$ ]\mapsto
  F_2= \Tree[ $\gamma$ $\delta$ ] \sqcup \Tree[ $\eta$ $\xi$ ] \sqcup \Tree[ $\alpha$ $\beta$ ] $$
  and finally
  $$ F_2 \mapsto \Tree[ $\gamma$ $\delta$ ] \sqcup \Tree[ $\alpha$ $\beta$ ]  \otimes  \Tree[ $\eta$ $\xi$ ]\mapsto F \, . $$
  Thus, while at each step the extractions are smaller (and in themselves less costly), one needs a larger number of
  operations. Comparing the costs of these two derivations of $F$ from $T$ gives for the first one
 $1- \frac{4}{\ell(T)}=1/3$ and for the second $\fc=2-\frac{\ell(F_{\underline{v}})}{\ell(T)}-
 \frac{\ell(F_{\underline{w}})}{\ell(F_1)}=\frac{2 (\ell(T)-2)}{\ell(T)}=4/3 $, so the cost of the
 repeated application of SM is higher than the cost of the single application. In that sense the SM operations
 involving larger $T_v$ are more cost effective, as they make the overall derivation more efficient. 
 We will return to discuss this cost function and Minimal Search in \S \ref{HopfMArkovSec}.

\medskip
\subsection{Resource Restriction and Minimal Yield}\label{MYsec}

We now focus on the main constraints that will be relevant to our discussion.
The constraints of Minimal Yield discussed in \cite{ChomskyElements} 
(part of what is there more generally called Resource Restriction constraints)
aim at optimizing the Merge operators $\fM_{S,S'}$ in terms of their effect on
the size of workspaces. As discussed in \S 1.6 of \cite{MCB}, this can be 
measured by two counts: how the number of components of the workspace
is modified and how the number of accessible terms is modified.
The number of connected components of a workspace $F$ (notation $b_0(F)$, as it is also called
the zero-th Betti number by mathematicians) is the number of syntactic
objects  in the workspace (counting possible repetitions), or equivalently the
number of binary rooted trees $T_a$ in the forest $F=\sqcup_a T_a$. We denote by $\alpha(F)$
the number of accessible terms of the workspace $F$, by which we mean here
the sum $\alpha(F)=\sum_a \alpha(T_a)$ of the number of accessible terms of each components. 
Cancellations of deeper copies are no longer counted as accessible for the purpose of this
size counting. Also full components of the workspace are not counted as
accessible terms, so the size of the entire amount of computational material
available in the workspace is the sum $\sigma(F)=\alpha(F)+b_0(F)$, which is
also the total number of vertices in $F$ (not counting traces, that is, 
vertices corresponding to cancellations of deeper copies). 

The constraints on these measures of size express complementary optimality
requirements. We recall this definition from \cite{MCB}.

\begin{defn}\label{MYdef}
denoting by $\Phi: \fF_{\cS\cO_0} \to \fF_{\cS\cO_0}$ a given
transformation (in particular a Merge transformation $\Phi=\fM_{S,S'}$, we say that
it satisfies 
\begin{itemize}
\item {\em no divergence} if $b_0(\Phi(F)) \leq b_0(F)$, the number of
components is non-increasing, a condition that ensures that derivations
consisting of iterations of such transformations do not diverge;
\item {\em no information loss} if $\alpha(\Phi(F)) \geq \alpha(F)$, the
number of accessible terms is non-decreasing, namely no amount
of syntactic information is lost in the process.
\end{itemize}
\end{defn}

These two constraints are complementary to each other, and
tend to alter the total size $\sigma(F)=\alpha(F)+b_0(F)$ of the
computational material of the workspace in opposite directions,
so the minimal balance between them is expressed as the 
Minimal Yield condition that the total available
computational material is either stationary or
is minimally growing, namely $\sigma(\Phi(F))=\sigma(F)+1$.

It is shown in \S 1.6 of \cite{MCB} that EM and IM 
behave in the following way with
respect to the no divergence, no information loss, and
Minimal Yield constraints, with resulting change in sizes given by
\begin{center}
\begin{tabular}{| l | c ||r|r|r|}
\hline
{\em Type of Merge} &{\em Coproduct} & $b_0$ & $\alpha$ & $\sigma$  \\
\hline
External & $\Delta^c$ and $\Delta^d$ & $-1$ & $+2$ & $+1$  \\
\hline
Internal & $\Delta^c$ & $0$ & $+1$ & $+1$  \\
\hline
Internal & $\Delta^d$  & $0$ & $0$ & $0$  \\
\hline
\end{tabular}
\end{center}
where $\Delta^d$ and $\Delta^c$ are two different possible
forms of the coproduct (representing, respectively, the effect
of cancellation of the deeper copies at the SM and the CI
interfaces, see \S 1.3 of \cite{MCB}). The three forms of
SM, on the other hand produce the following change in the
size counting
\begin{center}
\begin{tabular}{| l  | c ||r|r|r|}
\hline
{\em Merge}  & {\em Coproduct} & $b_0$ & $\alpha$ & $\sigma$  \\
\hline
(1) &  $\Delta^c$ & $0$  &  $+1$  &   $+1$      \\
\hline
(1) & $\Delta^d$ &  $0$ & $0$  & $0$  \\
\hline
(2)  & $\Delta^c$ & $+1$  & $0$  &  $+1$     \\
\hline
(2)  &  $\Delta^d$ &  $+1$ & $-2$  & $-1$    \\
\hline
(3)  & $\Delta^c$ & $+1$   & $0$  & $+1$    \\
\hline
(3)  & $\Delta^d$ & $+1$  & $-2$  & $-1$     \\
\hline
\end{tabular}
\end{center}
We see from these tables of values that only EM, IM and
the form (1) of SM satisfy all three of the no divergence, 
no information loss, and Minimal Yield constraints, while
the remaining two forms of SM are excluded by these
constraints. The form (1) of SM is, from the point of view
of these counting, behaving exactly like an IM. 

\smallskip
\subsubsection{Variant of MY counting}

For comparison with the linguistics literature, it should be pointed
out that the counting for the Minimal Yield property used by
Chomsky in \cite{ChomskyGK} and in \cite{ChomskyUCLA} is
slightly different from the one here above. Indeed, in \S 4 (pp.~19--21)
of \cite{ChomskyGK} and pp.~38--39 and 47--48 of \cite{ChomskyUCLA},
the counting that is used is a slightly different combination of our
$b_0$ and $\alpha$, which is not the sum $\sigma=b_0+\alpha$ mentioned
here above, but rather the quantity that was called $\hat\sigma$ in the arXiv 
preprint \cite{MCB1}, which is given by $\hat\sigma= b_0 + \sigma$. This
form has the advantage of being more symmetric in EM and IM,
though it is less transparently interpretable in geometric terms:
while $b_0$ (connected components), $\alpha$ (non-root vertices), and
$\sigma$ (all vertices) have an immediate geometric interpretation
in terms of the tree structure, $\hat\sigma$ does not directly correspond
itself to a geometric quantity, so the formulation in \cite{MCB} was
finally expressed in terms of $\sigma$.  Since, given $b_0$, knowing
$\sigma$ or $\hat\sigma$ is equivalent, and in both cases SM is
disfavored compared to IM and EM, these formulations are
effectively equivalent, but use of $\hat\sigma$ instead of
$\sigma$ may be convenient in some cases.

\smallskip
\subsection{Resource Restriction and No Complexity Loss} \label{NCLsec}

There is an additional constraint considered in \S 1.6 of \cite{MCB},
which is referred to as ``no complexity loss" principle, and which
encodes the idea that Merge should build structures of increasing
complexity. Using the degree in the Hopf algebra (the number of
leaves, that is, the length of the sentence) as a measure, one
can state this principle as the requirement that for each component
$T_a$ of the workspace $F=\sqcup_a T_a$, the root vertex $v_a$
of the component $T_a$ becomes, after the transformation $\Phi$
a vertex of a component $T'_{a'}$ of the new workspace which
is of degree not smaller than that of $T_a$, namely
$\deg(T'_{a'})=\# L(T_{a'})\geq \#L(T_a)=\deg(T_a)$. This condition
is satisfied by EM and IM but fails for all forms of SM.

\smallskip

Note that, coming back to the formulation of the Extension Condition
and the comments immediately following Definition~\ref{ECdef} above, 
if we formulate EC as the combination of the two statements
\begin{enumerate}
\item  syntactic composition always grows the structure resulting in
more complex components
\item  this growth only happens at the
root of the tree(not by insertions at any lower vertices/edges),
\end{enumerate}
then the first statement is exactly formalized as the 
``no complexity loss" principle recalled here, while
the second statement is the formulation of EC
as given in Definition~\ref{ECdef}. To avoid confusion,
we will refer to the first statement as the NCL condition,
as described above, 
and to the second alone as the EC condition, consistently
with Definition~\ref{ECdef}, which we already
analyzed in \S \ref{ECdualHsec} above. 

It is preferable to maintain these two properties as distinct,
because of the different nature of the two constraints: the 
first one, the NCL principle, is a {\em soft constraint}, since
it is defined by an optimization property, while the second
property (the actual EC) is a {\em hard constraint} as it is 
directly imposed by the algebraic structure. 

\medskip
\subsection{Relation between Minimal Search and Resource Restriction}\label{MSRRrelSec}

Both Minimal Search and Resource Restriction (Minimal Yield and in the
formulation of \cite{MCB} also the No Complexity Loss principle discussed
above) have the effect of imposing soft optimality constraints on the action of Merge
on workspaces. It is sometimes unclear in the literature what exactly is the
relation between these two kinds of constraints: is there a redundancy or do they
actually play different roles, both of which are separately achieving different
forms of optimality?

\smallskip

The complementarity of MS and RR becomes more directly evident, if one
uses the cost function formulation that we described in Proposition~\ref{MSprop}.
External Merge and Internal Merge are optimal both with respect to MS and RR,
so in that respect these two kind of constraints achieve the same effect, but
what is more interesting is the comparative quantitative analysis of the
optimality violations for different forms of Sideward Merge, because there
the two types of constraints optimize in different ways, for different properties.

\smallskip

Consider again the simple example discussed in \S \ref{MSsev}, with the two
different Sideward Merge transformations
$$ T = \Tree[ [ $\eta$ [ [ $\alpha$ $\beta$ ] [ $\gamma$ $\delta$ ] ] ] $\xi$ ] \mapsto
 F = \Tree[ [ $\alpha$ $\beta$ ] [ $\gamma$ $\delta$ ] ] \sqcup  \Tree[ $\eta$ $\xi$ ] \, . $$
As we have seen in \S \ref{MSsev}, MS favors as less costly the shorter derivation
consisting of a single SM operation that extracts the two accessible terms
$$ T_v = \Tree[ $\alpha$ $\beta$ ] \ \  \text{ and }  \ \  T_w = \Tree[ $\gamma$ $\delta$ ] $$
and forms $\fM(T_v,T_w)$. On the other hand, as we will be discussing more at length in
\S \ref{MYSec}, Resource Restriction constraints favor the longer derivation consisting
of two SM operations and one EM operation, extracting and merging only atomic elements,
as these generate at each step of the derivation the least amount of RR violation.
In this very simple case one can see that one cannot simultaneously minimize both MS
and RR violations, but one can optimize for a combined cost. 

\smallskip

This shows that indeed MS and RR are not redundant constraints and they play
different roles. We will return in \S \ref{HopfMArkovSec} to discuss the effect on
the dynamics of Merge on workspaces of optimization with respect to one or
the other of these two constraints (or equivalently, the effect of optimality
violations in one or the other, or in both).

\smallskip
\section{Linguistic phenomena at near-violations of Resource Restriction} \label{MYSec}

Quantifying the optimality principles described above in terms of
size counting and grading makes it possible to 
characterize not only the operations that satisfy the
constraints but also the ``smallest possible" violations, namely  
the operations that are near-optimal but not optimal in terms of 
the same counting functions.

We focus here in particular on preserving the optimality of
no divergence, no information loss, and Minimal Yield constraints
and analyze which forms of type (1) SM (which we know
satisfy these constraints) are ``{\em as close as possible}"
to EM/IM.  

\subsection{Near-optimal Sideward Merge}\label{SMnearIM}
Given that in terms of the size counting type (1) SM behaves
like an IM, we want to find a good measure of how far the
different forms of SMs are from IM and which is the closest.

We know that type (1) SM violates the ``no complexity loss" 
principle mentioned above, so one first measure of proximity
to EM/IM would be minimization of the violation to this
constraint. 

As we recalled above, in type (1) SM the part of the workspace 
that is modified produces two components of the form 
$\fM(T_v, T')$ and $T/T_v$, replacing the original $T$ and $T'$,
with the rest of the workspace unchanged. The root of $T$
ends up as part of the structure $T/T_v$ and the root of $T'$
ends up as part of the structure $\fM(T_v, T')$. Simce
$\deg(\fM(T_v, T'))\geq \deg(T')$ the no complexity loss
is satisfied by the component $T'$, while $\deg(T/T_v)\leq \deg(T)$
so the component $T$ is where the violation of this constraint
occurs, and the amount of violation is $\deg(T)-\deg(T/T_v)=\deg(T_v)$.
This means that the smallest possible violation occurs when
$\deg(T_v)=1$, namely $T_v=\alpha$ is an ``atomic" structure
not further decomposable (a primitive term for the coproduct),
consisting of a single leaf with a label in $\cS\cO_0$ (a lexical item
with certain features). 

Thus, minimizing violation of the the ``no complexity loss" 
principle identifies SM cases of the form
$$ \fM(T', \alpha) \sqcup T/\alpha\, . $$
In more traditional linguistics notation, we can write these
in the form 
$$ [\, XP\,\,\,\, Y ] [ Z \, \ldots\,  \text{\sout{$Y$}} ] \, . $$

Note that by further application of EM (which satisfies all
the constraints) one obtains from this structures of the form
$$ \text{[ [ XP\,\, Y] [ Z\, $\ldots$\,  \text{\sout{$Y$}} ] ]} $$
or equivalently in our notation above
$$ \fM( \fM(T', \alpha), T/\alpha ) \, . $$
These cases can be seen as combinations of EM and SM
in which the SM part most closely resembles (in terms
of Resource Restrictions constraints) an IM operation.

We see then that this case (minimal violation of the NCL constraint)
identifies the class of Head-to-Phrase Movement discussed in \S \ref{HtoPsec},
in particular \eqref{8a}--\eqref{8b}, 
and includes also the particular case of 
Head-to-Head Movement discussed in \S \ref{HtoHsec} as in
\eqref{6a}--\eqref{6b}.

As we have discussed in \S \ref{ECsec} such Head-to-Phrase Movement
cases include {\em cliticization} (see \eqref{9ae}).
That such possibility of ``syntactic clitics" can occur as a minimal violation
of one of the constraints appears interesting in view of the fact that
other theories have typically appealed to forms of 
morphological ``readjustment rules" or to other post-syntactic 
devices.

\smallskip
\subsection{Near-optimal EM/SM combinations} \label{EMSMnearIM}

Next consider the case of composite operations. Compositions
of IM and EM are the optimal cases, and we want to identify
compositions involving SMs that are ``near-optimal" in terms
of constraints violations.

Consider the case of a composition EM$\circ$SM. 
Such a composition represents syntactically a form
of movement and as such it is natural to ask how far
it is, in terms of optimality constraints, from 
movement performed by Internal Merge.

\subsubsection{Resource Restrictions on EM$\circ$SM}
When we analyze this kind of compositions EM$\circ$SM 
from the point of view of the Resource Restrictions constraints,
we see from the tables above that, for the three cases of SM,
and the two forms $\Delta^d$ and $\Delta^c$, we obtain

\begin{center}
\begin{tabular}{| l  | c ||r|r|r|}
\hline
{\em Merge}  & {\em Coproduct} & $b_0$ & $\alpha$ & $\sigma$  \\
\hline
EM$\circ$SM(1) &  $\Delta^c$ & $ -1 $  &  $  +3$  &   $ +2 $      \\
\hline
EM$\circ$SM(1) & $\Delta^d$ &  $ -1 $ & $ +2 $  & $ +1 $  \\
\hline
EM$\circ$SM(2)  & $\Delta^c$ & $ 0 $  & $ +2 $  &  $ +2  $     \\
\hline
EM$\circ$SM(2)  &  $\Delta^d$ &  $ 0 $ & $ 0 $  & $ 0 $    \\
\hline
EM$\circ$SM(3)  & $\Delta^c$ & $ 0 $   & $ +2 $  & $ +2 $    \\
\hline
EM$\circ$SM(3)  & $\Delta^d$ & $ 0 $  & $ 0 $  & $ 0 $     \\
\hline
\end{tabular}
\end{center}
According to this table, the composition EM$\circ$SM(1),
for the first type of SM, namely
$$ \fM(\fM(T_v,T'),T/T_v)\, , $$
behaves exactly like a composition IM$\circ$EM in terms of
Resource Restrictions constraints. On the other hand, the
two compositions EM$\circ$SM(2) and EM$\circ$SM(3),
respectively of the form
$$ \fM(\fM(T_v,T'_w),T/T_v)  \ \ \ \text{ or } \ \ \  \fM(\fM(T_v,T'_w),T'/T'_w)  $$
and
$$ \fM(\fM(T_v,T_w),T/(T_v \sqcup T_w))\, , $$
behave like compositions IM$\circ$IM from the point of view of
the Resource Restrictions constraints. This means that the
point of view of the ``no divergence" and ``no information loss"
and Minimal Yield constraints, all the compositions EM$\circ$SM
are as good as compositions of IMs and EMs.  

However, if we consider the ``no complexity loss"
constraint and we impose that it holds at each step
in the composition (not only for the result of the
composite map) then the previous discussion limits
the form of SM involved to be a SM(1) with $T_v=\alpha$
an atomic structure, which involves a minimal
violation of this constraint. Note that both SM(2) and
SM(3), even with the restriction that the accessible
terms involved $T_v=\alpha$
and $T'_w=\beta$ (respectively, $T_v=\alpha$
and $T_w=\beta$) are both atomic, will still have
a larger violation of the ``no complexity loss" constraint
with respect to SM(1) with $T_v=\alpha$ atomic.

Thus, the cases of  EM$\circ$SM with minimal
violations of the Resource Restriction constraints
are again of the form 
$$ \fM( \fM(T', \alpha), T/\alpha ) \, .  $$

\subsubsection{Near-IM forms of EM$\circ$SM}

Next observe that the structure 
$\fM( \fM(T', \alpha), T/\alpha )$ as above
can itself be compared, in terms of Resource Restrictions constraints
to an IM. This is a different question from the one discussed
in \S \ref{SMnearIM} where one minimizes the discrepancy
between the SM operation and an IM: here one evaluates
the discrepancy between a composition EM$\circ$SM and IM. 
It is clear here that the discrepancy between the composition
of Merge operations $\fM( \fM(T', \alpha), T/\alpha )$ and
the Internal Merge $\fM(\alpha, T/\alpha)$ lies in the size
of the syntactic object $T'$. So one naturally expects that
the minimal violation should arise when $T'=\beta$ is also
an {\em atomic} structure, so that 
the more ``IM-like" movement that can be generated 
by a composition EM$\circ$SM with minimal constraints
violations is of the form
$$ \fM( \fM(\beta, \alpha), T/\alpha )\, ,  $$
where $\alpha$ and $\beta$ are leaves.

This can be formalized by describing more precisely what 
it means to be ``IM-like"  in the following way.
IM takes as input a syntactic object $T$ in a
certain degree $\ell =\# L(T)$ of the Hopf
algebra and gives a new syntactic object $\fM(T_v, T/T_v)$
in the same degree $\ell$ (using the quotient as in $\Delta^d$).
Similarly EM takes two syntactic objects $T,T'$ of
degrees $\ell, \ell'$ respectively, and returns a new
syntactic object $\fM(T,T')$ of degree $\ell+\ell'$, so
we can measure how ``IM-like" a transformation is
by how close it is to taking the syntactic object $T$ from
which the extraction of accessible terms is performed,
to an object of the same degree: the distance to IM
is here measured by the gap in degree. Thus, in
the case of the transformation
$\fM( \fM(T', \alpha), T/\alpha )$,
the gap in degree is $\deg(T')$ which is minimal
in the case
$\fM( \fM(\beta, \alpha), T/\alpha )$, 
where $\deg(\beta)=1$.
 
This type of Sideward Merge involving atomic elements
is exactly the one that we discussed in \S \ref{HtoHsec}
and describes the Head-to-Head movement case,
not realized in the form of \S \ref{HtoHsec} that violates the Extension Condition
but realized with the derivation described in \S \ref{AltSMsec} that uses
Sideward Merge followed by External Merge, 
where none of the steps violates EC. 

This shows that the form of Sideward Merge that allows for this Head-to-Head movement
is also the minimal constraints-violation operation involving SM and EM that
most closely resembles IM. 

Note that other compositions EM$\circ$SM(2) and EM$\circ$SM(3) where
both $T_v$ and $T'_w$ (respectively $T_v$ and $T_w$) is atomic would
also be near-IM in this degree-gap sense, but would have larger violations
of the NCL constraint so would not be preferable to
the EM$\circ$SM(1) of the form $\fM( \fM(\beta, \alpha), T/\alpha )$ that
model head movement. 
          
\smallskip          
      
Thus, we can conclude that one can look for near-IM compositions of
EM's and SMs, with the near-IM condition defined as above by the
degree gap. In a composition of the form
\begin{equation}\label{EMSMmin}
 \fM(\fM(T_v,T'),T/T_v)\, , 
\end{equation} 
allowing for larger distance from the IM case (larger
degree gap), while keeping the Resource
Restrictions constraints minimal or near-minimal 
corresponds to increasing the degree of $T'$ while fixing
$\deg(T_v)=1$, while maintaining the distance to IM minimal, $\deg(T')=1$,
and allowing for increasing violations of the Resource
Restrictions constraints (the ``no complexity loss" constraint
in this case) would lead to increasing values of $\deg(T_v)$.
These two cases correspond, respectively, to structures of
the form 
$$ \fM(\fM(T',\alpha), T/\alpha) $$
or equivalently in the notation we used in \S \ref{ECsec}, of the Head-to-Phrase form
$$ [ [ \,\, XP\,\,\,Y] [ Z\, \ldots\,  \text{\sout{$Y$}} ] ] \, , $$
and in the second case to structures of the form
$$ \fM(\fM(\beta, T_v), T/T_v)\, , $$
or equivalently of the Phrase-to-Head form
$$ [ [ \,\, X\,\,\, YP ] [ Z\, \ldots\,  \text{\sout{$YP$}} ] ] \, . $$
These are the .

\smallskip

\subsection{A hierarchy of constraint violations} 
Thus, we obtain a hierarchy of forms of EM$\circ$SM in terms of the
types of violations of the Resource Restrictions
and degree constraints:
\begin{enumerate}
\item {\em smallest overall violations:}  structures  \eqref{EMSMmin} of the form
$$ \fM( \fM(\beta, \alpha), T/\alpha ) $$
with both $T_v=\alpha \in \cS\cO_0$ and $T'=\beta \in \cS\cO_0$ atomic ({\em Head-to-Head movement});
\item {\em smallest Resource Restrictions violations, but larger degree violation}:
 structures  \eqref{EMSMmin} of the form
$$ \fM(\fM(T',\alpha), T/\alpha) $$
with $T_v=\alpha \in \cS\cO_0$ atomic and $T' \in \cS\cO$ an arbitrary syntactic object 
({\em Head-to-Phrase Movement});
\item  {\em smallest degree violation, but larger Resource
Restrictions violations}: 
structures  \eqref{EMSMmin} of the form
$$ \fM(\fM(\beta, T_v), T/T_v)\, , $$ 
with $T'=\beta \in \cS\cO_0$ atomic and $T_v$ not necessarily atomic ({\em Phrase-to-Head Movement});
\item {\em both large degree and Resource Restrictions violations}: structures  \eqref{EMSMmin}
where neither $T_v$ not $T'$ is atomic ({\em Phrase-to-Phrase Movement}). 
\end{enumerate}

\smallskip

It is clear that, among these, the first case has the least violations and
the fourth case the most. The two intermediate cases present violations
of a different nature: in the Resource Restrictions in the second case
and in the degree preservation in the third one. So a priori they are
not directly comparable in terms of which constitutes a ``worse" 
violation compared to the EM/IM case. 

One argument suggesting that the third case may be
regarded as a lesser violation than the second one is
the fact that the amount of possible violation in the third
case is bounded, as $\deg(T_v)\leq \deg(T)$, while
in the second case the amount of violation can be
arbitrarily large as $\deg(T')$ is unconstrained. On the
other hand, if we consider this as an operation for arbitrary
syntactic objects $T$, then $\deg(T)$ itself is also unconstrained,
so that the bound on $\deg(T_v)$ will also become arbitrary. 

One can instead argue that the constraints coming from
Resource Restrictions are more significant than the violations
of degree preservation. Indeed Resource Restrictions apply to
all Merge derivations and correspond to general economy
principles of the transformation of workspaces, hence they
are more fundamental, while degree preservation only
detects how ``similar to an IM" a certain combination of
EMs and SMs can be. In particular, EM itself does not
satisfy degree preservation, as it is additive on degrees.
Degree preservation simply means that the transformation
applied is a {\em movement} (like IM) rather than a {\em structure
building} operation (like EM), so this condition only describes 
to what extent a given Merge transformation involving EM and SM
can implement a form of movement. It is not surprising then
that it is satisfied by construction such as {\em predicate raising}
that can be seen as movement. 

Thus, the list above does indeed represent, in a precise way, an increasing
order of more severe violations of the optimality constraints
satisfied by EM and IM. 

\medskip

\section{Multiple wh-fronting, tucking in problem, and the SM formulation}\label{BulgSMsec}

In \S \ref{MWMsec} we discussed a possible approach to modeling
the multiple wh-movement and the associated tucking in phenomenon
without resorting to Sideward Merge, but replying instead on a different
proposed formulation of Minimalism through Chomsky's Box Theory.

We return now to compare that alternative solution
with the proposed one that uses Sideward Merge, which we
discuss here more explicitly. 
In the SM formulation as proposed in \S \ref{AltSMsec}, 
an SM operation produces in the workspace a structure of the form
$$ \Tree[ koj kakvo ] $$
which is then EM-merged to the remaining structure.
The tucking in problem, for why the form ``{\em koj kakvo e kupil}" 
is attested in Bulgarian but not a form with {\em kakvo koj}, becomes in this
approach part of the language-specific choice of {\em planarization}  
at Externalization for the resulting tree
$$ \Tree[ [ koj kakvo ] [ e kupil ] ] $$

The problem becomes more interesting when considering more than two
wh-movements. Cases with three wh-movements in Bulgarian were analyzed
in \cite{KrapCinq} and include for example:
\begin{itemize}
\item {\em Koj kogo kak e tselunai?} (who kissed whom how?)
\item {\em Koj k\^{a}de kolko e pohar\v{c}il?} (Who spent how much where?)
\item {\em Na kogo koga kak  \v{s}te pomogne\v{s}?} (Whom will you help when how?) 
\end{itemize}
On the basis of empirical evidence, in  \cite{KrapCinq} they propose a partial ordering for
wh-fronting in Bulgarian of the form
\begin{equation}\label{NwhBulg}
\text{{\em kogo} (whom)} > \text{ {\em na kogo} (to whom)} > \text{{\em koga} (when)}
\end{equation}
$$  > \text{{\em k\^{a}de} (where)} >
\begin{array}{l}
\text{{\em kakvo} (what)} \\
\text{{\em (na) kolko} (to how many)} 
\end{array}   > \text{{\em kak} (how)} \, . $$

A first observation: if we use SM as part of the Merge operations, there is an increasing
cost associated to more than two wh-frontings. For example, while {\em Koj kogo e tselunai?}
only involves a single SM which is of the lowest cost (as we will quantify in \S \ref{CostsSec} 
and \S \ref{MYSec}), passing to {\em Koj kogo kak e tselunai?} would require two SM
operations, one of which is of higher cost as it extracts accessible terms of the form
$$ \Tree[ koj kogo ] \sqcup \text{ kak } $$
and combines them into a structure
$$ \Tree[ [ koj kogo ] kak ] $$
that is then EM-ed to form the resulting structure ``{\em Koj kogo kak e tselunai?}".
Note that other choices of the two successive SM's can also planarize to give the
same ordering, for example
$$ \Tree[ kogo kak ] \sqcup \text{ koj } \longmapsto \Tree[ koj [ kogo kak ] ]  $$
On the other hand, other choices of successive SM would {\em not} give rise to the empirically observed
ordering. For example the tree resulting from
$$ \Tree[ koj kak ] \sqcup \text{ kogo } \longmapsto \Tree[ kogo [ koj kak ] ] $$
cannot be in any way planarized with an ordering of the three wh-frontings compatible with
\eqref{NwhBulg}.

Formulated in this way, the issue of the ordering of the wh-fronting appears to
be similar in nature to other planarization problems in Externalization (such as, for example,
the verb clusters in Germanic languages, discussed in terms of the mathematical formulation
of Minimalism in \cite{MHB}), where freely formed structures produced
by the free symmetric Merge are filtered in Externalization according to language-specific
parameters that filter out certain structures (that do not admit certain planarizations) and accept
others. This may explain the difference between Bulgarian and other Slavic
languages with respect to ordering of wh-fronting, in a similar way as one explains,
say, differences in the structure of verb clusters among Germanic languages.

On the other hand, the increasing cost of SM operations required to handle 
a larger number of wh-frontings should contribute to make the multiple wh-movement
an infrequent phenomenon, as one expects when increasingly large violations of optimality
are involved. 

A different issue, as mentioned above, arises with this approach based on 
a combination of SM and EM transformations, namely compatibility with the
coloring rules for the theta roles colored operad of \cite{MarLar} further discussed
in \cite{MHL}. 

\section{The role of SM in the dichotomy in semantics}\label{SMdisemSec}

First observe that, in the case of a single wh-movement, implemented by an 
Internal Merge operation, the movement is to a non-theta position. In terms of
the colored operad generators of \cite{MarLar} and \cite{MHL} this means that
IM, written as a composition $\fM_{T_v, T/T_v} \circ \fM_{T_v, 1}$ has the first
operation implemented by colored operad generators of the form
\begin{equation}\label{IMthetagen}
 \Tree[ .$\theta_0$ [ $c$ $(1,\theta_0)$ ] ]  
\end{equation} 
where $\theta_0$ is the color marker for a non-theta position, and $c$ is 
another arbitrary theta-color marker. As explained in \cite{MarLar} and \cite{MHL} in
this model the theta-role coloring remains attached to the trace of movement
(which is a reason why the trace is used for interpretation at the CI interface)
while the generator above means that the copy of the accessible term $T_v$
that $\fM_{T_v, 1}$ deposits in the workspace and that $\fM_{T_v, T/T_v}$ merges
with $T/T_v$ will have root labelled by the non-theta marker $\theta_0$
and not by its original theta-color $c$ that it had inside the structure $T$.

In order to accommodate a construction accounting for multiple wh-movement
in terms of Sideward Merge, we need to maintain the fact that such movement
is to a non-theta position. In the formulation in \cite{MarLar} and \cite{MHL}  of the
dichotomy in semantics between EM and IM (EM always creating theta positions
and IM always moving to non-theta position) the case of SM is dealt with by
the same coloring generators as EM (since the only generator that affects a 
color change to $\theta_0$ happens in an $\fM_{S,1}$ operation which can only
occur in IM). 

There are two ways to adapt the setting of \cite{MarLar} and \cite{MHL} to
accommodate SM to non-theta positions, as needed for instance for these
cases of multiple wh-movement. 

\medskip

\begin{enumerate}
\item One possibility is to add another generator to the colored operad of
theta roles of the form
\begin{equation}\label{extragenTheta}
  \Tree[ .$\theta_0$ [ $c$ $c'$ ] ]  
\end{equation}  
This choice has the apparent drawback that it can in principle be used also
by an EM, which would break the dichotomy requiring EM to create only
theta positions. This is not really the case, however. As discussed in \cite{MarLar}
the syntactic objects colored by the colored operad of theta roles should satisfy
a global conservation-law on theta roles (theta criterion), or else they are filtered
out as non-viable. In the formulation of \cite{MarLar}, all the generators except
\eqref{IMthetagen} already satisfy a local conservation law (theta roles that
flow into the vertex also flow out) and the global theta criterion ensures that
all the theta roles injected into the structure are also discharged. In the case of
the generator \eqref{IMthetagen}  the global conservation holds because
the theta role remains assigned at the trace position, so that global counting
of theta roles is unaffected. If we introduce a further generator of the form
\eqref{extragenTheta}, which is then further EM-merged to the rest of the
structure, the only way that the resulting object can satisfy the global conservation
law for theta roles is if both colors $c$ and $c'$ remain attached to traces:
this excludes that the generator \eqref{extragenTheta} can be used by an EM
operation (which creates no traces) and shows it can only occur in an SM.

\medskip

\item The other possibility to account for SM moving to a non-theta position
$\theta_0$ as needed for an SM-based explanation of multiple wh-movement,
is to maintain the same generators of the colored operad of theta roles
as in \cite{MarLar} and \cite{MHL}. This would maintain the fact that color-changes
to $\theta_0$ can only happen in an operation of the form $\fM_{S,1}$, which is
satisfactory in algebraic terms, as it singles out a particular role for the unit
element $1$ of the Hopf algebra of workspaces. This means that the two
different roles of SM (moving to a theta position like EM or to a non-theta
position like IM) are obtained by considering the second case that moves
to a non-theta position as a composition
\begin{equation}\label{SM11}
 \fM_{T_v, T_w} \circ \fM_{T_v,1}\circ \fM_{T_w,1} 
\end{equation}
where $\fM_{T_v,1}\circ \fM_{T_w,1}=\fM_{T_w,1}\circ \fM_{T_v,1}$ so
the order of this composition does not matter.  When considering
the corresponding colored operad generators for these three
operations, for the assignment of theta roles, we see that the first
two operations $\fM_{T_v,1}$ and $\fM_{T_w,1}$ have the effect 
of extracting the accessible terms $T_v$ and $T_w$ and placing
them in the workspace, but now with their root vertex marked as
a $\theta_0$ color. The additional SM operation $\fM_{T_v, T_w}$
will then necessarily be implemented by a colored operad generator
simply of the form
$$ \Tree[ .$\theta_0$ [ $\theta_0$ $\theta_0$ ] ] $$
At first it looks like this
choice has a possible drawback: we know by the cost counting that
the operations $\fM_{S,1}$ by themselves have non-optimal costs (which
disappears in the combination $\fM_{T_v, T/T_v} \circ \fM_{T_v, 1}$ 
that gives the optimal IM). So one can worry that, given an already
non-optimal SM operation $\fM_{T_v, T_w}$, a further composition
with other non-optimal operations $\fM_{T_v,1}\circ \fM_{T_w,1}$
will further rise the cost of this SM. However, this is not actually
the case. If we forget the theta-coloring, then we simply have
$$  \fM_{T_v, T_w} \circ \fM_{T_v,1}\circ \fM_{T_w,1} =  \fM_{T_v, T_w} $$
so the overall cost cannot exceed the one of the SM operation
$\fM_{T_v, T_w}$ involved (the cost problem only arises if the
operations $\fM_{S,1}$ occur in isolation).

\end{enumerate}

In either of these cases, the role of SM in the dichotomy in semantics
becomes more nuanced, in the sense that SM can act as either an
EM or an IM from the point of view of theta roles. 
We will see below that the second option described here appears more
natural in view of a unified view with the other cases (clitics and head-to-head movement)
that have an SM model.

\medskip

\section{EM$\circ$SM combination and filtering by theta-coloring and phases}\label{EMSMling}

We have seen above that the most optimal form of SM occurs when two
conditions are met:
\begin{itemize}
\item only atomic elements are involved in SM (single leaves)
\item SM is immediately followed by an EM operation
\end{itemize}

In particular, we argued that the EM$\circ$SM composition maximally
resembles an IM and is the closest in terms of costs counting.

This already rules out several possibilities that use an SM with only atomic
elements but not immediately followed by an EM. For example, consider the
following non-viable construction:
$$ \text{\em *John likes \text{\sout{ who }} after wondering who Bill hates \text{\sout{ who }} } $$
obtained as the result of the following Merge operations
$$ \text{ likes } \sqcup \text{ who } \sqcup \text{ wondering } \sqcup \text{ after } \sqcup \text{ Bill } \sqcup \text{ John } \sqcup \text{ hates } $$ $$ \stackrel{\text{\bf EM}}{\longrightarrow} \Tree[ likes who ] \sqcup \text{ wondering } \sqcup \text{ after } \sqcup \text{ Bill } \sqcup \text{ John } \sqcup \text{ hates } $$
$$ \stackrel{\text{\bf SM}}{\longrightarrow}  \Tree[ likes \text{\sout{ who }} ] \sqcup \Tree[ hates who ] \sqcup 
\text{ wondering } \sqcup \text{ after } \sqcup \text{ Bill } \sqcup \text{ John } 
$$ 
$$
\stackrel{\text{\bf EM}}{\longrightarrow}  \Tree[ likes \text{\sout{ who }} ]  \sqcup  \text{ wondering } \sqcup \text{ after }
\sqcup \text{ John }  \sqcup \Tree[ Bill [ hates who ] ] 
$$
$$
\stackrel{\text{\bf IM}}{\longrightarrow}  \Tree[ likes \text{\sout{ who }} ]  \sqcup  \text{ wondering } \sqcup \text{ after }
\sqcup \text{ John }  \sqcup \Tree[ who [ Bill [ hates \text{\sout{ who }} ] ] ]
$$
$$ 
\stackrel{\text{\bf EM$\circ$EM}}{\longrightarrow} \Tree[ likes \text{\sout{ who }} ] \sqcup \text{ John } \sqcup
\Tree[ [ after wondering ] [ who [ Bill [ hates \text{\sout{ who }} ] ] ] ] 
$$
$$ 
\stackrel{\text{\bf EM$\circ$EM}}{\longrightarrow} \Tree[ John [ [ likes \text{\sout{ who }} ] [ [ after wondering ] [ who [ Bill [ hates \text{\sout{ who }} ] ] ] ] ] ]
$$ 
We see that the problem here is that one cannot obtain the structure using an immediate EM merging 
of the substructure $\{ \text{ hates }, \text{ who } \}$ produced by SM with its own remainder 
part $\{ \text{ likes }, \text{\sout{ who }} \}$. This causes a higher cost of using SM compared to an EM$\circ$SM 
combination and a structure that is farther away from optimality and therefore less likely to be viable. 

\smallskip

However, even if we restrict to only the cases involving an EM$\circ$SM combination, 
not all the possible applications of operations of the form EM$\circ$SM
will pass the filtering either by theta roles or by well formed phases. A mathematical modeling
for these filtering systems in terms of colored operads and their bud generating systems 
has been described in \cite{MarLar} and in \cite{MHL}. 

We discuss in this section how these coloring rules restrict which EM$\circ$SM
combinations provide viable derivations. We will see that the set of generators
of the colored operads, which were chosen in \cite{MarLar} and in \cite{MHL} to
be the smallest set that accommodates the empirical data on theta roles and
the structure in phases of the Extended Projection, can be slightly adjusted to
also account for the type of movement analyzed in this paper that can be
modeled by a minimally-non-optimal Sideward Merge. As we have discussed
in the previous sections there are mainly three cases that admit a model via SM
with minimal optimality violations:
\begin{itemize}
\item Head-to-head movement
\item Multiple wh-movement
\item Clitics 
\end{itemize}

We will discuss here what coloring constraints by theta roles and
phases imply in each of these cases, and how those coloring rules
filter out non-viable structures that can be produced by SM in the free
action of Merge. 

\subsection{Filtering by theta roles}\label{ThetaEMSMsec}

While, as we showed above, 
theta roles compatibility for multiple wh-fronting can be satisfied, we can look at an
example like the following for a case where compatibility with theta roles does not hold:
consider the sentence ``{\em The guy looked at the bar}". We have a structure that
we can represent in the form $\{  {\rm NP} , \{ \text{-ed}, \{ v^*, \{ P, {\rm NP} \} \} \} \}$.
Suppose we then use a SM to extract and combine ``{\em The guy}" and ``{\em at the bar}"
and EM to merge it back into a single structure of the form
$$  \{ \{ {\rm NP} , \{ P, {\rm NP} \} \}  , \{ \text{-ed}, \{ v^*,  \text{\sout{ $\{ P, {\rm NP} \} $ }} \} \} $$
This would correspond to a non-viable derivation giving:  ``{\em The guy at the bar looked \text{\sout{ at the bar }}}".
This structure can be ruled out on the ground of either theta or phase coloring. We will illustrate here the
theta-coloring problem and we will later consider a similar example that fails  because of phase-coloring.

A way to see, in the formalism of generators for the colored operad of theta roles,
what goes wrong for this example, is to view the structure in the following way
\begin{equation}\label{nocolorTheta1}
 \Tree[ \qroof{(The guy, $\theta_E^\downarrow$)}.$\theta_E^\downarrow$ [ {$(\text{looked}, (\theta_E^\uparrow, \theta_I^\uparrow))$} 
\qroof{ (at the bar, $\theta_I^\downarrow$) }.{$\theta_I^\downarrow$} ] ] 
\end{equation}
The SM action has only two possibilities for theta roles. in the first one it can create a structure like
$$ \Tree[ .$\theta_E^\downarrow$ \qroof{(The guy, $\theta_E^\downarrow$)}.$\theta_E^\downarrow$ \qroof{ (at the bar, $\theta_I^\downarrow$) }.{$\theta_I^\downarrow$} ] $$
Thus structure is not acceptable as there is no colored operad generator for theta roles of the unbalanced form
$$ \Tree[ .$\theta_E^\downarrow$ $\theta_E^\downarrow$ $\theta_I^\downarrow$ ] $$
where there is no conservation at the vertex between what flows in and out. 
Note also that if this structure contains ``{\em at the bar}" in the  theta position $\theta_I^\downarrow$, that theta role
$\theta_I^\uparrow$ would have to be assigned by ``{\em The guy}" rather than by ``{\em looked}" and there is 
no combination of the colored operad generators that could change, in this SM movement, the theta-color of
``{\em The guy}".  In particular, with respect to ``{\em The guy}" the moved ``{\em at the bar}" would be in an
adjunct position (which is a non-mandatory relation) and not in a mandatory theta-role relation. The
theta-coloring generators have the possibility of non-mandatory adjunct positions, marked as non-theta position
(discussed below). 
This causes a mismatch in the counting of theta roles, as the trace of the movement  would
also retain a $\theta_I^\downarrow$ labeling that would then be overcounted, compared to the $\theta_I^\uparrow$'s. 

This reasoning about theta-coloring generators, that excludes a colored structure of the form \eqref{nocolorTheta1},
shows that our system of theta-coloring generators also implements the linguistic rule that 
one cannot insert a theta-marked argument into an adjunct position.

The second possibility would be that 
(by first acting on $\fM_{T_v,1}$ and $\fM_{T_w,1}$ as in the case of multiple wh-fronting)
SM creates a structure of the form
$$  \Tree[ .$\theta_E^\downarrow$ \qroof{(The guy, $\theta_E^\downarrow$)}.$\theta_E^\downarrow$ [ .{$\theta_0$}
$(1,\theta_0)$ \qroof{ (at the bar, $\theta_I^\downarrow$) }.{$\theta_I^\downarrow$} ] ] $$
Note that the construction would be asymmetric, as both ``{\em The guy}" and ``{\em at the bar}"
are extracted by SM but only the second is extracted via an $\fM_{T_v,1}$  that changes the theta-color to $\theta_0$.
In principle, one can consider such operations of the form $\fM_{T_w, T_v} \circ \fM_{T_v,1}$ where
$T_v$ changes its theta role to $\theta_0$ but $T_w$ retains its original theta-role, hence carried also
by the root of the resulting $\fM(T_w,T_v)$.  However, the only way to interpret in theta theory such a
theta-colored structure would be if the $\theta_0$-colored ``{\em at the bar}" has a non-theta position in
the sense of a non-mandatory (adjunct) relation with ``{\em The guy}", which is not a mandatory theta-role (unlike in its
role as $\theta_I^\downarrow$-role for ``{\em look}", which is a mandatory theta role, as we observed above. 

As discussed in \cite{MarLar}, one should really distinguish two different kinds of non-theta positions, say $\theta_0$ and
$\theta_0'$ where the latter indicates non-mandatory roles that are still part of the argument
structure in the propositional domain, while $\theta_0$ marks the positions that belong in the clausal domain
(like in the case of IM movement). This is indeed an example where this distinction is important. 
The generator \eqref{IMthetagen}
of the colored operad that is associated to the $\fM_{T_v,1}$ transformation only produces this 
$\theta_0$-type non-theta positions, not the $\theta'_0$ type, while EM can access generators for both forms
$$ \Tree[ .$c$ [ $c$ $\theta'_0$ ] ]   \ \ \ \text{ and } \ \ \  \Tree[ .$c$ [ $c$ $\theta_0$ ] ]  $$
hence EM$\circ$SM is not a viable derivation for ``{\em The guy at the bar looked}".

Thus, the incompatibility with theta-coloring, in this example, arises from the different status of the
moved term `{\em at the bar}", with a mandatory theta-role assignment at its original position and
an adjunct role at its moved position. We will see in \S \ref{PhaseEMSMsec} that similar examples
where this incompatibility is resolved (i.e.~both positions are adjunct or both positions are the
same kind of mandatory theta-role) can still fail to pass the other coloring filter, by phase-coloring.

\smallskip
\subsection{Filtering by phases}\label{PhaseEMSMsec}

Consider then the similar example 
$$ \text{\em *The main at the bar saw the woman \text{\sout{ at the bar }} } $$
This is similar to the example we discussed above, except that now there is no
mismatch of theta roles, because ``{\em at the bar}" is in an adjunct role in both
positions. 

However, we see that, if we want to realize this structure via a combination EM$\circ$SM,
with EM acting by merging the structure created by SM with its remainder part, this would
violate the constraints coming from the structure of phases. 

Indeed, if we first build by EM the structure
$$ \Tree[ saw [ [ the woman ] \qroof{at the bar}. ] ] $$
and we then act via SM to form
$$ \Tree[ \qroof{the man}. \qroof{at the bar}. ] \sqcup \Tree[ saw [ [ the woman ] \qroof{\text{\sout{ at the bar } }}. ] ] $$
followed by the EM that merges these two syntactic objects, we have the problem that the SM is 
extracting from the interior of a phase {\em after} the phase $v^*$P has already been constructed. 

When movement is realized by Internal Merge, movement across the phase edge is 
prevented (phase impenetrability). In \cite{MHL} we showed that a suitable choice of colored operad
generators suffices to implement this constraint on movement by IM.

In \cite{MHL} we did not discuss how to extend the same constraint to movement realized by SM
rather than IM. So in this section we show how the choice of the generators of the colored operad
of phases can be extended to include the compatibility, for all the cases of SM that are relevant,
of movement by SM and the structure of phases. 

In particular, this example will then be ruled out by the phase-coloring. A similar example,
in which the element moved by SM is in a theta position and maintains the same theta role
at both its initial and its moved position, is given by 
$$ \text{\em *Constructing the building is quite different from planning \text{\sout{ the building }}     } $$

In this case also one sees that, in order to produce this sentence via an EM$\circ$SM that
merges the component created by SM with its corresponding remainder part, the SM would have
to move across a phase boundary. Indeed, in this case there are two $v^*$P phases. If one 
applies SM to the already formed ``{\rm is quite different from planning the building}" 
to give ``{\em Constructing the building}" and ``{\rm is quite different from planning \text{\sout{ the building }}}"
and then merges them with EM one would have SM acting across the phase boundary, as in the
previous example. In order to avoid that, one would have to apply SM before the $v^*$P phases
are constructedm meaning creating ``{\em Constructing the building}" and ``{\em planning \text{\sout{ the building }}}"
before EM proceeds to form ``{\rm is quite different from planning \text{\sout{ the building }}}" to which
the structure formed by SM is finally merges, so it would not be an EM$\circ$SM operation
of the type considered above. 

\smallskip
\subsection{Bud generators and phase-coloring of SM: multiple wh-fronting}\label{SMphasecolSec}

In order to incorporate these constraints on phases in the structure building where SM
is involved, we consider the same setting described in \cite{MHL}, where the structure
of phases and the Extended Projection, and the associated rules for movement, are
encoded in a finite set of generators for a colored operad. Syntactic objects that 
can be colored according to the rules set by this system of generators (hence that
are in the associated language of the colored operad, in the sense introduced in \cite{Gira})
have well formed phases and are obtained through derivations that follow the rules of
structure formation and movement with respect to phases.

\smallskip

In the formulation of \cite{MHL}, for the colored operad of phases, 
movement realized by Internal Merge corresponds to a generator of the form
\begin{equation}\label{genIMphase}
\Tree[ .$\fs^\downarrow_{\omega'}$ [ {$\fc_\omega$} {$(1,\fm)$} ] ] 
\end{equation}
where the coloring $\fs^\downarrow_{\omega'}$
at the root indicates that the target location of the movement by IM is the edge of the phase
determined by the head $\omega'$, and 
$\fc_\omega$ is the color, for a lower head $\omega$ (within the same phase), 
at the previous location where the accessible term is extracted by IM. 

A first observation is that we want SM to realize the case of multiple wh-movement,
compatibly with the coloring by theta-roles discussed in \S  \S \ref{SMdisemSec} and 
\S \ref{ThetaEMSMsec} (see \cite{MHL} 
for a discussion of the compatibility between phase-coloring and theta-coloring). We have
already seen, in discussing theta-coloring in \S \ref{SMdisemSec}, that in order to
realize multiple wh-movement by SM one should think (for the purpose of theta-coloring)
of the SM transformation as a composition $\fM_{T_w,T_v} \circ \fM_{T_w,1} \circ \fM_{T_v, 1}$.
This should also fit the coloring by phases, since as argued in \cite{MHL} these two
systems of coloring should be combined compatibly. 

However, while for theta-coloring there is a generator with only non-theta positions $\theta_0$
that can be used for combining the two generators 
$$ \Tree[ .$\theta_0$ [  $\fc$  {$(1,\theta_0)$} ] ] $$
of $\fM_{T_w,1}$ and $\fM_{T_v, 1}$,  for phase-coloring we do not have, among the 
generators with an edge-of-phase position $\fs^\downarrow_{\omega'}$ at the root, one
that can also input the same $\fs^\downarrow_{\omega'}$ at the leaves, as one would need
for coloring the merge-generated nodes of $\fM_{T_w,T_v} \circ \fM_{T_w,1} \circ \fM_{T_v, 1}$ as
\begin{equation}\label{largegenEMSMphase}
\Tree[ .$\fs^\downarrow_{\omega'}$ [ [ .$\fs^\downarrow_{\omega'}$ [ {$\fc_{\omega_1}$} {$(1,\fm)$} ] ] [ .$\fs^\downarrow_{\omega'}$ [ {$\fc'_{\omega_2}$} {$(1,\fm)$} ] ] ] ] 
\end{equation}
The problem with introducing generators of the form
\begin{equation}\label{sssgen}
 \Tree[ .$\fs^\downarrow_\omega$ [ $\fs^\downarrow_\omega$ $\fs^\downarrow_\omega$ ] ]  
\end{equation} 
that ``split" the edge of phase is that these should only apply in the combination
EM$\circ$SM, as a single generator of the form \eqref{largegenEMSMphase} 
and not as separate generators \eqref{sssgen} usable by EM (as that would not be
compatible with the behavior of EM with respect to phases). Introducing a composite
generator like \eqref{largegenEMSMphase} has interesting consequences: it can
be viewed as saying that, for the purpose of structure formation by phases, only
EM$\circ$SM is a fundamental operation, not SM on its own (while of course EM
on its own is allowed). As discussed in \cite{MarSkig}, there are other advantages,
in terms of properties of Merge as a dynamical system for considering only EM$\circ$SM
rather than SM as one of the operations (along with SM and IM). 

We explain below the advantages and drawbacks of both hypothesis: an additional simple generator
of the form \eqref{sssgen}, or an additional more complex generator of the form \eqref{largegenEMSMphase}.

\subsubsection{Additional edge-splitting generator}\label{gensssSec}

We first analyze the possibility of adding a generator of the form \eqref{sssgen}, which has
the advantage of maintaining all the coloring generators in the simplest form with only one
vertex. As we explained in \cite{MHL}, this allows for an equivalence between the ``filtering"
viewpoint (structure building is unconstrained and realized by the free symmetric Merge
and these freely formed structures are filtered for theta roles and well-structured phases)
and the ``constrained Merge" viewpoint (constraints on movement and phases and
matching theta-roles are part of the structure building process of Merge). These two
complementary viewpoints are entirely equivalent, but only if all the coloring generators
are of the simplest form consisting of only one vertex. Thus, this is a clear advantage of
adding a generator of the form \eqref{sssgen}, to account for multiple wh-movement.

The drawback of this choice is that such generator should {\em only} occur in SM and
never in EM. We discuss how one can ensure that, both from the colored operad
point of view and from its complementary ``colored Merge" viewpoint as described in \cite{MHL},
so that the equivalence between these two formulations is maintained. 
In the colored operad formulation, this would require that the $\fs_\omega^\downarrow$ positions
be marked in two different ways, according to whether they correspond to an EM or an SM (or IM).
Note that adding a non-theta position $\theta_0$-marker does not serve this purpose, as some parts
of the Extended Projection built by EM do not carry theta roles. This would mean that, in addition
to the generators 
\begin{equation}\label{gen0}
  \Tree[ .$\fs_\omega^\downarrow$ [ {$(\fh^{\omega'}, \fs_{\omega'}^\uparrow)$} $\fs_{\omega'}^\downarrow$ ] ] 
\end{equation}  
as in \cite{MHL}, that are used by EM, one includes additional generators with a modified coloring
$\hat \fs_{\omega'}^\downarrow$ (that implements this splitting of the $\fs_{\omega'}^\downarrow$ position,
namely generators similar to the IM generator \eqref{genIMphase}, of the form
\begin{equation}\label{gensstar2}
\Tree[ .$\hat\fs^\downarrow_{\omega'}$ [ {$\fc_\omega$} {$(1,\fm)$} ] ] 
\end{equation}
and generators that build and terminate clustering at this edge position
\begin{equation}\label{gensstar3}
\Tree[ .$\hat\fs^\downarrow_{\omega'}$ [ {$\hat\fs^\downarrow_{\omega'}$} {$\hat\fs^\downarrow_{\omega'}$} ] ] 
\end{equation}
that splits the edge-of-phase for SM multiple wh-movement
\begin{equation}\label{gensstar3b}
\Tree[ .$\fs^\downarrow_{\omega'}$ [ {$\hat\fs^\downarrow_{\omega'}$} {$\hat\fs^\downarrow_{\omega'}$} ] ] 
\end{equation}
that terminates this splitting and connects back to the generator \eqref{gen0}.
In this way, the generator \eqref{gensstar3} cannot be accessed by an EM but can be accessed by an SM 
of the form $\fM_{T_w,T_v} \circ \fM_{T_w,1} \circ \fM_{T_v, 1}$ that forms a composition like \eqref{largegenEMSMphase}
(which is in this case not a generator but a composition of generators). 

In the case of more than two wh-frontings (for example the cases with
three wh-frontings discussed in \S \ref{BulgSMsec}), with this formulation we
would simply have compositions of multiple \eqref{gensstar3}, such as
\begin{equation}\label{triplesplit}
 \Tree[ .$\fs^\downarrow_{\omega'}$ [ [ .$\hat\fs^\downarrow_{\omega'}$
[ {$\hat\fs^\downarrow_{\omega'}$} {$\hat\fs^\downarrow_{\omega'}$} ] ]
$\hat\fs^\downarrow_{\omega'}$ ]  ]
\end{equation}
that then combine with three generators of the form \eqref{gensstar2}. 

Since all the generators here remain of the same simple form with only one vertex, the coloring
implemented by the colored operad (and the corresponding filtering of freely formed
structures according to whether they are in the language of the colored operad) is still
equivalent to the ``colored Merge" formulation where the free Merge action by
\begin{equation}\label{KmergeHMC}
 \cK = \sum_{S,S'} \fM_{S,S'} = \sum_{S,S'} \sqcup \circ (\cB\otimes {\rm id}) \circ \delta_{S,S'} \circ \Delta 
\end{equation} 
is replaced by the ``colored" Merge action by
$$ \cK_\Phi = \sum_{S,S'\in \cL(\bB_\Phi)} \sum_{c\in \Omega} \fM^c_{S,S'} $$
with $\cL(\bB_\Phi)$ the language of the colored operad of phases, $\Omega$ its set of colors, and
$$ \fM^c_{S,S'} = \sqcup \circ (\cB^c \otimes {\rm id}) \circ \delta^c_{c_S,c'_{S'}} \delta_{S,S'} \circ \Delta $$
with $c_S,c'_{S'}$ the root colors of $S,S'\in \cL(\bB_\Phi)$ and 
\begin{equation}\label{deltacolors}
 \delta^c_{c_S,c'_{S'}}=\left\{ \begin{array}{ll} 1 & \text{if } {\scriptsize  \Tree[ .$c$ [  $c_S$  $c'_{S'}$ ] ]  }  \in \cR_\Phi \\[10mm]
0 & \text{otherwise,} \end{array} \right. 
\end{equation}
with $\cR_\Phi$ the set of generators of the colored operad of phases. 

In this form, the terms of the coproduct that will be selected by 
$$ \delta^{\hat\fs^\downarrow_\omega}_{\hat\fs^\downarrow_\omega,\hat\fs^\downarrow_\omega} \ \ \text{ or } \ \ 
\delta^{\fs^\downarrow_\omega}_{\hat\fs^\downarrow_\omega,\hat\fs^\downarrow_\omega} $$
will be either colored structures $S, S'$ with root coloring of the form \eqref{gensstar2} (a double 
wh-movement realized by SM) or structures $S, S'$ with some combination of generators \eqref{gensstar3}
at the top of the tree followed by generators \eqref{gensstar2}  below (and other
colored structures below the vertices marked with $\fc_{\omega_1}$ and $\fc_{\omega_2}$), which
would realize multiple wh-movement.

\smallskip
 \subsubsection{Composite generator}\label{gensssLargeSec}

The second possibility, to incorporate phase-coloring for multiple wh-movement via SM,  is to
use a colored operad generator that is no longer of the simplest form with a single vertex, but
is instead of the composite form \eqref{largegenEMSMphase}. In this case, we do not need
to introduce two different colors $\fs$ and $\hat\fs$ for spec-positions, since the colored 
single vertex structure \eqref{gensstar2}  no longer occurs on its own but always only in
the combination  \eqref{largegenEMSMphase}.

On the other hand, this choice has its own drawbacks. One is that it makes the derivation of
multiple (more than two) wh-fronting more cumbersome, and the other is that it is no longer
directly equivalent to a colored Merge. There is, however, a way to restore the equivalence
between free structure building followed by color-filtering and structure building constrained
by coloring, is one considers that SM is not an operation occurring on its own but always
only in the combination EM$\circ$SM that EM-merges the component created by SM with
its complementary one. We discuss briefly both of these aspects. 

In the case of more than two wh-frontings (as in \S \ref{BulgSMsec}) one would have to
first perform a double wh-fronting with a generator of the form  \eqref{largegenEMSMphase}
implementing the composition $\fM_{T_w,T_v} \circ \fM_{T_w,1} \circ \fM_{T_v, 1}$. Since
now there is no way to produce a multiple splitting of the edge-of-phase position like \eqref{triplesplit},
the multiple case will have to use SM to extract again the structure  \eqref{largegenEMSMphase} and
combine it with the third wh-fronting via another  \eqref{largegenEMSMphase} generator, creating
a more complex structure of the form 
$$
\Tree[ .$\fs^\downarrow_{\omega'}$ [ [ .$\fs^\downarrow_{\omega'}$ [ [ .$\fs^\downarrow_{\omega'}$ [ [ .$\fs^\downarrow_{\omega'}$ [ {$\fc_{\omega_1}$} {$(1,\fm)$} ] ] [ .$\fs^\downarrow_{\omega'}$ [ {$\fc'_{\omega_2}$} {$(1,\fm)$} ] ] ] ] 
 {$(1,\fm)$} ] ] [ .$\fs^\downarrow_{\omega'}$ [ {$\fc'_{\omega_3}$} {$(1,\fm)$} ] ] ] ] 
$$
instead of a structure 
$$ 
 \Tree[ .$\hat\fs^\downarrow_{\omega'}$ [ .$\hat\fs^\downarrow_{\omega'}$
[ [ .$\hat\fs^\downarrow_{\omega'}$ [ {$\fc_{\omega_1}$} {$(1,\fm)$} ] ] [ .$\hat\fs^\downarrow_{\omega'}$ [ {$\fc_{\omega_2}$} {$(1,\fm)$} ] ]  ] ]
[ .$\hat\fs^\downarrow_{\omega'}$ [ {$\fc_{\omega_3}$} {$(1,\fm)$} ] ]     ] 
$$ 
as would be obtained by the previous method discussed in \S \ref{gensssSec}. Multiple wh-fronting
can still be obtained, only with a slightly more complicated form of the phase-coloring. 

The issue of the equivalence between filtering freely formed structures by colored operads or
forming structures by a constrained colored Merge is more interesting here. Because in this
formulation not all coloring generators are single-vertex, the coloring cannot be implemented as
constraints on the Merge action of the form \eqref{deltacolors}
as above (as in the formulation of \cite{MHL}). Thus, the only way that the filtering can
be seen as equivalent to a constrained structure building is if SM can only occur
in structure building in the form EM$\circ$SM rather than SM alone. This would mean
replacing the simple unifying form \eqref{KmergeHMC} of the Merge action, which is
in a transparent way a case of the general theory of Hopf algebra Markov chains (see 
\cite{DiPaRa}, \cite{Pang}, \cite{MarSkig}) with a more complicated and less transparent
form. This would still give rise to a Markov chain on the space of workspaces, as analyzed
in \cite{MarSkig}, but with a less interesting behavior as a dynamical system. (This will
be discussed in separate work.) 

\smallskip

Given this direct comparison between the formal  properties of both proposed
solutions, it seems that the one described in \S \ref{gensssSec} is more natural and
more closely fitting the overall algebraic structure of Merge. 

\smallskip

We then analyze the other cases of SM (head-to-head movement and clitics), to
see how those fit into the constraints of coloring by phase-structure.

\smallskip
\subsection{Bud generators and phase-coloring of SM: head-to-head}\label{SMphasecolSec2}

In the case of head-to-head movement, we have a combination EM$\circ$SM of the form
$$ \beta \sqcup T \stackrel{\text{\bf SM}}{\longrightarrow} \Tree[ $\beta$ $\alpha$ ] \sqcup T/^c \alpha 
 \stackrel{\text{\bf EM}}{\longrightarrow} \Tree[ [ $\beta$ $\alpha$ ] {$T/^c \alpha$} ] $$
that replaces a structure of the form
$$  \beta \sqcup T \stackrel{\text{\bf EM}}{\longrightarrow}  \Tree[ $\beta$ $T$ ] $$
with the head-to-head movement of $\alpha$ (extracted from $T$)

First observe that we can assume here that the structure $\fM(\beta,T)$ is well formed
and therefore is both in the language $\cL(\bB_\Phi)$ for the colored-operad of phases
and in the language $\cL(\bB_\Theta)$ for the colored operad of theta-roles. For the  
colored operad of phases, this means that $\beta$ should be colored by either
$(\fh^\omega, \fz^\uparrow_\omega)$ or $(\fh^\omega, \fs^\uparrow_\omega)$ or
$(\fh^\omega, \fz^\uparrow_\omega, \fs^\uparrow_\omega)$ and the root vertex of $T$ is correspondingly colored
either $\fz^\downarrow_\omega$ or $\fs^\downarrow_\omega$ or $(\fs_{\omega'}^\uparrow, \fz^\downarrow_\omega)$,
according to the coloring-generators of \cite{MHL}. 

This means that the structure $\fM(\beta,\alpha)$ created by SM must be colored consistently with these
existing constraints. First we can assume that, when $\alpha$ is extracted from $T$, it gets a coloring
(similarly to the case of IM extractions or of the multiple wh-movement SM extraction discussed above) as
part of some structure (to be determined) of the form
\begin{equation}\label{HtoHgen1}
 \Tree[ .{$\hat\fm$} [ {$(\alpha, \fc)$} {$(1,\fm)$} ] ]
\end{equation} 
where $\fc$ is the coloring of $\alpha$ inside the syntactic object $T\in \cL(\bB_\Phi)$, and where $\hat\fm$
is an additional color that is needed to provide consistent phase coloring
for the head-to-head movement produced by SM but is not used by IM. This should then be combined with
a generator
\begin{equation}\label{HtoHgen2}
 \Tree[ .{$(\fh^\omega,\fc_\omega^\uparrow)$} [ \text{$\hat\fm$  } {$(\fh^\omega,\fc_\omega^\uparrow)$} ] ] 
\end{equation} 
where $\hat\fm$ is the same color and $(\fh^\omega,\fc_\omega^\uparrow)$ can be any of the colors 
$(\fh^\omega, \fz^\uparrow_\omega)$ or $(\fh^\omega, \fs^\uparrow_\omega)$ or
$(\fh^\omega, \fz^\uparrow_\omega, \fs^\uparrow_\omega)$. Moreover, the color $\hat\fm$
should not occur in any other generator of the colored operad. We use the notation $\hat\fm$ simply
because the second generator \eqref{HtoHgen2} behaves like the case of a modifier color $\fm$ that
lets the other color (here $(\fh^\omega,\fc_\omega^\uparrow)$) go through unchanged. This notation
does not mean, however, that the generator \eqref{HtoHgen1} produces a modifier of the head
$\fh^\omega$, only that it behaves like one for the purpose of coloring rules. 

Again the same argument given in \S \ref{gensssSec}  shows that, with these additional
generators \eqref{HtoHgen1} and \eqref{HtoHgen2} the coloring by filtering freely
formed structures according to whether they can be colored as elements of $\cL(\bB_\Phi)$
is equivalent to the colored Merge formulation with constrained structure building. This
depends on the fact that all the generators remain of the simple form with a single vertex,
as in \cite{MHL}. In order to ensure compatible theta-coloring for these additional
generators it suffices to require that the addditional color $\hat\fm$ introduced here
always comes paired with a theta-coloring $\theta_0$.

The last case that we need to account for, in terms of phenomena that can be
modeled by a SM with minimal violations of optimality, in a combination EM$\circ$SM, is
the case of clitics.

\smallskip
\subsection{Bud generators and phase-coloring of SM: clitics}\label{SMphasecolSec3}

The main delicate issue, in the case of clitics, is the insertion in the Extended Projection
so that it does not require movement across phase boundaries. More precisely, consider
the example ``{\em Pietro l'ama}". Head-to-head movement via SM that raises the clitic
CL to INFL is not compatible with the structure of phases because INFL and CL
belong to two different phases in the structure
$$ \{ {\rm INFL}, \{ {\rm EA}, \{ v^* , \{ {\rm Root}, {\rm CL} \} \} \} \} \, . $$
We would see here a chain of Merge operations of the form
$$  \{ {\rm Root}, {\rm IA} \} \}  \sqcup v^* \sqcup {\rm EA} \sqcup {\rm INFL} \stackrel{\rm SM}{\longrightarrow}
\{ v^*, {\rm Root} \} \sqcup \{ \text{\sout{ Root }}, {\rm IA} \} \}  \sqcup v^* \sqcup {\rm EA} \sqcup {\rm INFL} $$
$$  \stackrel{\rm EM}{\longrightarrow} \{ {\rm EA}, \{ \{ v^*, {\rm Root} \} , \{ \text{\sout{ Root }}, {\rm IA} \} \} \} \sqcup {\rm INFL}  $$
$$ \stackrel{\rm SM}{\longrightarrow} \{ {\rm INFL}, \{ v^*, {\rm Root} \}  \} \sqcup 
\{ {\rm EA}, \{ \text{\sout{ $\{ v^*, {\rm Root} \}$ }} , \{ \text{\sout{ Root }}, {\rm IA} \} \} \} $$
$$ \stackrel{\rm SM}{\longrightarrow} \{ \{ {\rm INFL}, \{ v^*, {\rm Root} \}  \}, {\rm CL} \} \sqcup 
\{ {\rm EA}, \{ \text{\sout{ $\{ v^*, {\rm Root} \}$ }} , \{ \text{\sout{ Root }}, \text{\sout{  IA }} \} \} \} $$
$$ \stackrel{\rm EM}{\longrightarrow} \{ \{ \{ {\rm INFL}, \{ v^*, {\rm Root} \}  \}, {\rm CL} \} , \{ {\rm EA}, \{ \text{\sout{ $\{ v^*, {\rm Root} \}$ }} , \{ \text{\sout{ Root }}, \text{\sout{  IA }} \} \} \} \} $$
$$ \stackrel{\rm IM}{\longrightarrow} \{ {\rm EA},  \{ \{ {\rm INFL}, \{ v^*, {\rm Root} \}  \}, {\rm CL} \}  , \{ \text{\sout{ EA }}, \{ \text{\sout{ $\{ v^*, {\rm Root} \}$ }} , \{ \text{\sout{ Root }}, \text{\sout{  IA }} \} \} \} \} \} \, . $$
The structure $ \{ \{ {\rm INFL}, \{ v^*, {\rm Root} \}  \}, {\rm CL} \}$ would then externalize to ``{\em l'ama}".

The problem here are the SM transformations that form the workspace
$$ \{ {\rm INFL}, \{ v^*, {\rm Root} \}  \} \sqcup \{ {\rm EA}, \{ \text{\sout{ $\{ v^*, {\rm Root} \}$ }} , 
\{ \text{\sout{ Root }}, {\rm IA} \} \} \} $$
and then the workspace
$$ \{ \{ {\rm INFL}, \{ v^*, {\rm Root} \}  \}, {\rm CL} \} \sqcup 
\{ {\rm EA}, \{ \text{\sout{ $\{ v^*, {\rm Root} \}$ }} , \{ \text{\sout{ Root }}, \text{\sout{  IA }} \} \} \} $$ 
because both are among the more costly SM movements (involving non-atomic structures) and
moreover the movement crosses the edge of the phase, violating PIC. 

\smallskip

One possibility is to reinterpret this chain of operations as operations that
happen after structure formation by Merge (in Externalization), as  we will 
discuss further. We just point out here that operations in Externalization 
bypass the edge-of-the-phase problem, since after deletion 
of copies and pruning of branches, phases disappear in Externalization 
and so clitic placement is local. It is
less clear that this would solve the cost problem, as operations come
with some natural cost functions whether they occur in the core structure
building process of Merge of in Externalization, and the two operations that
give rise to more costly SM's, as depicted above, would also give rise to 
more costly operations when relegated to Externalization. 

\smallskip

There is another possibility, which suggests that this case of clitics may be
handled by the same mechanism that we already have in place for
handling the multiple wh-movement via SM. In fact, it shows that from
the point of view of the colored operad generators these have essentially
the same structure.

\smallskip

To see this, we can make two preliminary observations: one that clitics
form clusters (like the multiple wh-movement), and the other is the
existence of subject clitics.

Clusters of clitics are typical of some Romance languages like Italian, where examples of clusters
of two, three, and even four clitics can be constructed. Pairs of clitics
are very abundant, ``{\em glielo dico}" (I tell him), ``{\em te lo mando}" (I send it to you), ``{\em ci si sente}" 
(we'll hear each other, for ``talk to you later"), etc. Triples
also occur easily in phrases like ``{\em ce la si vede brutta}" (we will see it ugly: idiom for ``we get into trouble"). 
Clusters of four clitics in Italian are very rare and somewhat controversial, like 
``{\em vi ci se ne pu\`o trovare}" (there where one can find of those), indicating increasing 
costs (hence unlikeliness) of clustering. This last example sounds awkward 
compared to a common form like  ``{\em ce ne si pu\`o trovare}" with only three clitics. 

Subject clitics do not occur in standard Italian but are typical in Northern Italian,
``{\em Mario el magna}" (Mario eats). They occur frequently paired to the actual
subject, as in this example, which however can be dropped by Null Subject, leaving 
only the subject clitic, as in  ``{\em el magna}".

The case of the subject clitics like ``{\em Mario el magna}" suggests that the subject  
position is split, in a way that is reminiscent of the colored operad generator \eqref{gensstar3b}
that we discussed for the proposed SM explanation of multiple wh-movement. The difference
here is that the EA with its assigned $\theta_E^\downarrow$ theta-coloring
is also present here and co-occurring with the subject clitic. This suggests that instead of a structure
$$ \{ {\rm EA}, \{ {\rm INFL}, \{ \text{\sout{ EA }}, \{ v^*, {\rm Root} \} \} \} $$
with the EA movement by IM, we see an SM extracting a clitic and combining
it with the Subject that is then EM-ed into the Spec position
$$ \{  \{ {\rm Mario}, {\rm el} \} , \{ {\rm INFL}, \{ \text{\sout{ el }}, \{ v^*, {\rm magna} \} \} \} \, . $$

In terms of coloring generators, this would require one modification of the generator \eqref{gensstar3b},
which would distinguish between the case of (clusters of) clitics and the case of (cluster of)
multiple wh-movements, namely a generator of the form
\begin{equation}\label{gensstar4b}
\Tree[ .$\fs^\downarrow_{\omega'}$ [ {$\fs^\downarrow_{\omega'}$} {$\hat\fs^\downarrow_{\omega'}$} ] ] 
\end{equation}
that matches a theta-coloring generator
$$ \Tree[ .$\theta_E^\downarrow$ [ $\theta_E^\downarrow$ $\theta_0$ ] ] $$
unlike \eqref{gensstar3b} that corresponds to only non-theta positions. 

Given this, we would then also be able to accommodate 
additional clitics and their clustering as happening
again at the edge of phase where the EA is usually placed (rather than as movement to
the INFL head in a different phase). 

Indeed SM would be extracting a $\fc_\omega$ colored position (like a IA)  
and moving it via a $\fM_{T_w,T_v}\circ \fM_{T_v,1}$ to
a combination of generators
$$   \Tree[ .$\hat\fs^\downarrow_{\omega}$ [ $\hat\fs^\downarrow_{\omega}$ 
[ .$\hat\fs^\downarrow_{\omega}$ [ {$\fc_\omega$} {$(1,\fm)$} ] ]  ] ]  $$
that can thus insert a combination of clitics in the $\hat\fs^\downarrow_{\omega}$
position of \eqref{gensstar4b}, including clusters of multiple clitics.
Since each additional clitic in the cluster requires one more SM (even
with minimal violations) the costs increase making larger clusters
increasingly costy. 

 \medskip
 \subsection{Possessor agreement constructions}
 
 We mention here another phenomenon that appears to be suitable
 for a formulation in terms of the same type of coloring generators
 that we discussed in the multiple wh case and in the clusters of clitics case,
 namely the possessor agreement constructions in Korean, \cite{Cho}. 
 
 The Korean equivalent of English transitives like {\em Mary kicked John’s leg} have an expected form like (1a) below, where the object has a genitive possessor, just like in English. But they also have a form like (1b), where {\em John} and {\em tali} (‘leg’) are both marked with the accusative:
  \begin{center}
  \includegraphics[scale=0.65]{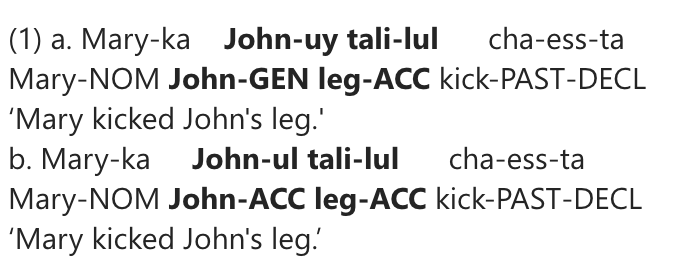} 
 \end{center} 

The latter are sometimes described as showing “possessor agreement”. The term is meant here
as {\em Case agreement} rather than {\em phi-feature agreement} (the regular case as, e.g., in Turkish). 

The difference in case-marking appears to reflect a difference in constituency. Whereas the 
genitive construction (GC) cannot be interrupted by a VP adverb ({\em seykey} ‘hard’), the 
possessor agreement construction (PAC) can, see (2a-b) below:  
 \begin{center}
  \includegraphics[scale=0.65]{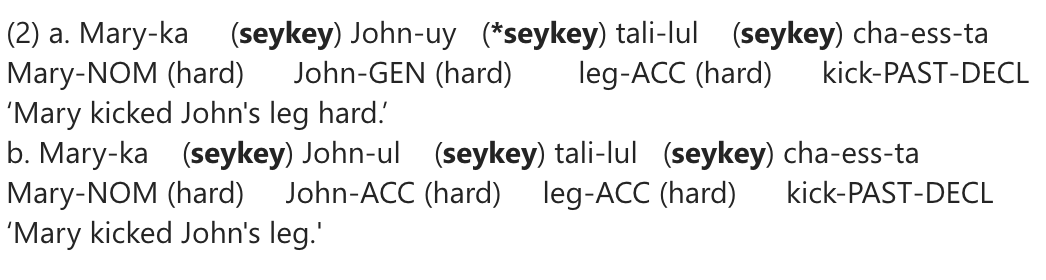} 
 \end{center} 

This suggests the GC constitutes phrase (as in English), but the PAC involves two independent NPs, see (3a-b) below:
 \begin{center}
  \includegraphics[scale=0.65]{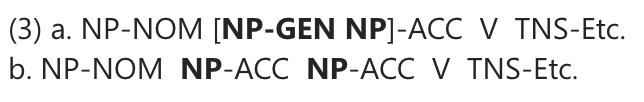} 
 \end{center} 
 
When one passivizes genitive construction examples like (1a) above, one gets the expected result 
where the whole accusative phrase becomes nominative as in (4) below:
 \begin{center}
  \includegraphics[scale=0.65]{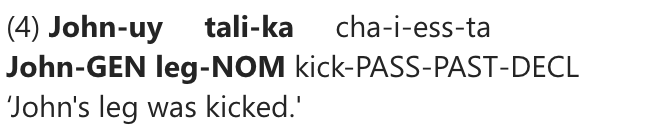} 
 \end{center} 
 
 On the other hand, when passivizing a case like (1b) above, at least two possibilities become available. 
 One is passivizing only the possessor nominal ({\em John}), so that it alone shows up as the nominative 
 subject, as in (5a) below. The second possibility is where (apparently) both accusatives are simultaneously 
 made subjects, despite their non-constituency, as in (5b) below.
  \begin{center}
  \includegraphics[scale=0.65]{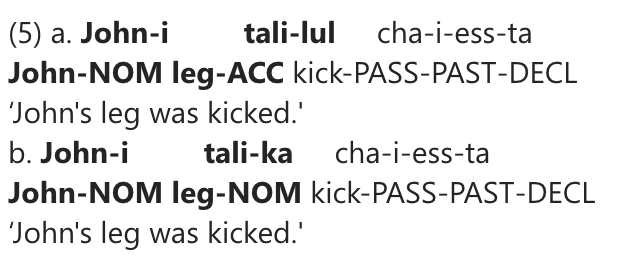} 
 \end{center} 
 
 This is the case that appears related to the other phenomena we have been discussing in this section,
 since the two accusatives (which do not form a constituency) being simultaneously made subjects
 appears to require again a splitting of the Spec position, in the coloring by phases. To provide a
 possible model, in the light of our previous discussion of multiple wh-fronting and clusters of clitics,
 we can analyze the distinction between the passivization of (4) and that of (5b) in terms of colored
 operad generators (for phases and for theta roles). The usual passivization as in (4) can be analyzed
 as in \S 8.1 of \cite{MHL}. We do not reproduce here all the generators involved, as one can directly
 adapt the case discussed in \cite{MHL}, but we note the two key features here: the generator of the form
 $$   \Tree[.{($\fz^\downarrow$,$\theta_I^\downarrow$)}  {(John's, $\theta'_0$, $\fm$)} {(leg, $\theta_I^\downarrow$, $\fz^\downarrow$)} ] $$
 in the structure for ``{\em kick John's leg}" is transformed via movement into a structure (composition of
 two generators) of the form
 $$  \Tree[.{$(\theta_0, \fs^\downarrow)$} [ [ {(John's, $\theta'_0$, $\fm$)} {(leg, $\theta_I^\downarrow$, $\fz^\downarrow$)} ] 
 {$( 1, \theta_0, \fm)$} ] ] $$
 where we combine the coloring by theta roles and phases, as in \cite{MHL}. We also note that, with our formulation
 of theta-coloring as in \cite{MarLar} and \cite{MHL}, in the coloring structure for ``{\em John's leg was kicked}", the
 external theta role $\theta_E^\downarrow$ is discharged at the ``{\em was}" position (see the discussion of
 this point in \S 8.1 of \cite{MHL}). On the other hand, in the PAC form of passivization as in (5b) we find a 
 different form of coloring. We start, in the structure for ``{\em John-ACC leg-ACC kick-PAST-DECL}", with a
 colored generator of the form
 $$  \Tree[.{($\fz^\downarrow$,$\theta_I^\downarrow$)}  {(John-ACC, $\theta'_0$, $\fm$)} {(leg-ACC, $\theta_I^\downarrow$, $\fz^\downarrow$)} ] $$
 which now, however, admits two possible different coloring choices: the one above (as in the previous case) but also
  $$  \Tree[.{($\fz^\downarrow$,$\theta_I^\downarrow$)}  {(John-ACC, $\theta_I^\downarrow$, $\fz^\downarrow$)} {(leg-ACC, $\theta'_0$, $\fm$)} ] $$
where in this second one ``John" received the theta role and ``leg" plays a $\theta_0'$ modifier role (akin to a
case like ``kicked John in the leg"). We return to the interpretation of this ambiguity of coloring below.
 Now, in the case of (5b), instead of extracting (with a single admissible cut) the accessible term
 $$ \Tree[ John's leg ] $$
 and use it with Internal Merge action, one extracts the two terms at the leaves of this generator (with two admissible 
 cuts under the same vertex), 
  $$ (\text{John-ACC}, \theta'_0, \fm) \sqcup (\text{leg-ACC},  \theta_I^\downarrow, \fz^\downarrow) $$
 or in the second coloring
 $$ (\text{John-ACC}, \theta_I^\downarrow, \fz^\downarrow) \sqcup (\text{leg-ACC}, \theta'_0, \fm) \, , $$
 and use them to form generators (via two applications of $\fM_{S,1}$)
 $$  \Tree[.{($\theta_0$, $\hat\fs^\downarrow$)}  {(John-ACC, $\theta'_0$, $\fm$)} {$(1,\theta_0,\fm)$} ] \sqcup
 \Tree[.{($\theta_0$, $\hat\fs^\downarrow$)}  {(leg-ACC, $\theta_I^\downarrow$, $\fz^\downarrow$)} {$(1,\theta_0,\fm)$} ] $$
or the analogous structures with the second choice of coloring. We can now use the same type of generator
used in the case of multiple wh movement and of clusters of clitics,
$$ \Tree[ .{$(\theta_0,\fs^\downarrow)$} {$(\theta_0,\hat\fs^\downarrow)$} {$(\theta_0,\hat\fs^\downarrow)$} ] $$
to form by SIdeward Merge the structure
$$ \Tree[ .{$(\theta_0,\fs^\downarrow)$} [ .{$(\theta_0,\hat\fs^\downarrow)$} {(John-ACC, $\theta'_0$, $\fm$)} {$(1,\theta_0,\fm)$}  ] [ .{$(\theta_0,\hat\fs^\downarrow)$}   {(leg-ACC, $\theta_I^\downarrow$, $\fz^\downarrow$)} {$(1,\theta_0,\fm)$}  ] ] $$

Note that this example is an empirical case in favor of the inclusion of the possibility of
 cutting two edges below the same vertex in the SM cases. We will return to discuss this point 
 further in \S \ref{HopfMArkovSec} (and more in detail in \cite{MarSkig}).
 
Regarding the presence of the two possible coloring choices in the case of  ``{\em John-ACC leg-ACC}"
we can argue that this appears to be the condition that allows for possessor agreement constructions (PACs).
Indeed,  while PACs in Korean are available with the equivalent of {\em Mary hit John’s arm}, they are not 
 available with the equivalent of {\em Mary hit John’s car}.  That is, one cannot get two accusative 
 marked nominals in the latter case. One can argue that this may have to do with ``alienable" vs. ``inalienable" 
 possession--i.e., PACs are only available with the latter. However, that is not quite the correct reason. 
 Indeed, one gets both (6a-b) below, but not both (7a-b) below, despite both pairs involving inalienable 
 possession:
 \begin{center}
  \includegraphics[scale=0.65]{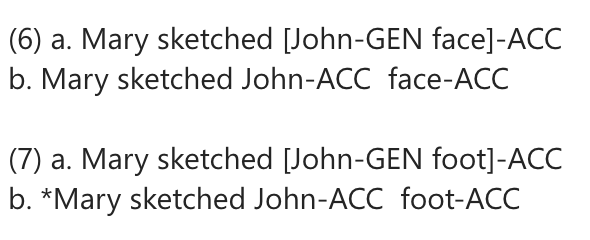} 
 \end{center} 
 
 It turns out that the possessor agreement construction is possible according to the condition:
 $$ 
 \text{ NP1-ACC NP2-ACC V possible } \Longleftrightarrow ( \text{[NP1-GEN NP2]-ACC V} 
 \Longrightarrow \text{ NP1  V})\, . 
 $$
In other words, {\em Mary kicked John’s leg} is possible as a PAC since kicking John’s leg entails 
kicking John. However, hitting John’s car does not entail hitting John, hence PAC is not possible
for this case.  

The comparison of (6a-b) and (7a-b) above is particularly interesting.
If John tells Mary “sketch me”, it is perfectly okay for Mary either to sketch 
the whole of John, or just John's face, but it is not okay for Mary to sketch John's foot. 
In other words, sketching John’s face entails/counts as sketching John, but sketching 
John’s foot does not.

This suggests that in the coloring of Korean PACs, the two ACC objects must be able to
receive the same color --which is exactly the different possible coloring that we have
seen in the example above, so that case agreement reflects some kind of theta-agreement. 
The fact that we can move them by passive as a ``cluster" then reflects the fact that,
under this symmetry, each of the ACC terms can play the role of the $\theta_I^\downarrow$
that is moved to the non-theta position $\fs^\downarrow$ by a generator
$$ \Tree[ .{$(\theta_0,\fs^\downarrow)$} {$(\theta_I^\downarrow,\fz^\downarrow)$} {$(1,\theta_0,\fm)$} ] $$
thus causing the split of the $\fs^\downarrow$ position via the generator
$$ \Tree[ .{$(\theta_0,\fs^\downarrow)$} {$(\theta_0,\hat\fs^\downarrow)$} {$(\theta_0,\hat\fs^\downarrow)$} ] $$
for compatibility with this symmetry. 


\medskip
\section{Markovian property and optimality violations} \label{HopfMArkovSec}

We have argued in the previous section that certain proposed explanation of
phenomena apparently violating EC condition can be described in terms of
SM transformations that do not violate the hard EC constraint and that are only 
minimal violations of the soft Resource Restrictions constraints. 

\smallskip

We point out here that the possible existence of such minimal
violations of RR constraints is predicted by the theoretical
algebraic model, through the Hopf algebra Markov chain
property of Merge, discussed in \S 1.9 of \cite{MCB}. 

\smallskip

The action of Merge on workspaces, written in the form
$$ \fM_{S,S'}=\sqcup \circ (\cB \otimes {\rm id} )\circ \delta_{S,S'} \circ \Delta $$
for a selected pair of syntactic objects $S,S'\in \fT_{\cS\cO_0}$ or as
$$ \cK = \sum_{S,S'} \fM_{S,S'}=\sqcup \circ (\cB \otimes {\rm id} )\circ \Pi_{(2)} \circ \Delta\, , $$
when one considers all possible Merge transformations applicable to a given workspace, 
has the typical form of a class of linear operators that are known as
Hopf algebra Markov chains, introduced and studied in \cite{DiPaRa},
\cite{Pang}, and \cite{Pang2}. They generally consist of operators of the
form $\sqcup \circ \cQ \circ \Delta$, where $\sqcup$ and $\Delta$ are the
product and coproduct of a combinatorial Hopf algebra (in the sense of 
\cite{LodayRoncoComb}), and $\cQ$ is a linear operator. In most cases
previously considered, motivated by probabilistic questions about shuffling 
phrased in terms of an associated Hopf algebra, the operator $\cQ$ 
would be a projector. Here the operator is of the form 
$$  \cQ=(\cB\otimes {\rm id})\circ \Pi_{(2)} =  \sqcup \circ (\cB_{(2)}\otimes {\rm id}) \circ \Delta \, ,   $$
with the notation as we discussed in \S \ref{ECdualHsec}. 

\smallskip

\begin{defn}\label{HopfMarkov}
A Hopf algebra Markov chain is a linear map of the form $\cK=\sqcup \circ \cQ \circ \Delta$ as above,
on a combinatorial Hopf algebra $\cH=(\cV(X),\sqcup,\Delta)$ with linear basis of combinatorial objects
$x\in X$, such that the matrix representation $K_X$ of $\cK$ in the basis $X$ satisfies 
\begin{enumerate}
\item $K_X(x,y)\geq 0$ and, for all $x\in X$ there is at least one $y\in X$ such that $K_X(x,y)>0$;
\item $K_X$ has a Perron--Frobenius eigenfunction $\eta=\sum_x \eta(x)\, x$, with eigenvalue $\lambda>0$, 
$$ \sum_y K_X(x,y) \eta(y)=\lambda \eta(x) \text{ with } \eta(x)> 0, \,\, \forall x\in X\, . $$
\end{enumerate}
Then after a rescaling of the basis by $x\mapsto x/\eta(x)$, the matrix representation of $\cK$ in
the rescaled basis is given by
$$ \hat K_X(x,y) = \frac{1}{\lambda} \frac{\eta(y)}{\eta(x)} K_X(x,y) $$
and is a stochastic matrix, namely
$$ \sum_y \hat K_X(x,y) = \frac{1}{\lambda} \frac{1}{\eta(x)} \sum_y  K_X(x,y) \eta(y)= 1 \, . $$
\end{defn}

\smallskip

Thus, to show that a given map of this form is indeed a Hopf algebra Markov chain, one
needs to show that the two properties listed above hold. In the case of the Merge action,
for $\cK=\sqcup \circ \circ (\cB_{(2)}\otimes {\rm id}) \circ \Delta$ as above, the first property
is clearly verified. The general strategy to prove that the second property is satisfied
is to ensure that the Perron--Frobenius  theorem holds. For a non-negative matrix $K_X$
(where here the combinatorial basis is $X=\fF_{\cS\cO_0}$), the applicability of the
Perron--Frobenius  theorem depends on the {\em strong connectedness} of the graph
with vertex set $X$ and a directed edge from $x$ to $y$ iff $K_X(x,y)>0$. Strong connectedness
means that any two vertices in this graph are connected by a path of directed edges. 

\smallskip

In \S 1.9 of \cite{MCB} strong connectedness is proved for the graph with vertices
given by all the forests $F\in \fF_{\cS\cO_0}$ with nonempty set of edges
and with an assigned set $L=L(F)$ of leaves, and with directed edges 
whenever $K_X(F,F')>0$. The proof relies on the fact that the map $\cK$
includes also the Sideward Merge operations, which play a crucial role in
obtaining the strong connectedness property.

\smallskip

We give here a slightly more detailed account of this result of \S 1.9 of \cite{MCB},
where we refine the argument by showing that, in fact, cases of Sideward Merge
with small violations of the Resource Restriction constraints suffice to ensure
strong connectedness (hence Perron--Frobenius  theory), although they need
to involve violations of size up to $2$ in both Resource Restriction and 
Complexity Loss.

\smallskip

For the purpose of this paper we will focus on illustrating one simple example
(the case with $3$ leaves, where a complete computation can be shown easily)
and we prove a general property of the Hopf algebra Markov chain (the
strong connectedness), which shows the role fo the SM operations. For a more
general and detailed analysis of Merge as a Hopf algebra Markov chain and its
properties as a dynamical system we refer the readers to the forthcoming
paper \cite{MarSkig}. 

\smallskip

One should keep in mind that here all operations EM, IM, and SM
are considered as applying freely and completely unconstrained: this means that 
we do not apply the I-language filters that check head and complement structure, 
labeling, and theta-roles. 

\smallskip

Thus, in particular, the occurrence of these optimality violating SM operations at this
stage does not reflect their occurrence at the interfaces, after this filtering. Indeed,
the previous analysis of \S \ref{AltSMsec} and \ref{ExtSec} shows that phenomena
that appear to require SM operations may have alternative explanation, reducing
SM to an optimality violating operation that only has the internal structural function
described in this section. (We comment more at the end of this section on why
the Hopf algebra Markov chain property is important.) 

\smallskip

We analyze first the simplest example, with a set of $3$ labelled leaves.
We then discuss, inductively, the general case.

\subsection{A simple example}\label{ExGsec}

\begin{figure}
    \begin{center}
    \includegraphics[scale=0.35]{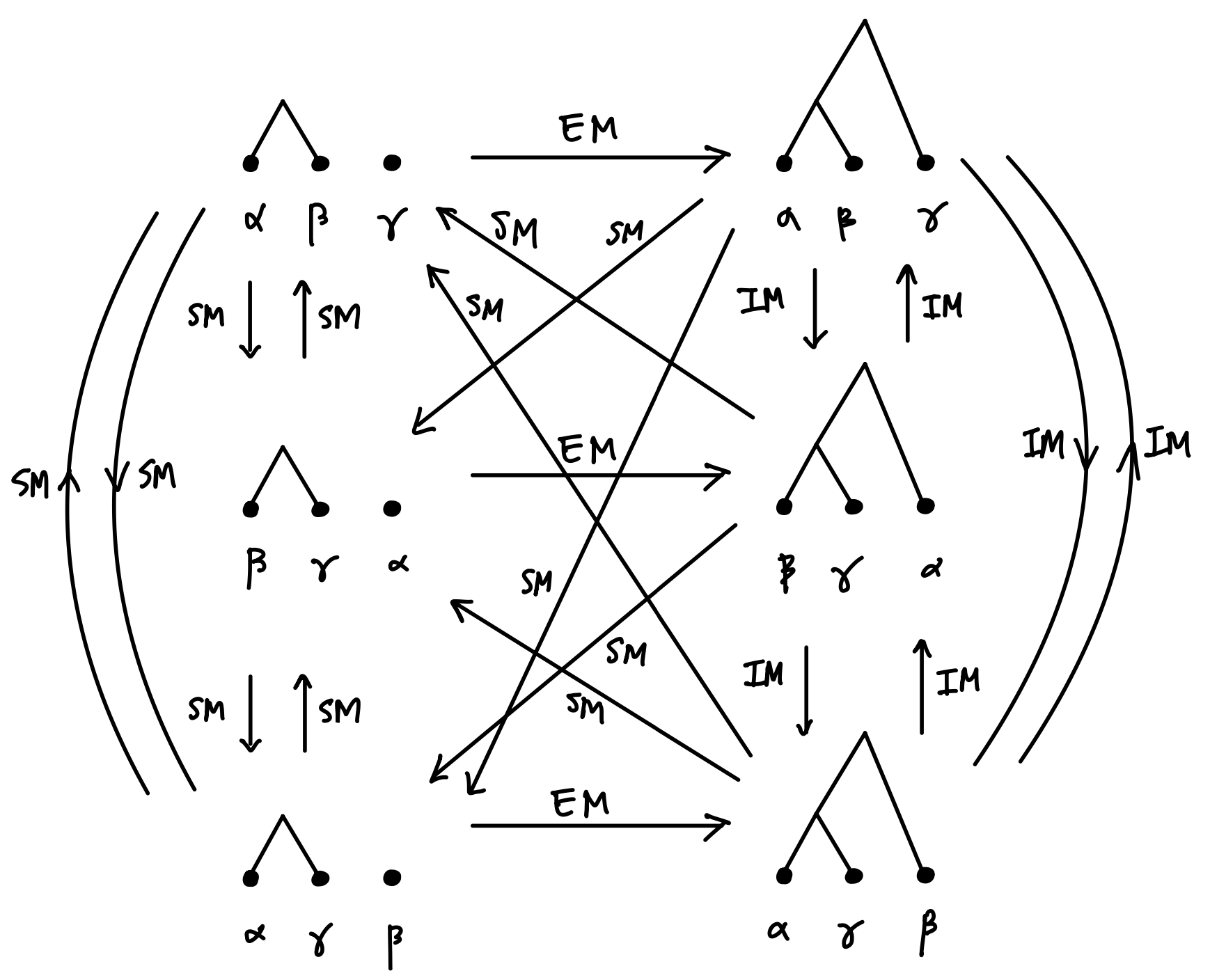} 
       \caption{Strong Connectedness for the Hopf algebra Markov chain illustrated in the case of $3$ leaves. \label{StrongConnectFig}}
    \end{center}
\end{figure}

We consider a given set $L=\{ \alpha, \beta, \gamma \}\subset \cS\cO_0$. Note that,
in principle, repetitions are allowed (``Buffalo buffalo buffalo", for example) but, for
simplicity, we can assume this is just a set rather than a multiset, as the argument
does not change. We consider all the possible forests $F\in \fF_{\cS\cO_0}$ with
set of leaves $L(F)=L$, and with nonempty set of edges, $E(F)\neq \emptyset$.
There are six such forests, shown in Figure~\ref{StrongConnectFig}. We put
these six forests at the vertices of a graph with directed edges given by the
possible Merge operations connecting them. This results in the graph of 
Figure~\ref{StrongConnectFig}, where the edges are marked according to
whether they are EM, IM, or SM. We use everywhere here the coproduct of
the form $\Delta^d$, as we discuss more in \S \ref{GenGsec}. 

\smallskip

Observe that, in this specific case, the SM arrows are always of two types: 
\begin{itemize}
\item {\bf SM(3)} with $\fM(T_v,T_w) \sqcup T/^d (T_v \sqcup T_w)$ where $T_v=\alpha$ and $T_w=\beta$ are atomic (single leaves):
$$ \Tree[[ $\alpha$ $\beta$ ] $\gamma$ ] \ \stackrel{\Delta^d}{\mapsto} \ (\beta \sqcup \gamma) \otimes \alpha \ \stackrel{\cB_{(2)}\otimes {\rm id}}{\mapsto} \  \Tree[ $\beta$ $\gamma$ ] \otimes \alpha \ \stackrel{\sqcup}{\mapsto} \ \Tree[ $\beta$ $\gamma$ ] \sqcup \alpha \, , $$
and similar cases with permuted leaves;
\item {\bf SM(1)} with $\fM(T_v, T') \sqcup T/^d T_v$ where $T_v=\alpha$ is atomic (single leaf), and in this specific case with only $3$ leaves also $T'=\beta$ is atomic:
$$ \Tree[ $\alpha$ $\beta$ ] \sqcup \gamma \ \stackrel{\Delta^d}{\mapsto} \  (\beta \sqcup \gamma)\otimes \alpha \ \stackrel{\cB_{(2)}\otimes {\rm id}}{\mapsto} \  \Tree[ $\beta$ $\gamma$ ] \otimes \alpha \ \stackrel{\sqcup}{\mapsto} \  \Tree[ $\beta$ $\gamma$ ] \sqcup \alpha \, ,
  $$
  and similar cases with permuted leaves. 
\end{itemize}
We will discuss in \S \ref{GenGsec} the case of structures with
arbitrarily large number of leaves, where the second case will include more general forms of SM,  
$\fM(\alpha, T') \sqcup T/\alpha$, where $T'$ need not be atomic. Note that all these cases create
{\em violations of size at most $2$} in both the Resource Restriction and the degree (No Complexity Loss) constraints.

\smallskip

We also make two simplifying choices here. Since we are using the coproduct $\Delta^d$, the IM operations
applied to an accessible term $T_v$, for $v$ one of the two vertices immediately below the root of $T$, are just
the identity operation so we will not include these. Moreover, in the SM transformations we will not include the
case where the admissible cut $C$ cuts the two edges below the same vertex, as in, for example
$$ \Tree[ [ $\alpha$ $\beta$ ] $\gamma$ ] \mapsto \alpha \sqcup \beta \otimes \gamma \mapsto \Tree[ $\alpha$ $\beta$ ] \sqcup \gamma \, . $$
The main reason for not including these SM transformations is that the coproduct $\Delta^d$, has better
algebraic properties when these admissible cuts are excluded. On the other hand, the general
analysis in \cite{MarSkig} of the resulting Hopf algebra Markov chain shows that the dynamical system
has better properties when these admissible cuts are {\em not} excluded. For the purpose of the example
we discuss here the presence or absence of the three arrows that would correspond to these SM transformation
do not alter significantly the result, so we just choose to illustrate the first case, with this type of admissible cuts 
not included (as one can see in Figure~\ref{StrongConnectFig} there is no SM arrow from a tree
in the right column to the forest immediatley adjacent to it in the left column). 

\smallskip

\begin{figure}
    \begin{center}
    \includegraphics[scale=0.3]{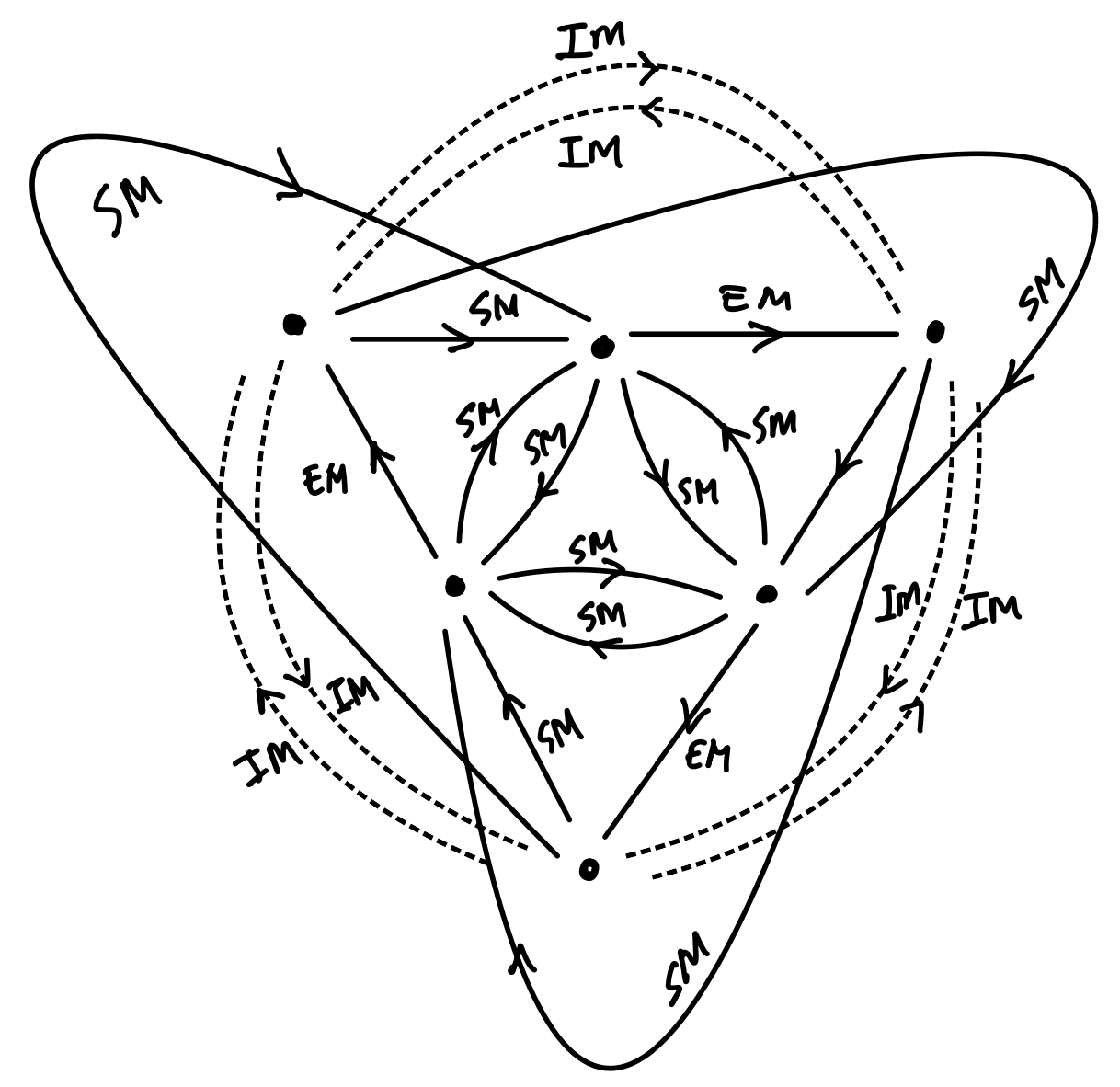}  \ \ \ \ 
    \includegraphics[scale=0.3]{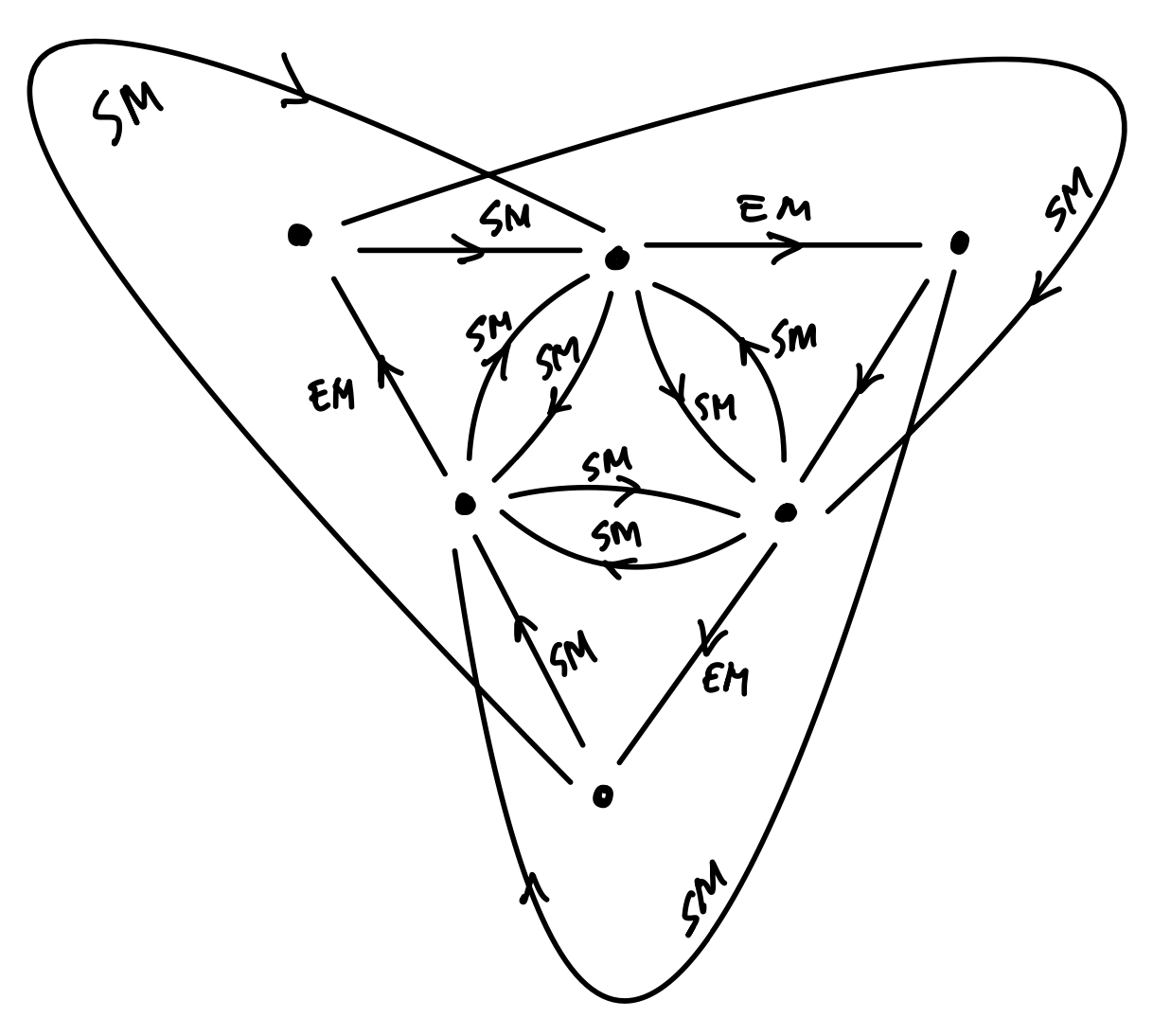} 
       \caption{Strongly connected graph for structures with $3$ leaves, with and without Internal Merge arrows. \label{StrongConnectFig2}}
    \end{center}
\end{figure}

As we will discuss in \S \ref{GenGsec}, it suffices to show strong connectedness for the
graph where we keep only the EM and SM arrows, which slightly simplifies the argument.
In this example, the comparison between the graph of Figure~\ref{StrongConnectFig} that
includes IM and the graph without the IM arrows is shown in Figure~\ref{StrongConnectFig2}.
Direct inspection of the two graphs in Figure~\ref{StrongConnectFig2} shows that both are
indeed strongly connected, as one can find a directed path between any two vertices. 

\smallskip

The matrix $K_X$, in the basis $X=\fF_{\cS\cO_0}$ (restricted to the case of $3$ leaves and a 
nonempty set of edges) is given in the first case (including IM) by
\begin{equation}\label{KXwIM}
K_X = \begin{pmatrix} 0 & 1 & 1 & 0 & 1 & 1 \\
1 & 0 & 1 & 1 & 0 & 1 \\
1 & 1 & 0 & 1 & 1 & 0 \\
1 & 0 & 0 & 0 & 1 & 1 \\
0 & 1 & 0 & 1 & 0 & 1 \\
0 & 0 & 1 & 1 & 1 & 0 
\end{pmatrix}  \ \ \ \  \text{including IM,}
\end{equation}
and in the second case (not including IM)
\begin{equation}\label{KXwoIM}
K'_X = \begin{pmatrix} 0 & 0 & 0 & 0 & 1 & 1 \\
0 & 0 & 0 & 1 & 0 & 1 \\
0 & 0 & 0 & 1 & 1 & 0 \\
1 & 0 & 0 & 0 & 1 & 1 \\
0 & 1 & 0 & 1 & 0 & 1 \\
0 & 0 & 1 & 1 & 1 & 0 
\end{pmatrix}  \ \ \ \  \text{only EM and SM.}
\end{equation}

The matrix $K_X$ has 
Perron-Frobenius eigenvalue $\lambda=2+\sqrt{2}$ and 
Perron-Frobenius eigenvector $\eta=(\sqrt{2}, \sqrt{2},\sqrt{2}, 1,1,1)$ 
(written as a column vector).
The matrix $K'_X$ has 
Perron-Frobenius eigenvalue $\lambda'=1+\sqrt{3}$ and
Perron-Frobenius eigenvector 
$\eta'=(2/\lambda',2/\lambda',2/\lambda', 1,1,1)$ 
(again written as a column vector).
While for showing strong connectedness one can replace $K_X$ with the simpler $K'_X$, only the case of
$K_X$ has linguistic relevance, when it comes to obtaining information about the behavior of the
Markov chain using Perron--Frobenius theory. 

\smallskip

When computing the stochastic matrix $\hat K_X$ the
scaling factors in the four sectors of the matrix $K_X$ are given by
$$   \begin{array}{c | c}  \begin{array}{l}  \eta(x)=\sqrt{2} \\ \eta(y)=\sqrt{2} \end{array}
 &   \begin{array}{l}  \eta(x)=\sqrt{2} \\ \eta(y)=1 \end{array} \\ \hline  \begin{array}{l}  \eta(x)=1 \\ \eta(y)=\sqrt{2} \end{array} &   \begin{array}{l} \eta(x)=1 \\ \eta(y)=1 \end{array}
\end{array}  $$
so that we obtain
$$ \hat K_X =   \frac{1}{\lambda}   \frac{\eta(y)}{\eta(x)} K_X(x,y)= \frac{1}{2+\sqrt{2}} \begin{pmatrix}
0 & 1 & 1 & 0 & 1/\sqrt{2}   & 1/\sqrt{2}    \\
1 & 0 & 1 & 1/\sqrt{2}   & 0 & 1/\sqrt{2}    \\
1 & 1 & 0 & 1/\sqrt{2}   & 1/\sqrt{2}   & 0 \\
\sqrt{2}  & 0 & 0 & 0 & 1 & 1 \\
0 & \sqrt{2}  & 0 & 1 & 0 & 1 \\
0 & 0 & \sqrt{2}  & 1 & 1 & 0 
\end{pmatrix} \, . $$
This results in the bistochastic matrix (each row and each column adds up to $1$),
\begin{equation}\label{bistohatKX} \hat K_X =  \begin{pmatrix}
0 & \frac{1}{2+\sqrt{2}} & \frac{1}{2+\sqrt{2}} & 0 & \frac{1}{2+2\sqrt{2}} & \frac{1}{2+2\sqrt{2}} \\
\frac{1}{2+\sqrt{2}} & 0 & \frac{1}{2+\sqrt{2}} & \frac{1}{2+2\sqrt{2}} & 0 & \frac{1}{2+2\sqrt{2}} \\
\frac{1}{2+\sqrt{2}} & \frac{1}{2+\sqrt{2}} & 0 & \frac{1}{2+2\sqrt{2}} & \frac{1}{2+2\sqrt{2}} & 0 \\
\frac{\sqrt{2}}{2+\sqrt{2}} & 0 & 0 & 0 & \frac{1}{2+\sqrt{2}}  & \frac{1}{2+\sqrt{2}}  \\
0 & \frac{\sqrt{2}}{2+\sqrt{2}} & 0 & \frac{1}{2+\sqrt{2}}  & 0 & \frac{1}{2+\sqrt{2}}  \\
0 & 0 & \frac{\sqrt{2}}{2+\sqrt{2}} & \frac{1}{2+\sqrt{2}}  & \frac{1}{2+\sqrt{2}}  & 0 
\end{pmatrix} \, . 
\end{equation}
Since $\hat K_X$ is bistochastic, the left Perron-Frobenius eigenvector for $\hat K_X$ is then the uniform distribution $\xi(x)=1$, 
that satisfies $\xi \hat K_X =\xi$. So we see that the stationary distribution of the Markov chain is 
the uniform probability distribution over all the structures with assigned leaf set $L$ and nontrivial
set of edges. 

This can be compared with the case of $K_X'$ where we only consider EM and SM, but without IM, which
gives a stochastic (but not bistochastic) matrix
$$ \hat K'_X =  \begin{pmatrix} 
0 & 0 & 0 & 0 & \frac{1}{2} & \frac{1}{2}  \\
0 & 0 & 0 & \frac{1}{2} & 0 & \frac{1}{2}  \\
0 & 0 & 0 & \frac{1}{2} & \frac{1}{2}  & 0  \\
\frac{2}{(1+\sqrt{3})^2} & 0 & 0 & 0 & \frac{1}{1+\sqrt{3}} &  \frac{1}{1+\sqrt{3}}  \\
0 & \frac{2}{(1+\sqrt{3})^2} & 0 & \frac{1}{1+\sqrt{3}} & 0 &  \frac{1}{1+\sqrt{3}} \\
0 & 0 & \frac{2}{(1+\sqrt{3})^2} & \frac{1}{1+\sqrt{3}} &  \frac{1}{1+\sqrt{3}} & 0 
\end{pmatrix} \, . $$
This has left Perron-Frobenius eigenvector  $\xi'  \hat K'_X=\xi'$
given by the non-uniform distribution 
$\xi'=Z^{-1} (2-\sqrt{3},2-\sqrt{3},2-\sqrt{3},1,1,1)$ with the
normalization factor $Z=3(3-\sqrt{3})$. 
In other words, one sees in this very simple example that, in the absence
of IM, the presence of SM would push the dynamics with higher probability towards disconnected states
(forests that are not trees), while when all possible forms of Merge are included (EM, IM, and SM) the
stationary distribution is uniform over all the workspaces. (This behavior is specific to the case $n=3$,
though it is illustrative of the general effect of the different forms of Merge on the dynamics.)

\smallskip

A more general analysis of the
dynamics of the Hopf algebra Markov chain describing the action of Merge on workspaces 
is discussed in \cite{MarSkig}. We only include here the simplest example to provide some intuition
for the behavior. In \S \ref{GenGsec} we prove a statement about the dynamics in the general
case that refines the result of \S 1.19 of \cite{MCB} and that shows that one always has
strong connectedness for the graph formed by EM and the two forms of SM:
{\bf SM(3)} with $\fM(\alpha,\beta) \sqcup T/^d (\alpha \sqcup \beta)$ and 
{\bf SM(1)} with $\fM(\alpha, T') \sqcup T/^d \alpha$.

\smallskip
\subsubsection{Identity SM and dynamics}

We use this simple example to also outline an aspect of the use of the coproduct $\Delta^d$
that was not discussed in detail in \cite{MCB}. It is known from the discussion in \S 1.2
and 1.3 of \cite{MCB} that the coproduct $\Delta^d$ only satisfies cocommutativity up to
an error term that has to do with mismatched multplicities in the counting of terms. A more
precise discussion of the resulting algebraic properties will be presented in forthcoming
work \cite{MarWal}. We just mention here the fact that in the coproduct $\Delta^d(T)$
one also has a term of the form $T_1 \sqcup T_2 \otimes 1$, which gives rise to
an identity SM operation that simpy reassembles $T$ itself by merging back $T=\fM(T_1,T_2)$.
This identity SM operation, if we include it as part of the Hopf algebra Markov chain, has the
effect of adding a single looping edge attached to each vertex of the graph $G_L$.

\smallskip

In the example we discussed here with $3$ leaves, we have analyzed the dynamics by
including only the nontrivial Merge transformation and not these trivial identity SM loops.
If we include these as well, the graph of Figure~\ref{StrongConnectFig2} is extended
by adding a looping edge to each of the six vertices. This means that the matrix $K_X$
of \eqref{KXwIM} is replaced by the matrix
$$ K_X'' ={\rm Id} + K_X = \begin{pmatrix} 1 & 1 & 1 & 0 & 1 & 1 \\
1 & 1 & 1 & 1 & 0 & 1 \\
1 & 1 & 1 & 1 & 1 & 0 \\
1 & 0 & 0 & 1 & 1 & 1 \\
0 & 1 & 0 & 1 & 1 & 1 \\
0 & 0 & 1 & 1 & 1 & 1 
\end{pmatrix} \,
$$
which has Perron-Frobenius eigenvalue $3+\sqrt{2}$ and corresponding
Perron-Frobenius eigenvector $\eta=(\sqrt{2},\sqrt{2}, \sqrt{2},1,1,1)$ (as
a column vector), resulting in the matrix $\hat K''_X$ of the form
$$ \hat K_X'' =\begin{pmatrix}
\frac{1}{3 + \sqrt{2}} & \frac{1}{3 + \sqrt{2}} & \frac{1}{3 + \sqrt{2}} & 0 & \frac{1}{\sqrt{2}(3 + \sqrt{2})} & \frac{1}{\sqrt{2}(3 + \sqrt{2})} \\
\frac{1}{3 + \sqrt{2}} & \frac{1}{3 + \sqrt{2}} & \frac{1}{3 + \sqrt{2}} & \frac{1}{\sqrt{2}(3 + \sqrt{2})} & 0 & \frac{1}{\sqrt{2}(3 + \sqrt{2})} \\
\frac{1}{3 + \sqrt{2}} & \frac{1}{3 + \sqrt{2}} & \frac{1}{3 + \sqrt{2}} & \frac{1}{\sqrt{2}(3 + \sqrt{2})} & \frac{1}{\sqrt{2}(3 + \sqrt{2})} & 0 \\
\frac{\sqrt{2}}{3 + \sqrt{2}} & 0 & 0 & \frac{1}{3 + \sqrt{2}} & \frac{1}{3 + \sqrt{2}} & \frac{1}{3 + \sqrt{2}} \\
0 & \frac{\sqrt{2}}{3 + \sqrt{2}} & 0 & \frac{1}{3 + \sqrt{2}} & \frac{1}{3 + \sqrt{2}} & \frac{1}{3 + \sqrt{2}} \\
0 & 0 & \frac{\sqrt{2}}{3 + \sqrt{2}} & \frac{1}{3 + \sqrt{2}} & \frac{1}{3 + \sqrt{2}} & \frac{1}{3 + \sqrt{2}} 
\end{pmatrix} $$
which is again a bistochastic matrix, so that one again obtains
the same uniform distribution as in the case of $K_X$. 

\smallskip
\subsubsection{Optimality violations and dynamics} 

In the Hopf algebra Markov chain that we considered here, we have weighted every
operation, EM, IM, and SM, in the same way in forming the matrix $K_X$. On the
other hand, we can also take into consideration the optimality constraints, both of
Minimal Search and of Resource Restriction. Thus, to each arrow in the graph $G_L$
we can assign a weight, which we can write in the form $t^c$ for some
parameter $t >0$ and with $c=c(K_X(x,y))$ the cost associated to $K_X(x,y)$.

\smallskip

We have considered three different types of cost functions, that can be viewed
individually or in combination: Minimal Search, Minimal Yield, No Complexity Loss.
We illustrate here, in this simple example with $3$ leaves, how incorporating
these costs affects the Hopf algebra Markov chain dynamics. 

\smallskip

\begin{prop}\label{KXcosts}
The matrix $K_X$ with entries weighted by the costs, either under
Minimal Search, or Minimal Yield, or Complexity Loss, or under the 
overall combined cost, takes the form:
\begin{equation}\label{KXms}
 K^{c_{MS}}_{X, t}= \begin{pmatrix} 0 & 1 & 1 & 0 & t^{1/3} & t^{1/3} \\
1 & 0 & 1 & t^{1/3} & 0 & t^{1/3} \\
1 & 1 & 0 & t^{1/3} & t^{1/3} & 0 \\
1 & 0 & 0 & 0 & t^{1/2} &  t^{1/2} \\
0 & 1 & 0 &  t^{1/2} & 0 &  t^{1/2} \\
0 & 0 & 1 &  t^{1/2}  &  t^{1/2} & 0 
\end{pmatrix}  \ \ \  \text{with Minimal Search costs,}
\end{equation}
\begin{equation}\label{KXmy}
 K^{c_{MY}}_{X, t}= \begin{pmatrix} 0 & 1 & 1 & 0 & t^{-1} & t^{-1} \\
1 & 0 & 1 & t^{-1} & 0 & t^{-1} \\
1 & 1 & 0 & t^{-1} & t^{-1} & 0 \\
t & 0 & 0 & 0 & 1 & 1 \\
0 & t & 0 & 1 & 0 & 1\\
0 & 0 & t & 1 & 1  & 0 
\end{pmatrix}  \ \ \  \text{with Minimal Yleld costs,}
\end{equation}
\begin{equation}\label{KXcl}
 K^{c_{CL}}_{X, t}= \begin{pmatrix} 0 & 1 & 1 & 0 & t^2 & t^2 \\
1 & 0 & 1 & t^2 & 0 & t^2 \\
1 & 1 & 0 &t^2 & t^2 & 0 \\
1 & 0 & 0 & 0 & t & t \\
0 & 1 & 0 & t & 0 & t \\
0 & 0 & 1 & t  & t & 0 
\end{pmatrix}  \ \ \  \text{with Complexity Loss costs,}
\end{equation}
\begin{equation}\label{KXc}
 K^c_{X, t}= \begin{pmatrix} 0 & 1 & 1 & 0 & t^{4/3} & t^{4/3}  \\
1 & 0 & 1 & t^{4/3}  & 0 & t^{4/3}  \\
1 & 1 & 0 & t^{4/3}  & t^{4/3}  & 0 \\
t & 0 & 0 & 0 & t^{3/2}  & t^{3/2}   \\
0 & t & 0 & t^{3/2}   & 0 & t^{3/2}   \\
0 & 0 & t & t^{3/2}   & t^{3/2}   & 0 
\end{pmatrix}  \ \ \  \text{with all costs combined,}
\end{equation}
\end{prop}

\proof
Under all of these cost functions IM has zero cost. In the case of Minimal Yield we
are using the coproduct $\Delta^d$, so IM has $0$ change in number of
components and number of accessible terms and in their sum, as discussed in \S \ref{MYsec}.
It also incurs no complexity loss, and is cost zero for Minimal Search. Thus, there is
no power of $t$ rescaling the IM sector of the matrix $K_X$ (the top first quadrant). 
EM has zero cost for MS, but achieves a change of $+1$ (Minimal Yield) in the sum
of the changes to number of components and number of accessible terms, with
the first decreasing by one and the latter increasing by two, \S \ref{MYsec}. So the
MY weight for the EM sector is $t$. There is no complexity loss in EM so this is
the overall weight of the EM sector. The two SM sectors in $K_X$ are different
forms of SM. The top right quadrant  (arrows from the single trees to the forests) is an {\bf SM(3)} that produces (with the
coproduct $\Delta^d$) a combined change of $-1$ in number of connected
components (changing by $+1$) and number of accessible terms (changing by $-2$),
see \S \ref{MYsec}. Moreover, the cost $\fc$ of Minimal Search for this 
SM is $\fc_2=1/3$. In this case the SM also incurs in a complexity loss of $+2$ since
extraction of two leaves maps the original root to the remaining structure that has a single leaf. 
In the case of the bottom right quadrant (arrows between the forests),
we have a form of {\bf SM(1)} Sideward Merge, which behaves like IM for Minimal Yield
(so overall change zero), but has cost $\fc_1=1/2$ for Minimal Search and incurs
in a complexity loss of $+1$. 
\endproof

\smallskip

The following is a direct linear algebra computation.

\begin{prop}\label{ulambdalem}
Consider a matrix of the form 
\begin{equation}\label{Kabc}
 K_{a,b,c}(t):= \begin{pmatrix} 0 & 1 & 1 & 0 & t^a & t^a  \\
1 & 0 & 1 & t^a  & 0 & t^a  \\
1 & 1 & 0 & t^a  & t^a  & 0 \\
t^c & 0 & 0 & 0 & t^b  & t^b   \\
0 & t^c & 0 & t^b   & 0 & t^b   \\
0 & 0 & t^c & t^b   & t^b   & 0 
\end{pmatrix}  
\end{equation}
for an assigned triple $(a,b,c)$ of real parameters. Then the Perron-Frobenius eigenvector is of
the form $\eta=(u,u,u,1,1,1)$ (as a 
column vector), with the function $u$ given by
\begin{equation}\label{uabc}
u = t^{-c} (1-t^b + ((1-t^b)^2 + 2 t^a t^b )^{1/2} )
\end{equation}
and with the Perron-Frobenius eigenvalue $\lambda$ given by
 \begin{equation}\label{abclambda}
 \lambda = 1 + t^b + ((1-t^b)^2 + 2 t^a t^b )^{1/2} 
 \end{equation}
The corresponding matrix $\hat K_{abc}$ is given by
\begin{equation}\label{hatKabc}
 \hat K_{abc}(t) = 
\begin{pmatrix} 0 & \lambda^{-1} & \lambda^{-1} & 0 & \lambda^{-1} u^{-1} t^a
& \lambda^{-1} u^{-1} t^a \\
\lambda^{-1} & 0 & \lambda^{-1} & \lambda^{-1} u^{-1} t^a & 
0 & \lambda^{-1} u^{-1} t^a \\ 
\lambda^{-1}  & \lambda^{-1}  & 0 & 
\lambda^{-1} u^{-1} t^a & \lambda^{-1} u^{-1} t^a & 0 \\
\lambda^{-1} u t^c & 0 & 0 & 0 & \lambda^{-1} t^b & \lambda^{-1} t^b \\
0 & \lambda^{-1} u t^c & 0 & \lambda^{-1} t^b & 0 & \lambda^{-1} t^b \\
0 & 0 & \lambda^{-1} u t^c & \lambda^{-1} t^b & \lambda^{-1}t^b & 0 
\end{pmatrix} \, . 
\end{equation} 
In turn, the stochastic matrix $\hat K_{abc}(t)$ obtained in this way has 
Perron-Frobenius eigenvector $\xi \hat K_{abc}=\xi$ which is the non-uniform
probability distribution  $\xi=Z^{-1}(v,v,v,1,1,1)$, with 
$$ v=\frac{u t^c}{\lambda-2} = \frac{1-t^b + ((1-t^b)^2 + 2 t^a t^b )^{1/2}}{-1+t^b  + ((1-t^b)^2 + 2 t^a t^b )^{1/2}} \, , $$
with normalization factor $Z=3v +3$. 
\end{prop}

Note that the asymptotic distribution $v$ is independent of the cost $c$ assigned to the EM sector.
Thus, EM behaves as a zero cost operation, even though from the MS perspective it has a 
nontrivial effect on size counting by altering the number of connected components and of accessible terms. 

\smallskip

In the cases of the cost functions in \eqref{KXms}, \eqref{KXmy}, \eqref{KXcl}, and \eqref{KXc}, we find, the following cases.

When weighting according to Minimal Search, we have 
\begin{equation}\label{vms}
v_{{\rm MS}}= \frac{1-t^{1/2} + ((1-t^{1/2})^2 + 2 t^{5/6} )^{1/2}}{-1+t^{1/2}  + ((1-t^{1/2})^2 + 2 t^{5/6} )^{1/2}} \, ,
\end{equation}
and the asymptotic distribution of the form
$$ \xi_{{\rm MS}}= \left( \frac{v_{{\rm MS}}}{Z_{{\rm MS}}} , \frac{v_{{\rm MS}}}{Z_{{\rm MS}}} , \frac{v_{{\rm MS}}}{Z_{{\rm MS}}} ,
\frac{1}{Z_{{\rm MS}}} , \frac{1}{Z_{{\rm MS}}}, \frac{1}{Z_{{\rm MS}}} \right) \, . $$
This gives the behavior of the asymptotic distribution at the limits $t\to 0$ and $t \to 1$ of the form
\begin{equation}\label{xims}
\begin{array}{lcl}
\displaystyle{
 \frac{ v_{\rm MS} }{ Z_{\rm MS} } 
 } 
 & \sim_{t\to 0} &    \frac{1}{3} - \frac{t^{5/6}}{6} - \frac{t^{4/3}}{3}+ \frac{t^{5/3}}{4} + \cdots    \\[4mm]
\displaystyle{ 
 \frac{1}{ Z_{\rm MS} }  
} 
& \sim_{t\to 0} &       \frac{t^{5/6}}{6} + \frac{t^{4/3}}{3}-  \frac{t^{5/3}}{4} + \cdots                     \\[7mm]
\displaystyle{ 
\frac{ v_{\rm MS} }{ Z_{\rm MS} } 
} 
& \sim_{t\to 1} &   \frac{1}{6} - \frac{\sqrt{2}}{24} (t - 1) + \frac{\sqrt{2}}{36} (t - 1)^2  + \cdots      \\[4mm]
\displaystyle{ 
 \frac{1}{ Z_{\rm MS} } 
 } 
 & \sim_{t\to 1} & \frac{1}{6} + \frac{\sqrt{2}}{24} (t - 1) - \frac{\sqrt{2}}{36} (t - 1)^2  + \cdots 
\end{array}
\end{equation}
This shows that, in the asymptotic limit of $t\to 0$ the limiting behavior of the Hopf algebra Markov chain
settles on the uniform $1/3$ distribution
over the $3$ connected structures, with the disconnected structures not contributing to the asymptotic
behavior. This indicates the situation where all structure formation is performed and IM remains the only
possible operation. SM is automatically excluded in the long term behavior. On the other hand, in the
limit of $t\to 1$ the asymptotic behavior is the uniform distribution over all the workspaces, reproducing in
the limit the behavior of the un-weighted case. 

In the case of the MY constraints, the matrix $\hat K$ remains the same as in \eqref{bistohatKX}, as
the weights $t$ and $t^{-1}$ of EM and SM(3) in \eqref{KXmy} compensate exactly when
computing $\hat K$ 
\begin{equation}\label{vmy}
v_{{\rm MY}}= 1 \, ,
\end{equation}
since in this case the matrix $\hat K$ remains bistochastic for all $t>0$ with the uniform distribution as
asymptotics. Thus, weighting according to MY does not affect the dynamics at all with respect to the
un-weighted case with $t=1$. 

In the case of weighting with respect to complexity loss, we obtain 
\begin{equation}\label{vcl}
v_{{\rm CL}}=  \frac{1-t + ((1-t)^2 + 2 t^3)^{1/2}}{-1+t  + ((1-t)^2 + 2 t^3)^{1/2}}\, ,
\end{equation}
the behavior of the asymptotic distribution $\xi$ satisfies
\begin{equation}\label{xicl}
\begin{array}{lcl}
\displaystyle{
 \frac{ v_{\rm CL} }{ Z_{\rm CL} } 
 } 
 & \sim_{t\to 0} &    \frac{1}{3} - \frac{1}{6} t^3 - \frac{1}{3} t^4  + \cdots    \\[4mm]
\displaystyle{ 
 \frac{1}{ Z_{\rm CL} }  
} 
& \sim_{t\to 0} &      \frac{1}{6} t^3 + \frac{1}{3} t^4  + \cdots                        \\[7mm]
\displaystyle{ 
\frac{ v_{\rm CL} }{ Z_{\rm CL} } 
} 
& \sim_{t\to 1} &      \frac{1}{6} - \frac{\sqrt{2}}{12} (t - 1) + \frac{\sqrt{2}}{8} (t - 1)^2  + \cdots    \\[4mm]
\displaystyle{ 
 \frac{1}{ Z_{\rm CL} } 
 } 
 & \sim_{t\to 1} & \frac{1}{6} + \frac{\sqrt{2}}{12} (t - 1) - \frac{\sqrt{2}}{8} (t - 1)^2  + \cdots  
\end{array}
\end{equation}
Again we see that, while the effect on the asymptotic distribution for a fixed value of
the parameter $t>0$ is different in the MS and the CL cases, both recover, in the limit $t\to 0$
the uniform distribution $1/3$ on the three connected structures with no role of the disconnected ones,
while for $t\to 1$ it recovers the uniform distribution $1/6$ on all the structures. 

Finally, we can consider the case where we add the total costs, MS, MY, and CL, together as a single
cost function. We then have 
\begin{equation}\label{vc}
v_{{\rm total\, cost}}=\frac{1-t^{3/2} + ((1-t^{3/2})^2 + 2 t^{17/6} )^{1/2}}{-1+t^{3/2}  + ((1-t^{3/2})^2 + 2 t^{17/6} )^{1/2}} \, .
\end{equation}
In this case we obtain
\begin{equation}\label{xitot}
\begin{array}{lcl}
\displaystyle{
 \frac{ v_{\rm total\, cost} }{ Z_{\rm total\, cost} } 
 } 
 & \sim_{t\to 0} &     \frac{1}{3}-\frac{t^{17/6}}{6}-\frac{t^{13/3}}{3}       + \cdots    \\[4mm]
\displaystyle{ 
 \frac{1}{ Z_{\rm total\, cost} }  
} 
& \sim_{t\to 0} &    \frac{t^{17/6}}{6} + \frac{t^{13/3}}{3}           + \cdots                        \\[7mm]
\displaystyle{ 
\frac{ v_{\rm total\, cost} }{ Z_{\rm total\, cost} } 
} 
& \sim_{t\to 1} &       \frac{1}{6} - \frac{\sqrt{2}}{8} (t - 1) + \frac{7\sqrt{2}}{48} (t - 1)^2      + \cdots    \\[4mm]
\displaystyle{ 
 \frac{1}{ Z_{\rm total\, cost} } 
 } 
 & \sim_{t\to 1} &    \frac{1}{6} +  \frac{\sqrt{2}}{8} (t - 1) -  \frac{7\sqrt{2}}{48} (t - 1)^2        + \cdots  
\end{array}
\end{equation}
Again, the behavior is analogous to the cases of weighting according to MS or CL separately: although
the specific form of the asymptotic distribution differs, for a fixed $0<t<1$, from the MS and CL cases,
the limit $t\to 0$ again gives the uniform $1/3$ distribution on the connected structure, with only IM
remaining as transformation cycle between these three configurations, and no contribution from
the multi-component structures, while for $t\to 1$ the asymptotic distribution becomes just the
uniform $1/6$ distribution over all workspaces. 

\smallskip

Note how, in each of the MS, CL, and total weighting, the contribution
of SM disappears in the limit $t\to 0$, and for small $t$ one has an asymptotic probability that favors IM
over SM, implying that one should expect to encounter rarely phenomena requiring an SM explanation. 
Note that EM will also necessarily stop contributing in the asymptotic behavior as the only
way to return to a multi-component structure to which EM can be applied is via an SM, but 
those become increasingly rare. The role of EM is however dominant in the early phases 
of structure formation, as the dynamics moves towards connected structures (and EM is the
only part of the dynamics that reduces the number of connected components). 

\smallskip

Thus, what we see from the full explicit analysis of the simplest example with 3 leaves, 
can be summarized as follows:
\begin{itemize}
\item  The MS, MY, and CL cost functions represent different and independent forms of 
optimization, which weight different forms of SM differently, not all of them 
identifying the same form of SM as preferable.
\item Weighting the action of Merge according to either MS or CL, or the combination of all MS, MY, CL costs,
gives quantitatively different Hopf algebra Markov chains and asymptotic distributions, depending on a
weight parameter $0<t<1$.
\item Weighting by MY costs has not effect on the dynamics, which remains the same bistochastic matrix
with uniform asymptotic distribution, as in the un-weighted case (because the MY effects of the EM and SM(3) 
components of the dynamics exactly compensate).
\item In all the other cases, the behavior of the resulting Hopf algebra Markov chain in the limits
$t\to 0$ and $t\to 1$ is the same, regardless of the choice of MS, CL, or combined cost function.
In the limit $t\to 0$ the asymptotic dynamics reduces to a cycle of IM transformations between the
three connected structures; while the limit $t\to 1$ recovers the un-weighted bistochastic matrix with
asymptotic distribution uniform over all 6 workspaces. 
\end{itemize}

In the presence of more complex structures with a larger set of leaves, the dynamics
will necessarily become more complicated to analyze, and we will not discuss the
general case here, but the simplest case is sufficient to clarify the different forms of 
optimization (MS, MY, CL, or combined) and their effects on the dynamics of the
Merge action on workspaces. 

\smallskip

\subsection{Strong connectedness in the general case}\label{GenGsec}

As in \S 1.9 of \cite{MCB}, we first observe that, since the operation $\cB_{(2)}$ involved in $\cK$ creates two
edges, anything in the image will have a non-empty set of edges. Thus, to obtain a graph that can have the
strong connectedness property we need to exclude the workspaces that only contain lexical items with
no other structure, and start with those where at least one External Merge has already been applied (since
all binary trees in $\fT_{\cS\cO_0}$ are full, these are exactly the forests with nonempty set of edges). We
can also fix the elements of $\cS\cO_0$ at the leaves, as the set of leaves is preserved by $\cK$. Thus,
for each choice of a multiset $L$ of elements of $\cS\cO_0$, we have a corresponding graph $G_L$, with
vertices given by all the forests $F\in \fF_{\cS\cO_0}$ with set of leaves $L(F)=L$ and with nonempty
set of edges, $E(F)\neq \emptyset$. 

We show that the graph is $G_L$ is strongly connected when the edges are given by the
nonzero entries of the matrix $K_X$ representing $\cK$ in the basis $X=\fF_{\cS\cO_0}$.
Note that, technically, this covers only External Merge and Sideward Merge, as incorporating 
Internal Merge requires a slightly more complicated form of $\cK$ as discussed in 
\S 1.9 of \cite{MCB}. However, if the graph is already strongly connected with EM and SM,
a fortiori it also is when IM is also considered.

Moreover, again because the operation $\cB_{(2)}$ adds two edges, we need to use
the form $\Delta^d$ of the coproduct, which in the quotient terms $T/^d F_{\underline{v}}$ on the
right-hand-side of the coproduct eliminates some edges (by cutting above the root
of the extracted accessible terms $T_{v_i}$ and by contracting to eliminate non-branching vertices). 
Thus, for example, the coproduct $\Delta^d(T)$ for $T$ with three leaves, contains a term of the form
$T_v \otimes \alpha$ where $\alpha$ is a single leaf and $T_v$ is a cherry tree, since $\alpha$ is the unique
full binary tree resulting from cutting above the root of the cherry tree.  

We discussed in \S \ref{ExGsec} the simplest example of the strong connectedness property illustrated in 
Figure~\ref{StrongConnectFig}, for the case of a set of three labelled leaves.  The strong
connectedness (with or without the IM arrows) is visible by direct inspection in this case.
One can see in this example that all the SM operations involved are coming from
coproduct terms that extract $F_{\underline{v}}=\alpha \sqcup \beta$ a forest consisting of
two single leaves, producing then, after applying $\cB_{(2)}\otimes {\rm id}$, a new separate
component of the form of a cherry tree $\fM(\alpha, \beta)$. 

We prove here the general result on strong connectedness. 
As we have seen in the simple example, it suffices
to show strong connectedness without Internal Merge to obtain also strong connectedness
for the full action of Merge, including IM. This slightly simplifies the argument. 
The first part of the statement is essentially the same argument given in \cite{MCB}, while
the second part is a refinement. 

\begin{prop}\label{strongconngen}
Consider the graph $G_L$ with vertices the
forests $F\in \fF_{\cS\cO_0}$ with $L(F)=L$ and a nonempty $E(F)\neq \emptyset$ and with
a directed edge from $F$ to $F'$ whenever there is either an EM of an SM transformation
from $F$ to $F'$. For any fixed set of leaves $L$, of any size $\ell=\# L$, the graph $G_L$ is
strongly connected. Moreover, consider the subgraph $G^a_L \subset G_L$ with the
same vertices and with edges corresponding to EM transformations and only  
particular SM transformation:
\begin{enumerate}
\item {\bf atomic SM(3)} with extraction and merge of two leaves: $$F \mapsto \fM(\alpha,\beta)\sqcup F/^d (\alpha \sqcup \beta)$$
\item {\bf atomic SM(1)} with extraction of one leaf and merge to another component:
$$F\mapsto \fM(\alpha, T) \sqcup \hat F/^d \alpha\, , \ \ \text{ for }  F=\hat F \sqcup T \, . $$
\end{enumerate}
For any fixed set of leaves $L$, of any size $\ell=\# L$, the graph $G^a_L$ is also
strongly connected.
\end{prop}

\proof Clearly the second part of the statement implies the first, as if $G^a_L \subset G_L$ is
strongly connected then $G_L$ also is, since they have the same set of vertices. However,
in the larger graph $G_L$ shortest paths that realize the strong connectedness property
are possible. The difference between the paths in $G^a_L$ and the shortest paths in $G_L$
is that they optimize for different properties: the paths in $G^a_L$ minimize Resource Restriction
violations at each step, while the minimal paths in $G_L$ minimize for Minimal Search and
the associated cost function. For this reason, instead of simply proving the statement for
$G^a_L$, we also comment explicitly on the difference between
paths in $G^a_L$ versus shortest paths in $G_L$.

\smallskip

We already know from the discussion in \S \ref{ExGsec} that the statement holds for $\ell=3$
where in fact the two graphs coincide, $G^a_L =G_L$.
We proceed by induction and assume that the statement holds for $\# L \leq \ell-1$. 
When we add a new labelled leaf $\alpha$, it is either added as a separate component in the 
resulting forest or it is added to one of the components of one of the previous forests. We can think of
all the possible ways of adding a new leaf $\alpha$ to a given tree $T$ as either merging it with EM,
forming $\fM(\alpha, T)$, or by splitting an edge $e$ of $T$ by a non-branching vertex and attaching it with a 
new edge to that vertex. The latter operation is the EC-violating insertion $T \lhd_e \alpha$ that we
analyzed earlier in this paper. 

\smallskip

Thus, the vertices of the graph $G_{L \sqcup \alpha}$ consist of two sets: the vertices labelled by
a forest of the form $F \sqcup \alpha$ with $F$ a vertex of $G_L$ and the vertices labelled either by
a forest of the form $F'\sqcup T \lhd_e \alpha$ where $F=F'\sqcup T$ is a vertex of $G_L$, or by $F'\sqcup \fM(\alpha, T)$.
The subgraph of $G_{L \sqcup \alpha}$ consisting only of vertices of the first type is strongly
connected because by induction hypothesis $G_L$ is and it also has the property that
$G^a_{L \sqcup \alpha}$ is strongly connected for the same reason. 

\smallskip

Thus, to prove strong connectedness of $G^a_{L \sqcup \alpha}$ (and hence $G_{L \sqcup \alpha}$)
it suffices to show that, for any vertex $F$ of $G_L$, there are directed paths in both directions between
the vertices $F'\sqcup T \lhd_e \alpha$ and $F \sqcup \alpha$, for any component $T$ of $F$,
that is, for $F=T\sqcup F'$, and direct paths in each direction between $F'\sqcup \fM(\alpha, T)$ and $F \sqcup \alpha$.

\smallskip

There is a directed path in $G^a_{L \sqcup \alpha}$ from $F \sqcup \alpha$ to 
$F'\sqcup T \lhd_e \alpha$, obtained by the following steps:
\begin{enumerate}
\item  If the edge $e$ is attached to a leaf $\beta$ of $T$, then there is
an SM transformation that uses the term $\alpha \sqcup \beta \otimes F'\sqcup T/^d \beta$ of the coproduct 
and forms a resulting forest $\fM(\alpha,\beta)\sqcup F'\sqcup T/^d \beta $. 
\item For every accessible term of $T/^d \beta$ (or of $T$ if $e$ is not attached to a leaf) 
that is a cherry tree $\fM(\gamma,\delta)$, perform an SM that extracts the term
$\gamma \sqcup \delta$. This chain of SM transformations results in 
a forest of the form $\sqcup_i T_i \sqcup F' \sqcup T/^d(\sqcup_i T_i)\sqcup \alpha$ (if $e$ is not
attached to a leaf), with $T_i=\fM(\alpha_i,\beta_i)$ all the cherry trees of $T$, or
$\sqcup_i T_i \sqcup \fM(\alpha,\beta)\sqcup F' \sqcup T/^d(\sqcup_i T_i)$ (if $\alpha$ is attached
to a leaf $\beta$), with $T_i$ all the cherries of $T$ not involving $\beta$. 
\item Check if the root vertex of one of the components obtained in the previous step is the vertex below
the edge $e$. If it is, then apply an External Merge of that component with $\alpha$. (If $e$ was attached to a leaf
this step was already done at Step (1).)
\item Act with all the possible External Merge operations on the result of the previous step that
give rise to accessible terms of $T$ (or of $T \lhd_e \alpha$ if $e$ was attached to a leaf). 
\item Repeat Step (3), if $\alpha$ is still a separate component. 
\item Since all the accessible terms of $T$ that could be built starting from the cherry trees of $T$
are already obtained, any further accessible term of $T$ (or of $T \lhd_e \alpha$, if $\alpha$ already merged into the
structure) requires merging one of the structures $T'$ obtained at this point with a single additional leaf $\beta'$ that is still
in the remaining quotient of $T$. Proceed by applying a chain of all the possible SM operations that extract $T' \sqcup \beta'$ 
and forms a structure $\fM(T',\beta')$ that is an accessible term of $T$ (or of $T \lhd_e \alpha$).
\item Repeat Step (3), if $\alpha$ is still a separate component. 
\item Repeat Step (4).
\item Repeat Step (3), if $\alpha$ is still a separate component. 
\item Repeat Step (6).
\item Keep repeating Steps (3)-(4)-(3)-(6)-(3) until all the leaves of $T$ have been extracted and
one is left with the structure $T \lhd_e \alpha \sqcup F'$.
\end{enumerate}

Note that there are shorter paths in $G_{L\sqcup \alpha}$ that are not in $G_{L\sqcup \alpha}^a$ as one
can see easily by starting the path with an SM operation that uses the term of the 
coproduct $T_v \sqcup \alpha \otimes F' \sqcup T/^d T_v$ forming
$\fM(T_v, \alpha) \sqcup F' \sqcup T/^d T_v$,
with $T_v$ the accessible term  with root at the vertex $v$ below $e$.

\smallskip

There is a directed path in $G^a_{L \sqcup \alpha}$ from $F'\sqcup T \lhd_e \alpha$
to $F \sqcup \alpha$, obtained by the following steps:
\begin{enumerate}
\item Extract by a chain of SM operations all the cherry trees $T_i=\fM(\alpha_i,\beta_i)$ of $T \lhd_e \alpha$, as above.
\item If one of these cherries is of the form $\fM(\beta,\alpha)$ then there is a leaf $\gamma$ of $T/^d (\sqcup_i T_i)$
such that $\fM(\beta,\gamma)$ is a cherry of $T$. Use the SM that extracts $\beta\sqcup \gamma$ and forms
the structure $\sqcup_j T_j \sqcup \fM(\beta,\gamma) \sqcup \alpha \sqcup T/^d (\gamma \sqcup_i T_i)$,
where $T_j$ are the remaing cherries of $T \lhd_e \alpha$, i.e., $\sqcup_j T_j \sqcup \fM(\beta,\gamma)$ is the
union of all the cherries of $T$. 
\item Keep building $T$ using EM and SM operations following the steps (4) and (6) of the construction
of the path in the opposite direction.
\item When the collection of accessible terms of $T$ constructed in this way contains both $T_v$, with root
the vertex $v$ below $e$ and $T_w$ with $w$ the twin of $v$ in $T$, the corresponding quotient of
$T\lhd_e \alpha$ that remains will be of the form $\fM(\alpha, T')$. Continue building $T$ following the
previous step, extracting with SM operation the remaining leaves of $T'$, until one obtains $F' \sqcup T\sqcup \alpha$,
with the component $\alpha$ as the last remaining quotient of $T \lhd_e \alpha$.
\end{enumerate}

Again, there is a much shorter paths in $G_{L\sqcup \alpha}$ that are not in $G^a_{L\sqcup \alpha}$,
as one can see by starting the path with an SM that extracts $T_v \sqcup T_w$ and merges them. 

\smallskip

For the remaining case of the vertex $F'\sqcup \fM(\alpha, T)$ there
is a single edge from $F \sqcup \alpha$ given by External Merge and in the opposite direction there is
a single edge given by an SM that extracts the two accessible terms $T_1$ and $T_2$ with
$T=\fM(T_1,T_2)$ and then merges them into $T$ resulting in $F \sqcup \alpha$.
This is the shortest path from $F'\sqcup \fM(\alpha, T)$ to $F \sqcup \alpha$ in $G_{L \sqcup \alpha}$
but is not a path in $G^a_{L \sqcup \alpha}$. To obtain a path of directed edges in $G^a_{L \sqcup \alpha}$ 
one can proceed in the same way discussed above for the paths from 
$F'\sqcup T \lhd_e \alpha$ to $F \sqcup \alpha$. 
\endproof

\medskip

The importance of the Hopf algebra Markov chain property for the action of Merge is
the fact of being able to study the long-term behavior of the dynamics, regardless
of knowledge of a specific initial distribution. Since the matrix $\hat K_X$ after the
rescaling of the basis $X$ is a stochastic matrix, it has the property that, for a vector
$\pi_0$ that is an arbitrary probability distribution, the new vector $\pi_1 = \pi_0 \hat K_X$
is also a probability distribution, with $\pi_{n+1}=\pi_n \hat K_X$ converging to the
limiting distribution $\xi = \xi \hat K_X$, the left-Perron Frobenius
eigenfunction. Thus, given any (not explicitly known) initial probability distribution
over the set of workspaces, with an assigned set of lexical items at the leaves, the
dynamics of free unconstrained applications of Merge will always converge to the
same stationary distribution $\xi$. This means that there is a probability distribution $\xi$
that is intrinsic to the structure formation process. It is completely independent of
the choice of the lexical items at the leaves and their semantic likelihood or lack of,
and is only sensitive to the possible chains of Merge operations that can obtain
a certain structure. The existence of these intrinsic (structural) probabilities can
 be used for comparing probabilities of sentence formations based uniquely on
 semantic proximity in texts (as in the case of LLMs and other manifestations
 of language in machines). This topic will be pursued elsewhere, but we
 mention it here as motivation.

\section{Alternative explanations and their mathematical formulation} \label{ThSec} 

As we discussed in \S \ref{ExtSec}, there are alternative explanations to the
use of Sideward Merge, for some of the linguistic phenomena analyzed in this paper. 
These alternative explanations rely in part on additional structure in the model, that is
not at present included in the mathematical formalization of Minimalism
of \cite{MCB}, and its subsequent elaborations. 

\smallskip
\subsection{Box Theory for multiple wh-movement}\label{BoxSec}

One such extensions is Box Theory, which is proposed here as an alternative
explanation to the Sideward Merge derivation in the case of multiple
wh-movement and the ``tucking in" phenomenon in Bulgarian.

\smallskip

Box Theory is an extension of the formulation of Merge in the Strong Minimalist
Thesis framework (as presented in \cite{ChomskyElements} and \cite{MCB}).
The setting of Box Theory was developed in Chomsky's 2023 preprint  ``Displacement", \cite{Chomsky23b},
and in his Keio Lectures of 2023, \cite{ChomskyKeio}. This further and still ongoing
development of the theory raises the question of how Box Theory fits with the
mathematical formulation of \cite{MCB}. This is a separate question that will need
to be addressed elsewhere, so we will not discuss it here, but we can make some
preliminary comments. There are two main aspects in which Box Theory differs from
the form of SMT developed in  \cite{ChomskyElements} and \cite{MCB} and formalized
mathematically in \cite{MCB}. One is the ``boxing" of elements moved by IM to
Spec-of-phase position (blocked from further Merge operations, eliminating
successive cyclic movement) and the second is the fact that the C/Q phase head can
``access" the boxed elements. The first requirement of ``boxing" is easily accounted for in the
mathematical model, via the bud generating system of the colored operad of
phases as introduced in \cite{MHL}. Indeed, it suffices to introduce a small change
to the generators. Instead of allowing all the possible generators of the form
$$ T^\fs_{\fc, (1,\fm)} = \Tree[ .$\fs^\downarrow_{\omega'}$ [ $\fc^\downarrow_\omega$ $(1,\fm)$ ] ] $$
which include, when $\fc^\downarrow_\omega=\fs^\downarrow_{\omega}$, the possibility of
cyclic movement, one allows only generators of the form
$$ T^\fs_{\fc, (1,\fm)} = \Tree[ .$\fs^\downarrow_{\omega'}$ [ $\fs^\downarrow_\omega$ $(1,\fm)$ ] ] \ \ \  \text{  with } \omega\in \Omega_{\fh,{\rm lex}}   $$
so that no further movement happens after first reaching the
Spec-of-phase position $\fs^\downarrow_{\omega'}$. This implements the ``boxing". The part that is
more delicate, in implementing Box Theory within the formalism of \cite{MCB} and \cite{MHL}, 
is the phase head C/Q ``accessing" the boxed elements, as here ``to access" is in need of a precise
mathematical formulation. This is likely also possible in terms of bud generating systems of
the colored operad of phases, but it will be discussed elsewhere. 

\smallskip
\subsection{Amalgam in Externalization}\label{AmalgSecExt}

The other alternative explanation that requires further work for implementing within
the mathematical model of \cite{MCB} is the proposed Amalgam in Externalization,
as an alternative to the use of Sideward Merge. This suggest the possibility of
certain operations at the interface of syntax and morphology. 

\smallskip

A formulation of the interface between syntax and morphology that is consistent with
the mathematical formulation of Merge of \cite{MCB} was developed in \cite{SentMar}. 
We will work here within that framework. We summarize quickly the main aspects of
the Morphosyntax formulation of  \cite{SentMar}. 

\smallskip 
 
Similarly to how syntactic objects are built as the free non-associative commutative
magma generated by the set $\cS\cO_0$ of lexical items and syntactic features, bundles of
morphological features are described by the same algebraic mechanism, with a finite
generating set $\cM\cO_0$ of morphological features and the free non-associative
commutative magma
$$ \cM\cO= {\rm Magma}_{na,c}(\cM\cO_0, \fM^{\rm morph})\, . $$
The resulting objects $M\in \cM\cO$ are interpreted as morphological bundles of features
which are matched to syntactic features in $\cS\cO_0$ via a ``feature correspondence"
$\Gamma_{SM}$. In fact, in order to allow for operations of
Distributed Morphology like impoverishment, one extends $\cM\cO$ to
include non branching vertices, by acting on elements of $\cM\cO$ via
the coproduct $\Delta^\rho$. We will not discuss the details of these aspects
here, since they are not relevant for the specific operations we are interested
in discussing, so we just refer the reader to \cite{SentMar} for more details. 
What is relevant here, from the construction of \cite{SentMar}, is the
notion of Morphosyntactic trees, obtained via operadic insertion 
(through the feature correspondence) of morphological bundles of features
at the leaves of syntactic objects. The relevant mathematical formulation of \cite{SentMar} 
involves algebras over operads and correspondences of algebras over
operads. 
 
 \medskip
 
 We the main notions from \cite{SentMar} that we need to use here. For our
 present purposes we will not make the distinction between morphological
 objects $M\in \cM\cO$ and what is called in \cite{SentMar} ``extended
 morphological objects", since that does not play a role in the cases 
 we are considering here. Thus, we just consider here a set of
 morphological workspaces $\cW(\cM\cO)$ that consists of forests
 whose components are morphological trees in $\cM\cO$. 
 
\medskip

We can now revisit the proposal of Amalgamation in Externalization described in  \S \ref{amalgSec}
and formulate it in terms of the mathematical model of Morphosyntax of \cite{SentMar}. As in \cite{MCB},
\cite{MHL}, and \cite{SentMar}, the Merge operad $\cM =\{ \cM(n) \}_{n\in \N}$ has $\cM(n)$ the set
of full non-planar binary rooted trees with $n$ leaves (with no leaf-decorations) and with the operad compositions
$$ \gamma: \cM(n) \times \cM(k_1)\times \cdots \times \cM(k_n) \to \cM(k_1+\cdots+ k_n) $$
that match the $i$-th leaf of $T\in \cM(n)$ to the root of $T_i\in \cM(k_i)$, and correspondingly,
in terms of a single insertion at a single leaf $e\\l \in L(T)$,
$$ \circ_\ell : \cM(n) \times \cM(k) \to \cM(n+k-1) $$
matching the leaf $\ell$ of $T\in \cM(n)$ to the root of $T'\in \cM(k)$. 

As discussed in \cite{MCB}, the set $\cS\cO$ of syntactic objects is an algebra over the Merge operad $\cM$, namely
there is an operad action
$$ \gamma_{\cS\cO}: \cM(n) \times \cS\cO^n \to \cS\cO $$
that grafts the root of each $T_\ell \in \cS\cO$ to the $\ell$-th leaf of $T\in \cM$. 

In turn, as shown in \cite{SentMar}, there are operad insertion maps that combine
syntactic objects $T\in \cS\cO$ and morphological objects $M\in \cM\cO$ to form 
morphosyntactic objects $\gamma_{\cS\cO, \cM\cO}(T,M_1,\ldots, M_n)\in \cM\cS$, where
the matching between the data in $\cS\cO_0$ at the leaves of syntactic trees and
the bundles of morphological features $B\in \cP(\cM\cO_0)$ (the powerset of $\cM\cO_0$) 
at the root of morphological trees is  ruled by a correspondence $\Gamma_{SM}\subset \cP(\cM\cO_0)\times \cS\cO_0$, 
with insertion maps
$$ \gamma_{\cS\cO, \cM\cO}: \cS\cO_n \times \cM\cO^n \to \cM\cS \, , $$
whenever the $\alpha_\ell\in \cS\cO_0$ at the leaves of the syntactic object $T$
and $B_\ell$ at the root of the morphological object $M_\ell$ satisfy $(B_\ell, \alpha_\ell)\in \Gamma_{SM}$. 
The resulting set $\cM\cS$ of morphosyntactic trees is again an algebra over the Merge operad $\cM$.

We also use here the notation
$$ \circ_\ell (T, M) \, , \ \ \text{ for } T\in \cS\cO \ \ \text{ and } \  M\in \cM\cO $$
for each individual leaf-insertion with matching $(B_\ell,\alpha_\ell)\in \Gamma_{SM}$,
with $\gamma_{\cS\cO, \cM\cO}(T,M_1,\ldots, M_n)$ obtained as the composition of the $\circ_\ell$
at all $\ell\in L(T)$. 

Syntactic objects $T\in \cS\cO$ determine operators on morphosyntactic trees of the form
$$ \cK_T= \sqcup \circ (\gamma_{\cS\cO, \cM\cO}(T,\ldots) \otimes {\rm id}) \circ \delta_{\underline{\alpha},\underline{B}} \circ \Delta $$
where $\delta_{\underline{\alpha},\underline{B}}$ ensures the matching conditions $(B_\ell, \alpha_\ell)\in \Gamma_{SM}$.

The operations of fusion and fission of Distributed Morphology can then be written as
operations on morphosyntactic trees in terms of these operad insertions (and other operations like Impoverishment and
Obliteration are obtained from these and the coproduct). A fusion operation transforms a
morphosyntactic tree $T=\gamma_{\cS\cO,\cM\cO}(T',(S_\ell)_{\ell\in L(T')})$ into a
morphosyntactic tree
$$ \gamma_{\cS\cO,\cM\cO}(T/^c \fM(\alpha_1,\alpha_2), S_{12}, (S_\ell)_{\ell\in L(T')\smallsetminus \{ \ell_1, \ell_2 \}} ) $$
with
$$ S_{12} = \ \ \ \  
\Tree[ .$(B_v=B_{v_1}\cup B_{v_2},\alpha_v)$  \qroof{tree $S_1$}.$B_{v_1}$  \qroof{tree $S_2$}.$B_{v_2}$ ]   $$
with $\alpha_v$ the head of $\fM(\alpha_1,\alpha_2)$.
This can be seen as pushing upward the morphosyntactic boundary, with the two syntactic leaves $\alpha_1,\alpha_2$
becoming part of the morphological object $S_{12}$. The fission operations, which we do not write out
explicitly here (see \cite{SentMar}) have the opposite effect of pushing the morphosyntactic boundary downward.

\smallskip

Consider then the following data:
\begin{itemize}
\item $\alpha\in \cS\cO_0$ a lexical item,
\item $T \in \cS\cO$ a syntactic object,
\item $\ell \in L(T)$ with $\alpha_\ell =\alpha$,
\item $\ell' \in L(T)$ such that $\ell'$ is a head that locally c-commands $\ell$
\item $M \in \cM\cO$ a morphological object with feature bundle $B$ at the root, such that
$$ (B,\alpha) \in \Gamma_{SM} $$
\item $\alpha_{\ell'}=\alpha'$ 
\item $(\alpha', \phi')\in \Gamma_{SM}$ for a feature $\phi'$
\item $M'=\Tree[ $\phi'$ $M$ ] \in \cM\cO$ is a morphological object.
\end{itemize}
Then the Amalgamation in Externalization can be seen as the 
statement that the operad insertion 
$$ \circ_{\ell'} (T, \Tree[ $\phi'$ $B$ ] ) $$
is possible whenever the insertion
$$ \circ_{\ell'} (\circ_\ell (T,B), \phi' ) $$
is possible. Given this, the operation of Amalgamation in Externalization
can be described in the following way, illustrated for 
$$ T= \Tree[ $\alpha'$ [ .$v_\alpha$ $\beta$ [ $\alpha$ $T'$ ] ] ] $$
with $v_\alpha$ the maximal projection of $\alpha$, and a
corresponding morphosyntactic object 
$$ \tilde T =\gamma_{\cS\cO,\cM\cO}(T, \{ M_u \}_{u \in L(T)} ) = \Tree[ [ .$\alpha'$ $\phi'$ ] [ [ .$\beta$ $M_\beta$ ] [ [ .$\alpha$ $M$ ] {$\gamma_{\cS\cO,\cM\cO}(T', \{ M_u \}_{u\in L(T')})$} ] ] ] $$
with $M_\ell=M$ and $M_{\ell'}=\phi'$. 
\begin{enumerate}
\item Start with a morphosyntactic workspace of the form 
$$ \phi' \sqcup  M_\beta  \sqcup \Tree[ $M$ 
{$\gamma_{\cS\cO,\cM\cO}(T', \{ M_u \}_{u\in L(T')})$} ] $$
\item Select the term of the coproduct of the form
$$  \phi' \sqcup M \otimes M_\beta \sqcup \Tree[ \sout{$\alpha$} {$\gamma_{\cS\cO,\cM\cO}(T', \{ M_u \}_{u\in L(T')})$} ] $$
which we read as
$$ \Tree[ .$\alpha'$ $\phi'$ ] \sqcup \Tree[ .$\alpha$ $M$ ] \otimes \Tree[ .$\beta$ $M_\beta$ ] \sqcup \Tree[ \sout{$\alpha$}
 {$\gamma_{\cS\cO,\cM\cO}(T', \{ M_u \}_{u\in L(T')})$} ] $$
\item we can form from this a new morphosyntactic workspace of the form
$$ \Tree[ [ .$\alpha'$ $\phi'$ ] [ .$\alpha$ $M$ ] ] \sqcup \Tree[ .$\beta$ $M_\beta$ ] \sqcup \Tree[ \sout{$\alpha$}  {$\gamma_{\cS\cO,\cM\cO}(T', \{ M_u \}_{u\in L(T')})$} ] $$
\item the fusion operation of morphosyntax then gives
$$ \Tree[ .$\alpha'$   $\phi'$ $M$  ] \sqcup  \Tree[ .$\beta$ $M_\beta$ ] \sqcup \Tree[  \sout{$\alpha$}  {$\gamma_{\cS\cO,\cM\cO}(T', \{ M_u \}_{u\in L(T')})$} ] $$
where the root of the first tree is labelled by $\alpha'$ because the local c-command relation between $\alpha'$
and $\alpha$ implies that $\alpha'$ becomes the head of $\fM(\alpha',\alpha)$ hence the vertex label in the fusion operation
(see \cite{SentMar}). 
\item We can then produce from this the morphosyntactic structure
$$ \Tree[ [ $\phi'$ $M$  ] [ $M_\beta$ [ \sout{$\alpha$} {$\gamma_{\cS\cO,\cM\cO}(T', \{ M_u \}_{u\in L(T')})$} ] ] ] $$
\end{enumerate}
Note that, formulated in this way, the Amalgamation in Externalization operation is an SM operation
accompanied by a fusion operation. This in particular implies that, in terms of cost counting, its
cost is at least as large as that of the SM operation involved (equal if the fusion is a zero cost operation). 

\smallskip

We see this, in the case of example of Amalgamation in Externalization described in  \S \ref{amalgSec}
in the following way. One has in that case a syntactic object of the form
$$ T = \Tree[ EA [ INFL [ \text{\sout{ EA }} [ $v^*$ [ Root  IA ] ] ] ] ]  $$
One starts with a syntactic workspace of the form 
$$ F = {\rm EA} \sqcup {\rm INFL} \sqcup v^* \sqcup \Tree[ Root IA ]   $$
and with a morphological object $M \in \cS\cO$ with $({\rm Root}, B)\in \Gamma_{SM}$ for
$B$ the bundle at the root of $M$, and with $(v^*, \phi')\in \Gamma_{SM}$ and $({\rm INFL}, \phi'')\in \Gamma_{SM}$. One then
obtains a morphosyntactic object (from an SM followed by a fusion) 
$$ \Tree[ .$v^*$  $\phi'$ $M$  ] $$
and one can then form
$$ \Tree[ [ .EA {$M_{\rm EA}$} ]   [ [ .$v^*$  $\phi'$ $M$  ]  [ \sout{Root} [ .IA $M_{\rm IA}$ ] ] ] ] $$
from which a further combination of an SM and fusion then gives a morphosyntactic object
$$ \Tree[ .INFL $\phi''$ [  $\phi'$ $M$  ] ] $$
from which we can then form
$$ \Tree[ [ .INFL $\phi''$ [   $\phi'$ $M$  ] ]  [ [ .EA {$M_{\rm EA}$} ]   [ \sout{$v^*$}  [ \sout{Root} [ .IA $M_{\rm IA}$ ] ] ] ]  ] $$
and from this
$$ \Tree[ EA [ [ .INFL $\phi''$ [ $\phi'$ $M$  ] ]  [ \sout{EA}   [ \sout{$v^*$}  [ \sout{Root} [ .IA $M_{\rm IA}$ ] ] ] ]  ] ] $$
This fits the description of Amalgamation in Externalization of \S \ref{amalgSec} into the framework
of \cite{MCB} and \cite{SentMar}, but in order to do so we reduce this operation to a combination of an SM
and a fusion operation of morphosyntax, hence this is again ends up relying on the same SM operation.

 \smallskip
 \subsection{Amalgam in I-language and FormCopy}\label{FCsec}
 
 We also comment here a bit more in depth on the proposal for
 Amalgam in I-language using EM and FormCopy as described in
 \S \ref{amalgSec}, from the perspective of the mathematical
 formulation of \cite{MCB}. As mentioned in  \S \ref{amalgSec}
 the issue here is the appropriate cost counting for this proposed
 formulation and its comparison with the cost counting for the
 SM-based explanation. 
 
 \smallskip
 
 Consider again the example discussed in  \S \ref{amalgSec}. We can start with
 a workspace of the form
 $$ F_1 = \text{John} \, \sqcup \, \Tree[ $v^*$ \text{read}$_1$ ]\, \sqcup \, \Tree[ INFL [ $v^*$ read$_2$ ] ] \, \sqcup\,
\Tree[ read$_3$ [ a  book ] ]   $$
An External Merge operation gives
$$ \text{John} \, \sqcup \, \Tree[ INFL [ $v^*$ read$_2$ ] ] \, \sqcup\,
\Tree[ [ $v^*$ \text{read}$_1$ ] [ read$_3$ [ a  book ] ]  ] $$
further External Merge operations give
$$ \Tree[ INFL [ $v^*$ read$_1$ ] ] \, \sqcup\, \Tree[  John   [ [ $v^*$ \text{read}$_2$ ] [ read$_3$ [ a  book ] ]  ] ] $$
and then
\begin{equation}\label{Tquot}
 \Tree[ [ INFL [ $v^*$ read$_1$ ] ] [  John   [ [ $v^*$ \text{read}$_2$ ] [ read$_3$ [ a  book ] ]  ] ] ] 
\end{equation} 
Internal Merge then transforms this into
$$ \Tree[ John [ [ INFL [ $v^*$ read$_1$ ] ] [  \text{\sout{John}}   [ [ $v^*$ \text{read}$_2$ ] [ read$_3$ [ a  book ] ]  ] ] ] ] \, . $$
The idea here is that one then performs FormCopy (for the two instances  read$_2$ and read$_3$, and then
of the two instances $\{ v^*, {\rm read}_1 \}$ and $\{ v^*, {\rm read}_2 \}$) 
and cancellation of the deeper copies resulting in 
\begin{equation}\label{Tquot2}
  \Tree[ John [ [ INFL [ $v^*$  read ] ] [  \text{\sout{ John }}   [ \sout{$\{ v^*, \text{read} \}$}  [ \text{\sout{ read }} [ a  book ] ]  ] ] ] ] 
\end{equation}  
 
 \smallskip
 
 According to the mathematical formulation of \cite{MCB}, cancellation of the deeper copies is
 a quotient operation that is performed by the coproduct (in the right channel) accompanied by
 extraction of the accessible term (in the left channel), so it does not take place as an operation
 on its own. In \S 3.8.2 of \cite{MCB}. in the formulation of Obligatory Control, the combination
 of FormCopy and cancellation of the deeper copies was formulated as a ``restriction of diagonals"
 in operad insertions at the leaves of syntactic objects in the Merge operad $\cM$. 
 
 \smallskip
 
 In order to examine the cost of such operations and compare them to the cost of the
 derivation based on Sideward Merge, we reformulate these operations in the following
 way. We can consider the combination of FormCopy and cancellation of the deeper copies
 as a single quotient operation, where FormCopy glues the corresponding parts of the
 graph together. Viewed in this way, the resulting graph is no longer a tree, and there is 
 no actual ``cancellation" of copies, rather the identification produced by the equivalence 
 relation that defines the quotient. This is, however, equivalent to the combination of
 FormCopy and cancellation of the deeper copie, since the result of these two operations
 can be seen as a {\em fundamental domain} for the quotient.
 
 \smallskip
 
 In the example discussed above, we would be taking two quotient operations.
A first one identifies the two substructures $\{ v^*, {\rm read} \}$ in 
\eqref{Tquot} resulting in a graph of the form
 \begin{center}
\begin{tikzpicture}[-latex ,auto ,node distance =1 cm]
\node (N1) {$\bullet$};
\node (N2) [below left=of N1] {$\bullet$};
\node (N3) [below right=of N1] {$\bullet$};
\node (EA) [below left=0.4cm of N3] {John};
\node (N4) [below right=of N3] {$\bullet$};
\node (V) [below left=2cm of N4] {$\bullet$};
\node (N5) [below right=of N4] {$\bullet$};
\node (A) [below left=of V] {$v^*$};
\node (B) [below right=of V] {$\text{read}$};
\node (C) [below left=0.4cm of N5] {read};
\node (D) [below right=of N5] {$\bullet$}; 
\node (D1) [below right=0.3cm of D] {book}; 
\node (D2) [below left=0.3cm of D] {a}; 
\node (E) [below left=of N2] {INFL};
\path[-] (V) edge (A);
\path[-] (V) edge (B);
\path[-] (N1) edge (N2);
\path[-] (N2) edge (E);
\path[-] (N1) edge (N3);
\path[-]  (N3) edge (EA);
\path[-]  (N2) edge (V);
\path[-]  (N3) edge (N4);
\path[-]  (N4) edge (N5);
\path[-]  (N5) edge (D);
\path[-]  (D) edge (D1);
\path[-]  (D) edge (D2);
\path[-]  (N5) edge (C);
\path[-] (N4) edge (V);
 \end{tikzpicture}
\end{center}
One can then further proceed to identify the two ``read" leaves, resulting
in graph of the form
\begin{center}
\begin{tikzpicture}[-latex ,auto ,node distance =1 cm]
\node (N1) {$\bullet$};
\node (N2) [below left=of N1] {$\bullet$};
\node (N3) [below right=of N1] {$\bullet$};
\node (EA) [below left=0.4cm of N3] {John};
\node (N4) [below right=of N3] {$\bullet$};
\node (V) [below left=2cm of N4] {$\bullet$};
\node (N5) [below right=of N4] {$\bullet$};
\node (A) [below left=of V] {$v^*$};
\node (B) [below right=of V] {$\text{read}$};
\node (D) [below right=of N5] {$\bullet$}; 
\node (D1) [below right=0.3cm of D] {book}; 
\node (D2) [below left=0.3cm of D] {a}; 
\node (E) [below left=of N2] {INFL};
\path[-] (V) edge (A);
\path[-] (V) edge (B);
\path[-] (N1) edge (N2);
\path[-] (N2) edge (E);
\path[-] (N1) edge (N3);
\path[-]  (N3) edge (EA);
\path[-]  (N2) edge (V);
\path[-]  (N3) edge (N4);
\path[-]  (N4) edge (N5);
\path[-]  (N5) edge (D);
\path[-]  (D) edge (D1);
\path[-]  (D) edge (D2);
\path[-]  (N5) edge (B);
\path[-] (N4) edge (V);
 \end{tikzpicture}
\end{center}
We can then view \eqref{Tquot2} as a tree that represents a fundamental domain for this
resulting quotient. 

\smallskip

This geometric interpretation as a quotient operation is useful to estimate an
associated cost to this operation of FormCopy and deletion of the deeper copies.
This allows us to compare the cost-effectiveness of this possible derivation
compared to the one via Sideward Merge. We compare costs according to
the different cost functions that we considered in this paper: Minimal Yield,
Complexity Loss, and (bottom-up) Minimal Search.

\smallskip
\subsubsection{Minimal Yield costs}

The Minimal Yield costs are associated to changes in the number of connected
components and number of accessible terms.

In a Sideward Merge derivation, we would have first an SM(1) operation
\begin{equation}\label{SM1aFC}
 v^* \sqcup \Tree[ read [ a book ] ] \stackrel{\rm SM}{\longrightarrow} 
\Tree[ $v^*$ read ] \sqcup \Tree[ \sout{read} [ a book ] ] 
\end{equation}
and then another SM(1) operation
\begin{equation}\label{SM1bFC}
{\rm INFL} \sqcup \Tree[ John [ [ $v^*$ read ] [ \sout{read} [ a book ] ] ] ] \stackrel{\rm SM}{\longrightarrow} 
 \Tree[ INFL [ $v^*$ read ] ]  \sqcup \Tree[ John [ \sout{$\{ v^*, {\rm read} \}$}  [ \sout{read} [ a book ] ] ] ]  \, .
\end{equation}
As we have seen SM(1) operations maintain the same number of connected components
and either also maintain the same number of accessible terms, with the $\Delta^d$ coproduct
which would give zero MY cost to these two SM operations
(or increase it by $+1$ with the $\Delta^c$ coproduct, which would give a MY cost of $+2$). 
There are, additionally, $5$ EM operations, each of which
has MY cost $+1$, so the total MY cost for a derivation based on EM+SM is $5$ (or $7$ with
the contraction coproduct).

In the derivation described above with FC and cancellation
of deeper copies, one first needs $8$ EM operations to form the tree  \eqref{Tquot},
for a total cost of $8$, and then one forms the two quotient operations described
above. These do not change the number of connected components, but alter the
number of accessible terms, reducing it by $-3$ and $-1$, respectively, for a
total YM cost of $4$.

\smallskip
\subsubsection{Complexity Loss costs} 

The Complexity Loss cost is computed in the case of trees by counting the maximal loss in
number of leaves over components, which is $+1$ for SM(1) operations and $+2$ for SM(2)
and SM(3) operations and zero for EM and SM. Thus, with the SM-derivation with
SM operations   \eqref{SM1aFC} and \eqref{SM1bFC} the total CL cost is $2$.

In the case of the quotient operations for FC with cancellation of the deeper copies
the total loss out of the initial number of leaves of the tree \eqref{Tquot} is $3$.

Adding up all the Resource Restriction costs (MY and CL) one gets
$5+2=7$ (or $7+2=9$ with the contraction coproduct) for the SM derivation
versus $4+3=7$ for FC with cancellation of deeper copies. In the case
of the deletion coproduct these costs match, while in the case of
the contraction coproduct the cost of EM+SM is higher.

\smallskip
\subsubsection{Minimal Search costs}

Consider then the cost function for Merge operations that we discussed in \S \ref{MSsev},
where we assigned costs to the extraction of accessible terms $T_v$ and to
the quotient operation $T/T_v$ on the basis of the size (in terms of number of
leaves, the grading of the Hopf algebra), with
$$ \fc(T_v) =\frac{\ell(T_v)}{\ell(T)} \, , \ \ \  \fc(T/T_v)=1-\fc(T_v) \, , \ \ \  
\fc(\fM(A, B))=\fb(A,B) -\fc(A)-\fc(B) $$
with $\fb(A,B)=1$ if $A,B$ are taken from the same component of the workspace
and $\fb(A,B)=2$ if they are from different components, so that for External and Internal Merge
$$ \fc(\fM(T,T'))=2-\fc(T)-\fc(T')=0, \ \ \ \text{ and } \ \ \ \fc(\fM(T-v, T/T_v))=1-\fc(T_v)-\fc(T/T_v)=0 $$
while for a Sideward Merge of type SM(1)
$$ \fc(T_v,T')= 1-\fc(T_v)  \, . $$

\smallskip

 To extend this cost function to the case of quotients of trees given by graphs that are 
 no longer trees (as in the description given above of FormCopy with cancellation
 of deeper copies), recall that, in the case of a full binary tree $T$ the number of 
 leaves $\ell(T)$ is related to the number of vertices by $v(T)=2 \ell(T) -1$, hence
 the cost can equivalently be written as
 $$ \fc(T_v) =\frac{\ell(T_v)}{\ell(T)}= \frac{v(T_v)+1}{v(T)+1} \ \ \ \text{ and } \ \ \
 \fc(T/T_v) =\frac{\ell(T/T_v)}{\ell(T)}= \frac{v(T/T_v)+1}{v(T)+1}\, . $$
 The number of vertices is a well defined quantity for arbitrary graphs so that
 we can set
 $$ \fc(G/\sim)=\frac{v(G/\sim)+1}{v(G)+1} \, . $$
 In the case of large graphs this would behave like $v(G/\sim)/v(G)$.
 
 In the case of the example given above, $v(T)=17$ is the number of vertices of the tree \eqref{Tquot},
 while the first quotient operation gives a graph $G=T/\sim$ with number of vertices $v(G)=14$, while
 the second quotient operation gives a graph with number of vertices $v(G/\sim)=13$. Thus, the overall
 cost is given by 
 $$ \fc(T/\sim)+ \fc(G/\sim)=\frac{14}{17} + \frac{13}{14} \sim 1.752\, . $$
 
 We can then compare this cost of the FormCopy operation with 
cancellation of deeper copies with the SM costs involved
in the derivation via Sideward Merge. In that case, we have the two
SM operations of \eqref{SM1aFC} and \eqref{SM1bFC}. The first one has
cost $\fc(T_v,T')= 1-\fc(T_v)=2/3$, and a second one 
has a cost of $\fc(T_v,T')= 1-\fc(T_v)=1-2/5=3/5$, for a total cost of
 $2/3+3/5\sim 1.266$. Thus, for these MS-costs, the EM+SM derivation
 is less costly.
 
\subsubsection{On comparison of cost functions} 
It seems clear from inspecting this example that there is no a priori argument
on the basis of these cost functions for ruling out one of the two methods in
favor of the other as the cost comparisons are sensitive to the specific
example and are not a priori bound to be higher or lower in one of the
two methods. This perhaps suggests that these might not be the optimal
cost functions to consider. The explicit example of the Merge Hopf algebra
Markov chain dynamics that we discussed in \S \ref{HopfMArkovSec}
also suggests that some more natural combinatorial way of weighting the arrows of the
Merge graph may lead to a better and simpler analysis of the resulting
dynamics and incorporate in a more natural way the different costs
separately accounted for by MS, MY, CL. We refer the readers to \cite{MarSkig} for further discussions.

\subsection{Sideward Merge and the flexible boundary of Morphosyntax}\label{MSboundarySec}

We return ti discuss some example mentioned earlier in the
paper, in terms of the interaction between two different mechanisms:
the minimal form of Sideward Merge discussed in this paper and
the flexible boundary of morphosyntax discussed in \cite{SentMar}.

We review in this light the case of the clusters of clitics and the clusters of
multiple wh-fronting. In both cases we have argued for a refinement
of the colored operad generators of phases and theta roles that will
allow for a splitting of the Spec-of-phase position that will permit
the insertion of a cluster of either of these kinds (depending on the
generator used). 

There is another possible way in which one can read this splitting of
the Spec-of-phase position, namely by allowing back and forth movement 
of the Morphosyntactic boundary with the operations of fusion and fission
as described in \cite{SentMar}. In the case of the multiple wh-movement
this interpretation would become then similar to the ``quantifier absorption"
(Higginbotham \& May \cite{Higg}) hypothesis for the explanation of multiple 
wh-fronting proposed in which the multiple wh-cluster would act as a single
unit. This would indeed be the case if the morphosyntactic boundary
is moved upward by fusions: this whole cluster becomes a morphological tree, 
with a single syntactic root-vertex that merges to the Spec-of-phase position.
It would then expand back into a syntactic tree by fissions to yield the resulting
cluster. This interpretation has the problem that there is no strong empirical 
evidence or conceptual considerations for the existence of such clusters as
morphological objects, so this intermediate step would have no empirical
confirmations. 

\smallskip

\subsection*{Acknowledgment} The first author was supported for this work 
by NSF grants DMS-2506176 and DMS-2104330, by Caltech's  Center 
of Evolutionary Science, and Caltech’s T\&C Chen Center for Systems Neuroscience.

\end{document}